\documentclass[sn-mathphys,iicol]{sn-jnl}% Math and Physical Sciences Reference Style
\jyear{2021}%

\theoremstyle{thmstyleone}%
\theoremstyle{thmstyletwo}%

\theoremstyle{thmstylethree}%

\begin{document}

\title[Article Title]{Ultra-high Resolution Image Segmentation via Locality-aware Context Fusion and Alternating Local Enhancement}


\author{Wenxi Liu, Qi Li, Xindai Lin, Weixiang Yang, Shengfeng He, Yuanlong Yu}

% \author*[1,2]{\fnm{First} \sur{Author}}\email{iauthor@gmail.com}



% \affil*[1]{\orgdiv{Department}, \orgname{Organization}, \orgaddress{\street{Street}, \city{City}, \postcode{100190}, \state{State}, \country{Country}}}

% \affil[2]{\orgdiv{Department}, \orgname{Organization}, \orgaddress{\street{Street}, \city{City}, \postcode{10587}, \state{State}, \country{Country}}}

% \affil[3]{\orgdiv{Department}, \orgname{Organization}, \orgaddress{\street{Street}, \city{City}, \postcode{610101}, \state{State}, \country{Country}}}


\abstract{Ultra-high resolution image segmentation has raised increasing interests in recent years due to its realistic applications. In this paper, we innovate the widely used high-resolution image segmentation pipeline, in which an ultra-high resolution image is partitioned into regular patches for local segmentation and then the local results are merged into a high-resolution semantic mask. In particular, we introduce a novel locality-aware context fusion based segmentation model to process local patches, where the relevance between local patch and its various contexts are jointly and complementarily utilized to handle the semantic regions with large variations. Additionally, we present the alternating local enhancement module that restricts the negative impact of redundant information introduced from the contexts, and thus is endowed with the ability of fixing the locality-aware features to produce refined results. Furthermore, in comprehensive experiments, we demonstrate that our model outperforms other state-of-the-art methods in public benchmarks. Our released codes will be available at: https://github.com/liqiokkk/FCtL.}

\keywords{ultra-high resolution image segmentation, geo-spatial image segmentation}



\maketitle

\newcommand{\wx}[1]{\textcolor{black}{#1}}
\newcommand{\wxr}[1]{\textcolor{black}{#1}}
\newcommand{\wxb}[1]{\textcolor{black}{#1}}
\newcommand{\wxo}[1]{\textcolor{black}{#1}}

\newcommand{\lqq}[1]{\textcolor{black}{#1}}

\begin{figure}[t]
    \centering
    \includegraphics[width=0.48\textwidth]{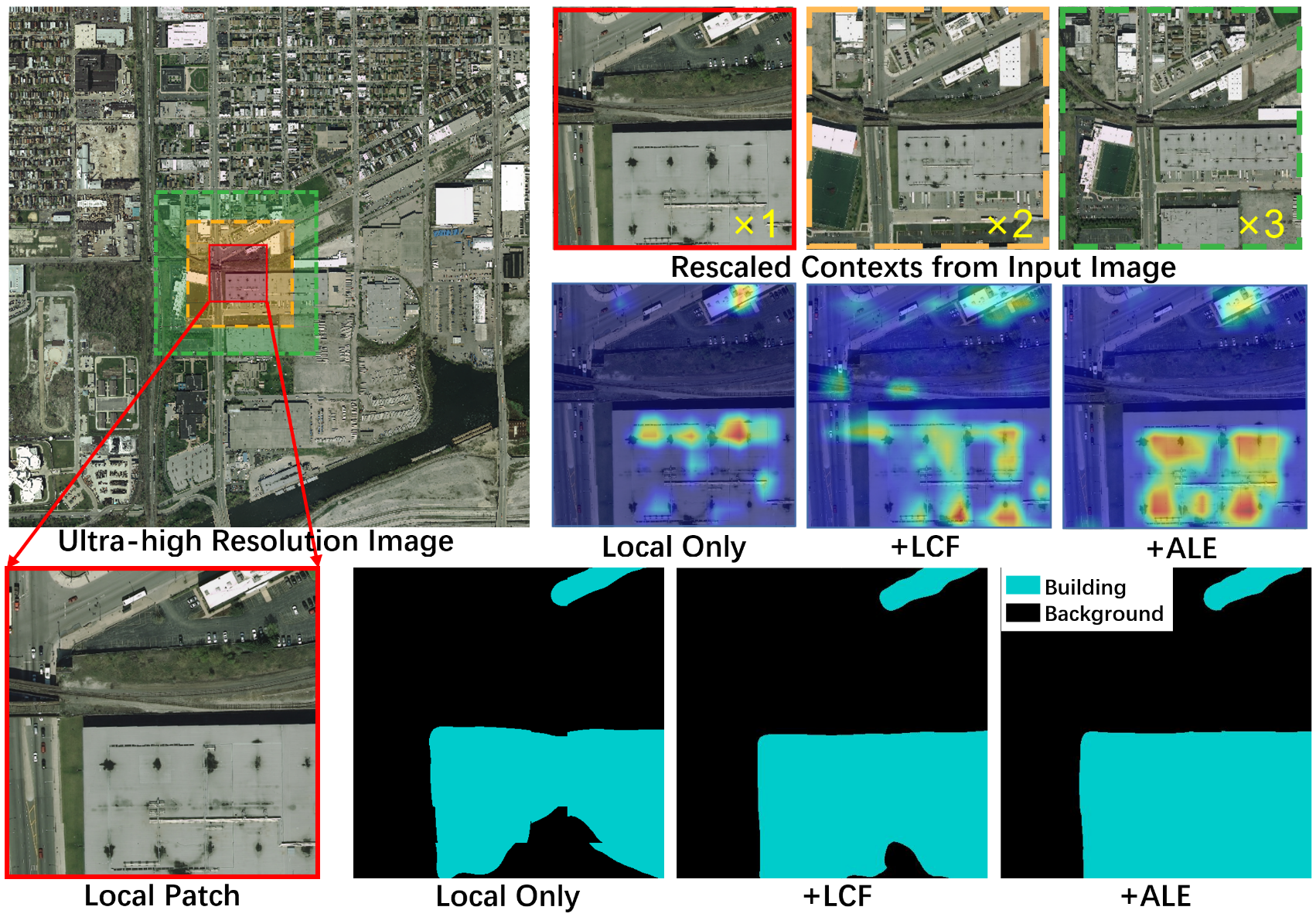}
    \caption{For the task of ultra-high resolution image segmentation, the most common approach is to segment cropped local patches and then combine them into a high-resolution mask. To address the core problem on local segmentation quality, we propose a locality-aware context fusion (LCF) based model that exploits rescaled various contexts ($\times 1, \times 2, \times 3$ large as local patches in the original image) \wxb{to enhance local features based on the complementarity of contexts. Moreover, we also present the alternating local enhancement module (ALE) to restrict the negative impact of the redundant information introduced from contexts and complement the locality-aware features to produce refined results. } }
    \label{fig:teaser}
\end{figure}

\section{Introduction}
\label{sec:introduction}

With the advance of photography and sensor technologies, the accessibility to ultra-high resolution images (i.e., 2K, 4K, or even higher resolution images) has opened new horizons to the computer vision community. It will benefits a wide range of imaging applications, e.g., urban planning and remote sensing based on high-resolution geospatial images and high-resolution medical image analysis, and thus the demand for studying and analyzing such images has urgently increased in recent years.
% Amongst the practical applications, segmentation of ultra-high resolution images plays important roles in a wide range of fields, such as urban planning and remote sensing [19, 20].

% [Difficulty of processing HD images]
In this paper, we aim at the specific task of semantic segmentation for ultra-high resolution geospatial images captured from aerial view. 
The recent development of deep convolutional neural networks (CNNs) has given rise to remarkable progress of semantic segmentation techniques. Yet, most CNN-based segmentation models target on full resolution images and perform pixel-level class prediction, which requires more computation resources comparing to image classification and object detection. This hurdle becomes significant when the image resolution grows to be ultra high, leading to the pressing dilemma between memory efficiency (even feasibility) and segmentation quality.

% To address the core problem on local segmentation quality, we propose a locality-aware contextual correlation based model that exploits rescaled various contexts ($\times 1, \times 2, \times 3$ large as local patch in the original image) to produce refined results. 

% [Existing methods and their limitations]
Particularly, in order to segment an ultra-high resolution image, the prevailing practice is either to downsample it to a smaller spatial dimension before performing segmentation, or to separately segment the partitioned patches and merge their results into a high-resolution one. These trivial practices sacrifice the segmentation quality for the model efficiency. 
Additionally, the recent attempts propose to utilize the well-pretrained segmentation models to obtain the coarse segmentation masks and another model to refine the contours of the masks \cite{cheng2020cascadepsp,zhou2020deepstrip}. 
However, these methods mainly focus on high-resolution natural images or daily photos concerning with large objects, while the high-resolution geospatial images are captured from aerial views covering a large field of view, which may contain many objects/regions with large contrast in scale and shape. Hence, it requires the segmentation model to be capable of capturing not only the semantics over large image regions but also the image details of different granularity. 
% For the former method, lots of image details will be lost during downsampling, which significantly degrades the segmentation quality. 
% As dealing with the ultra-high resolution images via image downsampling, a large amount of visual information will be discarded during the process, which thus degrades the segmentation quality. On the other hand, due to a lack of contextual information around each local patch, accumulating the overlapping patches may give rise to the artifacts on the boundaries of neighboring local patches. 
The recent work GLNet \cite{chen2019collaborative} proposes a two-stream network that separately processes the downsampled global image and cropped local patches, as well as a feature sharing module that shares the concatenated local and global features in both streams. Their method can achieve obvious improvements over existing methods, which embodies the importance of contextual information for segmentation performance.
% by capturing the local details under the guidance of the holistic information. 
Nevertheless, their feature sharing scheme does not spatially associate local features with the global ones and thus does not well exploit their correlation, which makes their model too complex to optimize and their performance suboptimal.
% , whereas it ignores the rich contextual information that can be explored in other scales. 

% [Proposed]
To thoroughly exploit the rich information within ultra-high resolution geospatial images, we present an
% Local-context-assisted High-Resolution Image Segmentation framework, named \textit{LoHRIS}, 
ultra-high resolution image segmentation paradigm featuring with \wxb{a locality-aware context fusion module and an alternating local enhancement module. }
% featuring with the locality-aware contextual correlation scheme \textcolor{blue}{featuring with a static scheme and a dynamic optimization}. 
% exploition of multi-scale contexts. 
Similar to \cite{ronneberger2015u,chen2019collaborative}, our framework is based on the widely used practice for high-resolution image segmentation, in which image patches are regularly cropped from the original image, then individually segmented, and finally their local results are overlayingly merged. 
However, each local patch of the ultra-high resolution geospatial images often contain semantic regions with large contrast in sizes (e.g., house and forest), which challenges the local segmentation model. 
Inspired by prior practices (e.g. \cite{chen2019collaborative}), contextual information turns out effective to resolve this problem.
But, unlike previous methods, we propose that the semantics within local patches can be structually and complementarily associated and inferred by their contextual regions of different scales. For instance, \wx{in Fig.~\ref{fig:teaser}, the contexts with varied coverage guide the model to the attentive regions relevant to the objects of different granularity in the image (e.g., small or large building).}
% small context can be more concentrated on small objects, while large context may be relevant with large objects due to its wide field of view.
% Unlike GLNet solely relying on the global context to communicate with local information, we assume that the semantics of local patch may be associated with contextual regions of different scales accordingly (see Fig. \ref{fig:teaser}). 
% It is inspired by the nature of the human's vision system, in which the field of view determines the levels of the perceived image details, while a human is able to adaptively adjust their field of view for better observation. 
% For instance, a closer fixation at an image may perceive more details (e.g., texture details or region boundaries), while a further look may observe more contextual information (e.g., the spatial relationship among regions or objects). During this process, human can freely adjust the perception field to capture the comprehensive information.

% [first contribution]

To accomplish this purpose, we first propose a locality-aware context fusion (LCF) module to exploit the correlation between local patch and its contextual regions. 
% \textcolor{blue}{Hence, we propose a deep network model based on a static scheme and a dynamic optimization, which exploits the correlation between local patch and its contextual regions and the interaction between local patch and its corresponding patch cropped from contextual regions, respectively.} 
% For instance, in Fig.~\ref{fig:teaser}, local patch and contexts can provide complementary cues to understand the scene. 
% As shown in the attention maps, the small building on the top-right can be identified in the smallest context (same as the local patch), in which the bottom-right large building is considered as background. On the other hand, larger contexts cover a wider field of view and thus can guide the model to attentively focus on the large building.
In concrete, the LCF is composed of two steps: locality-aware context correlation (LCC) and multi-context fusion (MCF).
LCC is designed to capture the positional relevance of local patch and contexts, which is able to attentively enhance the relevant features of the local patch, i.e., \textit{locality-aware features}. Then, an adaptive multi-context fusion (MCF) scheme is incorporated to balance and combine the locality-aware features associated by different contexts.
% which are captured from the ultra-high resolution image and rescaled to reduce the computation overhead.
\wx{As shown in Fig.~\ref{fig:teaser}, the contexts lead to different yet complementary locality-aware features, thus allow better tolerance towards misleading information brought by the context of a single scale.
To this end, the fusion weights of locality-aware features are predicted on-the-fly to accomplish the complementary fusion. }

\wx{However, computing locality-aware features concern with the contexts of different scales, so it inevitably brings in the redundant information from the regions outside local patch, and thus tends to generate undesired noises for semantic regions.
% with small contrast in sizes. 
% Moreover, downsampling context may cause missing critical details. 
To compensate the deficit, we propose a novel scheme, 
% the collaborative local representation optimization module (CLROM), 
denoted as the alternating local enhancement (ALE) module,
% the alternating optimization network 
to yield a cleaner local representation.
% , where the locally-cropped patches from contextual features are employed to refine local features. 
In practice, we crop and upscale the local features from the center of multiple contexts to avoid the interference of redundant information. Yet, upscaling cropped features may lead to the spatial misalignment during fusion and thus degrade the performance.
% To do so, we present a local feature interaction (LFI) block that enables to leverage reference patch features to enhance the target patch features.
% In order to make it not counterproductive in some cases, we introduce a dynamic optimization by the iterative interaction of local patch and cropped patch to restraint the negative influence of context. Noted that, the cropped patches contain a restricted context and is able to mutually confirm the content with local patch with details. In particular, we propose a local-context interaction module...
% Since LFI can be considered as an optimization function, 
To relieve this concern, we design a novel and effective network structure that recurrently utilizes the mutual relation of contexts to yield refined local representation. It will be further collaborated with LCF and used to suppress the negative impact of redundant contextual information. As shown in Fig.~\ref{fig:teaser}, LCF assists the understanding the main targets in the scene, whereas bring in irrelevant information. On the basis of LCF, ALE encourages the model to focus on relevant regions. 
} 

To evaluate our model, we conduct comprehensive experiments and demonstrate that our proposed model outperforms the state-of-the-art approaches on public ultra-high resolution arerial image datasets, DeepGlobe and Inria Aerial. The main contributions of our paper are summarized as below:
\begin{itemize}
    \item We present an ultra-high resolution image segmentation framework based on a novel local segmentation model featuring with two proposed modules, i.e., the locality-aware context fusion and \wxb{alternating local enhancement.}
    % collaborative local representation optimization modules. 
    \item We propose the locality-aware context fusion scheme accomplished via locality-aware contextual correlation and adaptive feature fusion, which associates and combines local-context information to strengthen local segmentation.
    % measures the relevances between local patches and multi-scale contexts to attentively enhance local features and adaptively fuse them in order to achieve the context-aware segmentation of local patches.
    % \item We present a contextual semantics refinement network that leverages the relevance of local segmentation and context mask to avoid boundary vanishing artifacts and refine the local semantic mask.
    \item To compensate the deficit caused by locality-aware features brought by redundant contexts, we introduce a novel \wxb{alternating local enhancement} scheme to yield local features to accompany locality-aware features.
    % \item To compensate the deficit caused by locality-aware features, we introduce a novel collaborative local representation optimization scheme to yield optimal local features. 
    \item Our method achieves the state-of-the-art semantic segmentation performance in several public ultra-high resolution geospatial image datasets.
\end{itemize}

\begin{figure}[t]
    \centering
    \footnotesize
    \begin{tabular}{c@{}c}
	\includegraphics[width=0.35\textwidth,height=3cm]
        {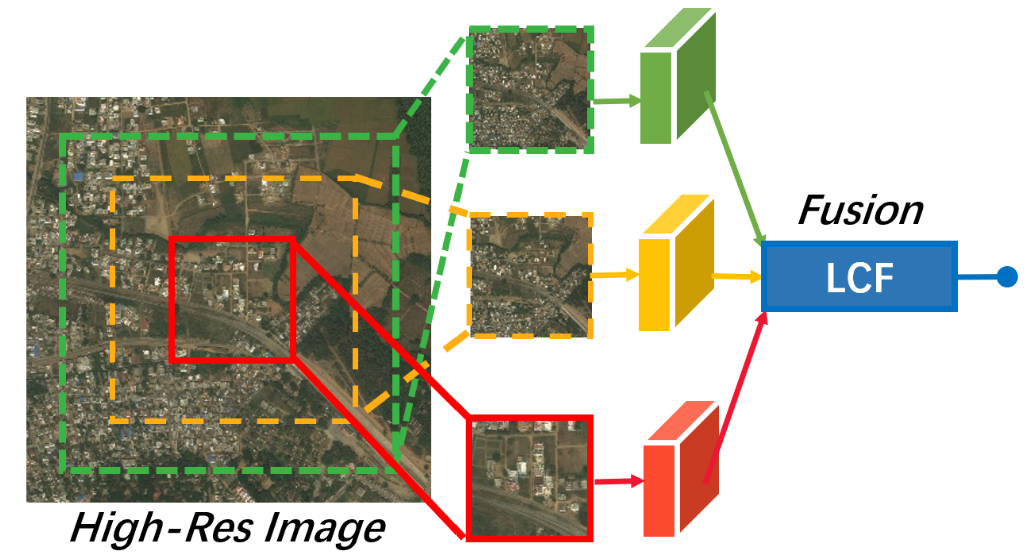} \\
        (a) Structure of LCC \cite{li2021contexts} \\
        \includegraphics[width=0.45\textwidth,height=3cm]
        {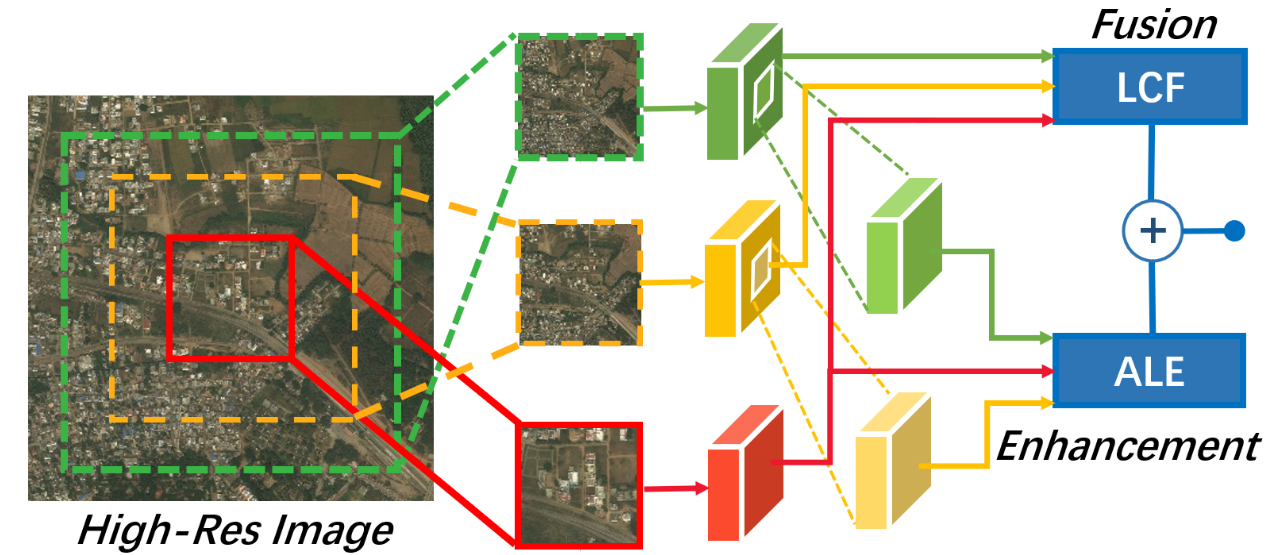} \\
        (b) Structure of our proposed model \\
	\end{tabular}
    \caption{Our proposed model versus LCC \cite{li2021contexts} (our preliminary version).}
    \label{fig:diff}
\end{figure}

Note that, a preliminary version of this work was presented as \cite{li2021contexts}. This submission extends \cite{li2021contexts} on methodology in the following aspects. First, we renovate our framework by simultaneously handling the contextual features in two different and complementary pipelines. We highlight the comparison between the structures of our proposed model and \cite{li2021contexts} in Fig.~\ref{fig:diff}. Specifically, we propose a novel alternating local enhancement module to address the limitations of the locality-aware features proposed in \cite{li2021contexts} caused by redundant contextual information. Under this scheme, we introduce a local feature interaction block and recurrently utilize it to yield optimal results from a specifically-designed network structure.
% the locality-confined attention sub-modules and we properly organize them to obtain the optimal results. 
Besides, we integrate the gating positional embedding technique to relieve the problem caused by position-agnostic spatial attention estimation.
Accordingly, we rephrase all the relevant sections, so as to better describe the motivation of ALE and its connection with LCF. 
Finally, we perform additional experiments to evaluate and discuss the effectiveness of our proposed model, and we show the state-of-the-art results produced by our model equipped with a latest vision transformer based network backbone. 

% In the remainder of this paper, we will survey the related works in Sec.~\ref{sec:related_works}. Then, we introduce the overall framework, and elaborate our proposed locality-aware context fusion module and alternating local enhancement in Sec.~\ref{sec:method}. Last, we demonstrate the experimental results in Sec.~\ref{sec:exp} and summarize our work in the Sec.~\ref{sec:conclusion}.

\section{Related Works}
\label{sec:related_works}
% \textcolor{cyan}{[more references and discussions]}
In this section, we will survey the relevant literature on semantic segmentation, context-guided models, and attention mechanism. 

% \noindent \textbf{Semantic Segmentation.}
\subsection{{Semantic Segmentation}}

In recent years, semantic segmentation has achieves remarkable progress \cite{long2015fully, 2018DeepLab, huang2019ccnet, wu2020cgnet, fu2019dual, he2019adaptive, luo2020multi,  kirillov2020pointrend, takikawa2019gated}. Fully convolutional network (FCN) \cite{long2015fully} was the first CNN architecture adopted for high-quality segmentation. U-Net \cite{ronneberger2015u} used skip-connections to concatenate low level features to high-level ones. Similar structures were also adopted by \cite{Noh_2015_ICCV, badrinarayanan2017segnet}. 
% DeepLab \cite{2018DeepLab} applied dilated convolutions to enlarge the field of view of filters. 
% Conditional random fields (CRF) were also utilized to model the spatial relationship. 
Unfortunately, these models suffer from prohibitively high GPU memory demand for ultra-high resolution images. ENet \cite{paszke2016enet} and ICNet \cite{zhao2018icnet} reduced GPU memory via model compression. However, these models were not effective on ultra-high resolution images. Recently, CascadePSP \cite{cheng2020cascadepsp} is proposed to refine the coarse segmentation results from a pretrained model to generate high-quality results. GLNet \cite{chen2019collaborative} preserves  both global and local information and interact each other through deeply shared layers, which is able to balance its performance and GPU memory usage. Compared with GLNet, the key difference rests in our proposed multi-context based local segmentation model, while GLNet relies on the holistic image as the only context and simply concatenates the local and cropped global features for segmentation. \wx{In addition, we also incorporate a novel and effective alternating local enhancement model to adaptively merge the local region from contexts.}

\subsection{Context-guided Models}
\wxb{In many image-based tasks, context plays a key role in encoding the local spatial neighborhood, or even non-local information~\cite{2016parsenet,zhao2017pyramid,2016ReSeg,2018DeepLab, wang2018non, chen2019collaborative, yang2020gated}. ParseNet \cite{2016parsenet} incorporated global pooling to aggregate different levels of contexts. DeepLab \cite{2018DeepLab} proposed dilated convolution and atrous spatial pyramid pooling module to aggregate global contexts into local information. In \cite{2018ContextNet,2018BiSeNet,2018Guided}, the deep/shallow branches were combined to aggregate global context and high-resolution details. GLNet \cite{chen2019collaborative} adopted the global context information to rich deep local feature by aggregating global branch and local branch. CPNet \cite{yu2020context} designed a context prior layer to embed the prior into the network and modelled the intra-class and inter-class dependencies. In recent works, ISNet \cite{jin2021isnet} proposed to integrate image-level and semantic-level contextual information to explore improving the pixel representations, and \wx{MCIBI} set up a feature memory module to store the dataset-level representations of various categories and mine contextual information beyond image. \wx{Unlike previous works, we tackle the contextual information from two aspects. First, we propose that the contexts of multiple scales can be spatially correlated and fused to enhance segmentation in a cross-attention manner. Moreover, a novel alternating local enhancement module is employed to suppress the inevitably introduced redundant information, so as to refine local features.} }

\begin{figure}[t]
    \centering
    \includegraphics[width=0.45\textwidth]{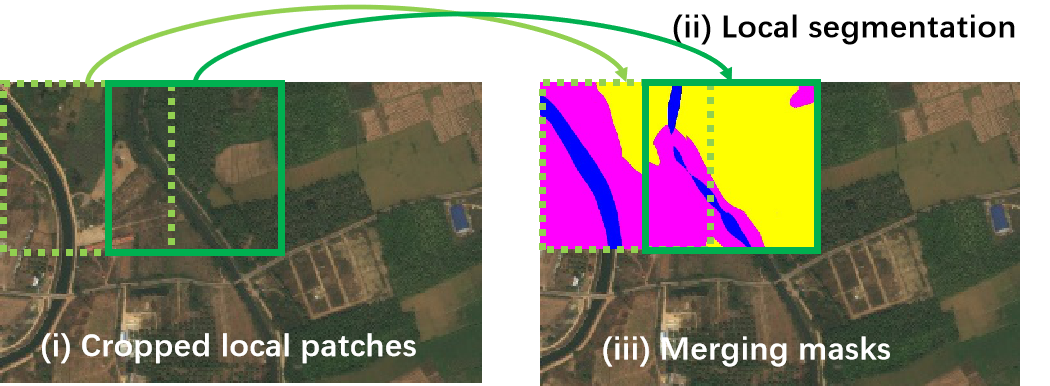}
    \caption{Main procedure of high-resolution image segmentation, consisting of (i) cropping local patches from the ultra-high resolution image; (ii) local patch segmentation; (iii) merging local masks into a high-resolution mask.}
    \label{fig:steps}
\end{figure}

\begin{figure*}
    \centering
    \includegraphics[width=0.95\textwidth]{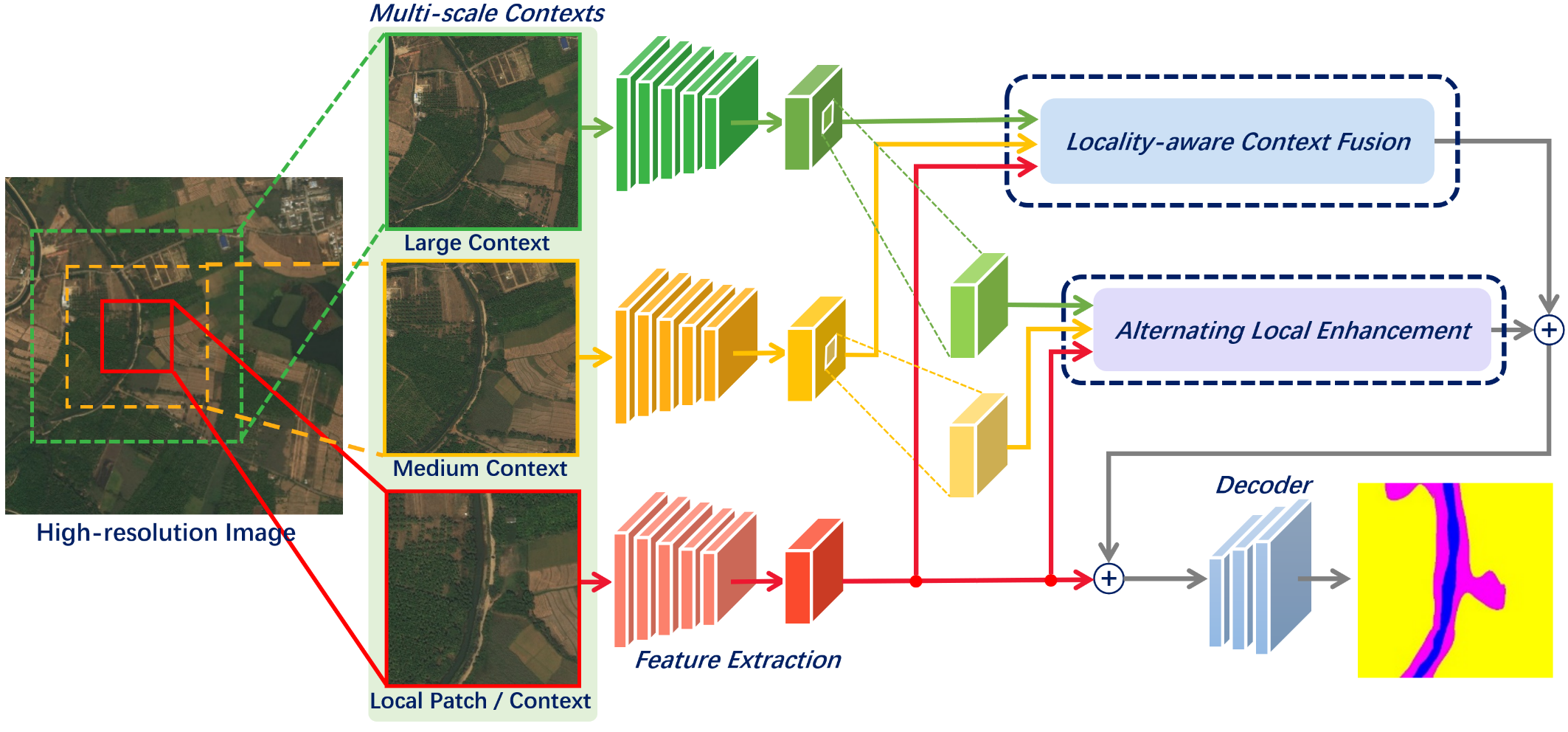}
    \caption{Illustration of our local segmentation model. In specific, a certain local patch cropped from the high-resolution image and its contexts are passed into the network branches separately to extract features and then measure their relevances against local patch to obtain the locality-aware features and adaptively fused via the \wxb{Locality-aware Context Fusion (LCF) module}. On the other hand, the local regions of  contextual features are upsampled and mutually optimized to yield robust local representation via the \wxb{Alternating Local Enhancement (ALE) module} for producing high-quality local segmentation results.}
    \label{fig:framework}
\end{figure*}

\subsection{Attention Mechanism}
\wxb{In recent years, it has witnessed the widespread application of attention mechanism in various fields \cite{guo2022attention}, including spatial attention, channel attention and self-attention mechanism. }

\noindent \textbf{Spatial Attention.} \wxb{Spatial attention is designed to adaptively select where needs attention in a spatial region. RAM \cite{mnih2014recurrent} first used recursive neural network and reinforcement learning for visual attention. Many subsequent works \cite{xu2015show, gregor2015draw} also have adopted RNN-based methods to recurrently predict important regions. STN \cite{jaderberg2015spatial} focused on the most relevant regions by learning invariance to rotation, scaling, translation and other more warps. Similar to STN, DCN \cite{dai2017deformable} adaptively enlarged the receptive field to pay attention to important regions by introducing the learnable offsets. GENet \cite{hu2018gather} combined part gathering and excitation to implicitly predict a soft mask to select important regions and capture long-range spatial information. PSANet \cite{zhao2018psanet} proposed a bidirectional information propagation
mechanism to aggregate global spatial information as an information flow and was added the end of CNN emphasizing the relevant features. \cite{huang2019ccnet, wang2018non, yin2020disentangled} all captured global spatial attention and strengthened the modeling ability of a regular CNN.}

\noindent \textbf{Channel Attention.} \wxb{Different channels tend to focus on different objects in different feature maps and channel attention autonomously adjusts the weight of each channel. SENet \cite{hu2019squeeze} used squeeze-and-excitation block to capture relationships between channels and
improve feature representation ability. Inspired by SENet, \cite{gao2019global, yang2020gated, qin2021fcanet, zhang2018context} improved squeeze 
or excitation module in various ways enhancing modeling capability Of network. Besides, some works \cite{woo2018cbam, fu2019dual, liu2020improving, wang2017residual} utilized both channel attention and spatial attention to capture feature dependencies in both domains.}

\noindent \textbf{Vision Transformer.} \wxb{In vision tasks, convolutional neural networks(CNNs) are considered the basic component \cite{he2016deep}, but nowadays
transformer \cite{vaswani2017attention} is showing it is a potential alternative to CNN. ViT \cite{dosovitskiy2021an} applied a pure transformer directly to sequences of image patches to classify the full image and achieves state-of-the-art performance. In addition to image classification \cite{liu2021swin, han2021transformer}, transformer has been utilized to address a variety of other vision problems, including object detection \cite{carion2020end, zhu2021deformable}, semantic segmentation \cite{zheng2021rethinking, xie2021segformer}, pose estimation \cite{lin2021end}, image generation \cite{jiang2021transgan}, image enhancement \cite{chen2021pre}. Transformer adopts self-attention mechanism, which lacks the ability to capture the positional information. In order to solve the problem, a common design is to first represent positional information using vectors and then infuse the vectors to the model as an additional input. In \cite{vaswani2017attention}, positional information was encoded as absolute sinusoidal position encodings. Later ways \cite{kenton2019bert, gehring2017convolutional} of representing absolute positions were to learn a set of positional embeddings for each position. Compared to hand-crafted position representation, learned embeddings are more flexible in that position representation can adapt to tasks through back-propagation. Another line of works \cite{shaw2018self, gu2019insertion, ramachandran2019stand} focused on representing positional relationships between tokens instead of positions of individual tokens. Methods following this principles are called relative positional representation. \cite{shaw2018self} proposed to add a learnable relative position embedding to keys of attention mechanism. ConViT \cite{d2021convit} used positional embeddings of the relative position of patches and introduces a gating parameter. Besides, Some research studies \cite{su2021roformer} have explored using hybrid positional representations that contains both absolute and relative positional information. TUPE \cite{ke2021rethinking} redesigned the computation of attention scores as a combination of a content-to-content term, an absolute position-to-position term and a bias term representing relative positional relationships. In addition to embedding at explicit positions in the encoder, some work study position representations without explicit encoding \cite{Wang2020Encoding} and position representation on decoders \cite{tsai2019transformer}.}

\wx{In our model, we also employ attention scheme in LCF to correlate contexts and local information, while we apply a spatial attention with different window kernels via depthwise convolution in ALE.
% gated relative positioal embedding in gaining position information to enhance the representation of locality-aware features in LCC while we utilize spatial attention with multi-window kernels via depthwise convolution to perform mutual optimization for local features in LFI.
}

\section{Our Proposed Methodology}
\label{sec:method}

\subsection{Overview}

Our proposed ultra-high resolution image segmentation framework follows the three-step procedure as shown in Fig. \ref{fig:steps}, which is consistent with the common practice applied in prior works (e.g., \cite{ronneberger2015u,chen2019collaborative}). 
% We first sample or crop several local patches from the input image, then compute the segmentation result for each patch, and combine the local segmentation into one piece as the final result. 
First, given an ultra-high resolution image $\dot{I}$ with width $W$ and height $H$, we evenly partition it into $N$ local patches $\{I_{k}\}$ ($k=[1,\cdots,N]$, $I_{k} \subset \dot{I}$) with width $w$ and height $h$ ($w<W$ and $h<H$) \lqq{along the row and column axis}. 
% In practice, the adjacent patches are partially overlapped. 
Next, a local semantic segmentation model computes the local result for each patch. Last, we merge the local results into one piece as the final high-resolution segmentation mask. 
% In this step, we replace the trivial merging approach with a semantics refinement network to refine the local masks.
% i.e., $\dot{M} = \mathcal{F}_{merge}(M_1,\cdots,M_N)$. 
Our main contributions rest in how to generate fine local segmentation. As follows, we will elaborate the technical details. 

% $\{\{I_{ij}\}_{j=1}^{n}\}_{i=1}^{N}$ with N scales and the different scales of each patch have the same center point. The segmentation branch receives $I_{1j}$ and the contextual branch receives $I_{ij},i \in [1,...,N]$ for patch $j$(see example). Note that $I$ is fully cropped into patches(instead of random cropping)to facilitate both training and inference. $I_{ij} \in \mathbb{R}^{h \times w}$, where $h \ll H$, and $w \ll W$. We adopt the same backbone for two branches and xxx Module acts on deep feature.

% \subsection{Our Proposed Local Segmentation Model}

As the core of our ultra-high resolution segmentation framework, we propose a novel local segmentation model to process each cropped patch (Fig.~\ref{fig:framework}).
% It is obvious that the quality of local patch segmentation affects the final segmentation result of the ultra-high resolution image. 
Yet, each local patch only covers a confined field of the ultra-high resolution image, which often contains regions of varied scales or truncated objects, and thus it tends to deliver incomplete information and may easily cause erroneous semantic segmentation. 
% it is difficult to infer the contextual and holistic information on image content. 
% Specifically, the object of interest (e.g. building) may be truncated, and thus confuse the local segmentation model for inferring its semantic meaning. 
% Inspired by prior practices \cite{chen2019collaborative}, involving contextual information relevant to the local patch during inference can be effective to mitigate the problems. 
To address this concern, we propose a locality-aware contextual correlation based segmentation model for processing each local patch.
% Hence, we propose a context-aware local segmentation model that can well exploit the image context without increasing too much computation burden. 
% to introduce the contextual information of a local patch to enhance the local segmentation. 

As illustrated in Fig. \ref{fig:framework}, our local segmentation model is based on a multi-stream encoder-decoder architecture, consisting of the feature extraction modules (i.e., encoder), locality-aware context fusion module (LCF), alternating local enhancement module (ALE), and decoder. In specific, an image patch and its context regions of different scales, which are rescaled into the same size for reducing computation overhead, are fed into the network for feature extraction. Then, the features of contexts are separately associated with the features of local patch via two proposed modules, LCF and ALE, to strengthen the local features. In final, the enhanced local features will be upsampled and decoded to obtain the local segmentation mask. 
As follows, we will first introduce how to choose the context of local patch, and then describe the proposed modules, respectively. 

% Hence, our local segmentation model mainly consists of two components: local-context relevance module and multi-scale context fusion scheme.
% As follows, we will first define the context of the local patch and then introduce our proposed local-context relevance module to attentively strengthen the visual features of local patches. Last, we present how to adaptively fuse multi-scale context to improve the local segmentation.

\begin{figure*}[h!]
    \centering
    \includegraphics[width=0.9\textwidth]{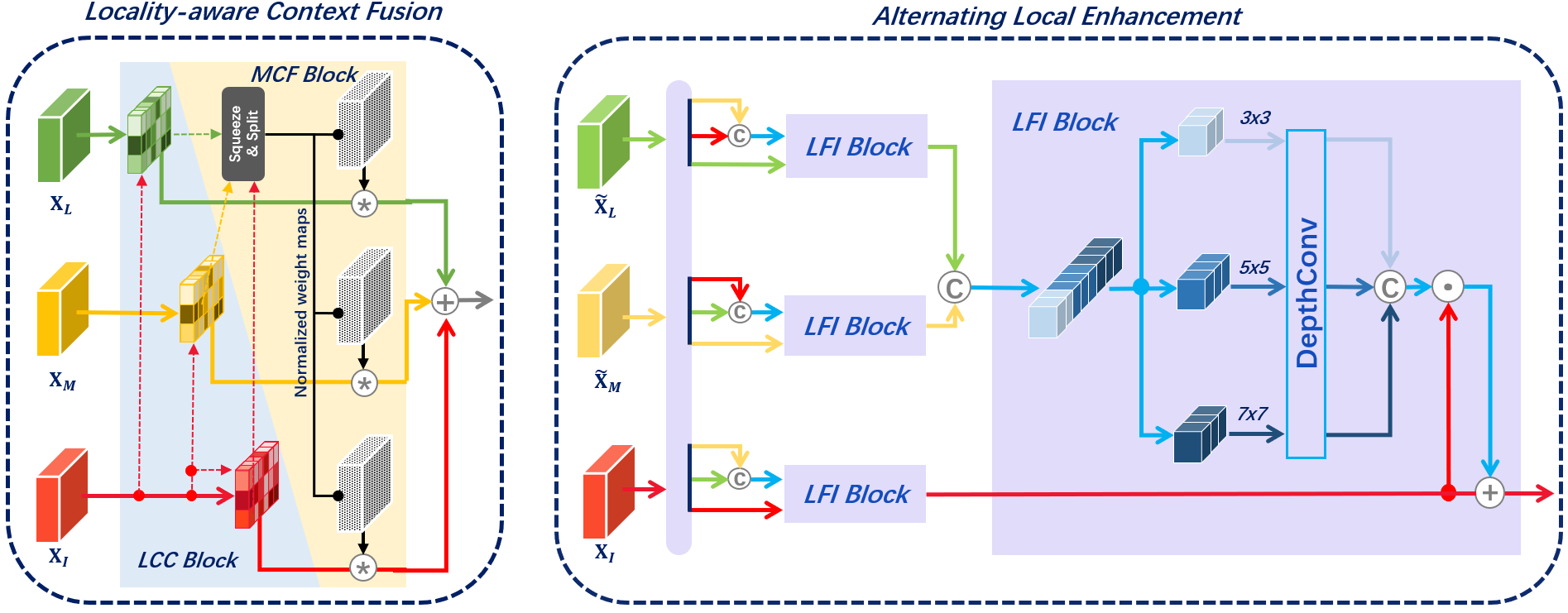}
    \caption{\wx{We illustrate the detailed structures of LCF (left) and ALE (right). As the main module for adaptively fusing contextual information, LCF is composed of locality-aware contextual correlation (LCC) block and multi-context fusion (MCF) block. In particular, it first correlates the contextual and local features, and then fuses them in a complementary manner. 
    On the right, ALE attempts to alleviate the erroneous information of LCF. ALE recurrently utilizes the Local Feature Interaction (LFI) blocks to optimize the contextual features and eventually yield the clean local representation. Specifically, LFI is based on the spatial attention scheme with different window kernels, which can effectively leverage the reference features of different scales to enhance the target ones.
    % In our segmentation model, we propose the locality-aware context fusion module(left) which mainly consists of locality-aware contextual correlation (LCC) block and multi-context fusion (MCF) block to enhance local features. To prevent from bringing in redundant information from locality-aware context fusion, we also present the alternating local enhancement module(right) that contains several Local Feature Interaction (LFI) blocks to complement locality-aware features.
    }}
    \label{fig:local_framework}
\end{figure*}

\subsection{Context of Local Patch} 
\label{sec:choose_context}

% Assume that the entire ultra-high resolution image $\dot{I}$ is equally divided into $K$ patches. 
Amongst all $N$ patches, regarding the $k$-th patch $I_k$, another image patch $U_k$ refers to the region within $\dot{I}$, which is not smaller than and covers $I_k$. $U_k$ has the width $w_u$ and height $h_u$, s.t. $w \leq w_u \leq W \text{ and } h \leq h_u \leq H$. 

Given a local patch, there are many candidate context regions. In practice, we design the following three types of context regions. (1) We set the size of the candidate context subject to $w_u = \lambda w \text{ and } h_u = \lambda h, (\lambda \geq 1, w_u \leq W, h_u \leq H)$, and its center aligned with the center of the local patch (see the examples in Figs.~\ref{fig:teaser} and ~\ref{fig:framework}). (2) The largest context we can utilize is exactly the whole image, i.e., $U_k \equiv \dot{I}$, dubbed \textit{global context}. (3) The smallest context is the patch itself, dubbed \textit{local context}, i.e., $w_u \equiv w \text{ and } h_u \equiv h$. 
Generally, the larger contexts offer more contextual cues that may be responsible for large regions or objects, while the smaller ones provide more details that may be attributed to small regions or objects. Before feeding the contexts along with local patch into network, they will be rescaled as the same dimension as the local patch in order to reduce the computation overhead.

% For the scale of the context region, there are several cases to consider: (1) $w_u \equiv w, h_u \equiv h$: The smallest context region is the local patch itself. In this case, measuring local-context relevance is essentially the evaluation of self-correlation, which is effective since certain regions within the patch may help identify other regions. (2) The context region matches the entire image, i.e., $U_k \equiv \dot{I}$: In this case, the entire image can provide the holistic information. (3) $w < w_u < W, h < h_u < H$: 
% Most selected context regions are larger than the local patch while smaller than the entire image. 
% Generally, the larger contexts offer more global cues, while smaller ones provide more details. 
% In practice, we let the center of the selected context regions to be aligned with the center of the local patch.
% Additionally, we choose the contexts with the identical aspect ratio with the local patch and normalize them into the same dimension as the local patch before feeding them into the network. 

% \input{lcf}

\subsection{Locality-aware Context Fusion}

To strengthen the segmentation of a local patch, we would like to associate the contextual information with the local information. Thus, as illustrated in Fig.~\ref{fig:framework}, apart from any target local patch, we involve two different contexts, the medium context and the large context for estimating the local semantic mask. 
For the sake of simplicity in description, in the following sections, we denote the target local patch as $I$, and their medium and large contexts as $M$ and $L$, respectively.
In the framework, we propose the LCF module $\mathcal{F}_{LCF}$ to evaluate the relevance of the target patch and its contexts, as well as its self-correlation, which are then employed to obtain the locality-aware features of the target patch. 

Concretely, the structure of our produced LCF module is specified in the left figure of Fig.~\ref{fig:local_framework}. It mainly consists of two steps: locality-aware contextual correlation (LCC) and multi-context fusion (MCF). Specifically, LCC discovers the relevance of local patch and contexts to enhance local features, and MCF then adaptively merges them so the contexts can well complement the local features.

\subsubsection{Locality-aware Contextual Correlation}
% \textcolor{red}{[Motivation and aim of LCC block]}
% In LCC block, we aim to correlate the contexts and local patches. 
\wx{Inspired by the recent development in vision transformers~\cite{dosovitskiy2021an}, we want to capture the positional relevance of local patch and contexts to enhance local patch. To accomplish this purpose, we propose the following procedure.}
First, the features of the local patch $I$ and its medium and large contexts, $M$ and $L$, are separately extracted through the identical yet independent backbone network structures (e.g., Conv1 to Conv3 of the pretrained VGG16~\cite{Simonyan15} \wx{or pretrained Segformer~\cite{xie2021segformer}).}
% For the sake of simplicity in description, we omit the subscript $k$ below. 
The extracted features can be denoted as $\textbf{X}_I$, $\textbf{X}_M$, and $\textbf{X}_L$ ($\textbf{X}_I,\textbf{X}_M, \textbf{X}_L\in \mathbb{R}^{c \times h_x \times w_x}$). 

Next, the relevance of $I$ and its context can be calculated via the outer product of the local features $\textbf{X}_I$ and its context features 
$\textbf{X}_U$ ($U \in \{I, M, L\}$), i.e., $\textbf{R}_U = \textbf{X}_I \otimes \textbf{X}_U$, which measures the non-local correlation by establishing the pairwise pixel-level relation for $\textbf{X}_I$ and $\textbf{X}_U$, as illustrated in the left figure of Fig.~\ref{fig:local_framework}.
Hence, the relevance can be further applied as an attention map to enhance the local features, $\textbf{X}_I$, which attends to the semantic regions more relevant with the locality. 
% Specifically, $\textbf{R}_k$ is passed through a softmax layer to obtain the attention map and then perform element-wise multiplication with $\textbf{X}_k^i$, i.e., $\overline{\textbf{X}_k} = \text{Softmax}(\textbf{R})_k \odot \textbf{X}_k^i$.
Specifically, $\textbf{R}_U$ is passed through a softmax layer to obtain its corresponding attention map as below: 
\begin{align}
    \textbf{M}_U=\text{Softmax}(\textbf{R}_U) = \text{Softmax}(\textbf{X}_I \otimes \textbf{X}_U), U \in \{I,M,L\}.
    \label{eq:m_k1}
\end{align}
% Then, we perform inner product with $\textbf{X}_k^U$, i.e., $\overline{\textbf{X}}_k = \langle M_k , \textbf{X}_k^U\rangle$.
% Last, $\overline{\textbf{X}}_k^i$ will be utilized to achieve the segmentation result $M_k$. 
% For the sake of clarity, we denote the entire procedure as $\overline{\textbf{X}}_k = \mathcal{F}_{lcc}(I_k, U_k)$.

% Overall, the inference process of the $k$-th local patch segmentation with regards to its certain context region can be expressed as below:
% \begin{align}
%     M_k &= \mathcal{F}_{seg}(I_k,U_k),\nonumber\\
%     &= F_{up}(F_{lcr}(F_{fe}(I_k), F_{fe}(U_k))).
% \end{align}

% To  enhance local branch deep feature, we introduce a xxx module. The xxx module enriches local feature with the wider range of contextual information  by using contextual branch deep feature. As illustrated in Figure, we first feed a local patch and a context separately to the coding network to obtain local deep feature  $L \in \mathbb{R}^{C \times H \times W}$ and contextual deep feature $C \in \mathbb{R}^{C \times H \times W}$.

% \noindent \textbf{Relative positional embeddings.}
% \textcolor{blue}{Specifically, $\textbf{R}_k$ is passed through a softmax layer to obtain the attention map $\overline{\textbf{R}}_k=\text{Softmax}(\textbf{R}_k)$.}
% (Why we add this?)
However, according to Eq. \ref{eq:m_k1}, $\textbf{M}_U$ is essentially position-agnostic. It contain insufficient information on how local features $\textbf{X}_I$ and contextual features $\textbf{X}_U$ are spatially correlated, which results in permutation equivariant and thus less expressive for our task. 
To incorporate their position information, one way is to add sinusoidal embeddings based on the absolute position of pixels, but early experimentation suggests that the employment of relative positional embeddings usually results in significantly better accuracies \cite{shaw2018self, ramachandran2019stand}.
% Another possibility is to model the spatial position between pixels with relative positional embeddings.
\lqq{
Therefore, inspired by \cite{d2021convit},
% which [does not ignore the smaller values between in the correlation and positional information to express their relative importance by learnable gating parameters], 
% we introduce the gated positional embeddings scheme that adaptively delivers the relative positional information to $\textbf{M}$ via the embeddings $\textbf{r}_{ij}$ of the relative positions of $i$-th and $j$-th patches. To adaptively combine the feature correlation and positional embeddings, the derivation of the attention map $M$ can be expressed as below:
% \begin{align}
%     \textbf{M}_U = \text{Norm}( &\sigma(z)\text{Softmax}(\textbf{X}_I \otimes \textbf{X}_U) \nonumber \\ 
%     &+  (1-\sigma(z))\text{Softmax}(\textbf{v}^{T}_{pos}\textbf{r}_{ij})),
%     % Gate = (1-\sigma(\lambda))\text{softmax}(\langle\textbf{X}_k^i,\textbf{X}_k^U \rangle)  \nonumber \\
%     %  + \sigma(\lambda)\text{softmax}(\textbf{v}^{T}_{pos}\textbf{r}_{ij}),
% \end{align}
% where $\textbf{v}_{pos}$ ($\textbf{v}_{pos} \in \mathbb{R}^{D_{pos}}$) represents a trainable embedding, and $\textbf{r}_{ij}$ ($\textbf{r}_{ij} \in \mathbb{R}^{D_{pos}}$) denotes the relative positional embeddings computed by the distance of the $i$-th and $j$-th pixel. $\text{Norm}(\cdot)$ normalizes the correlation of the $i$-th pixel and all pixels. Note that, the gating parameter $z$ is trainable and $\sigma(\cdot)$ refers to the sigmoid function, which can dynamically balance the contributions of feature correlation and positional embeddings.
we employ the gated positional embeddings scheme that adaptively delivers the relative positional information to $\textbf{M}$ via the embeddings $\textbf{r}$ of the relative positions of every two patches. To adaptively combine the feature correlation and positional embeddings, the derivation of the attention map $\textbf{M}$ can be expressed as below:
\begin{align}
    \textbf{M}_U = \text{Norm}( &\sigma(z)\text{Softmax}(\textbf{X}_I \otimes \textbf{X}_U) \nonumber \\ 
    &+  (1-\sigma(z))\text{Softmax}(\textbf{r} \otimes \textbf{W}_r)),
    % Gate = (1-\sigma(\lambda))\text{softmax}(\langle\textbf{X}_k^i,\textbf{X}_k^U \rangle)  \nonumber \\
    %  + \sigma(\lambda)\text{softmax}(\textbf{v}^{T}_{pos}\textbf{r}_{ij}),
\end{align}
where $\textbf{W}_r$ ($\textbf{W}_r \in \mathbb{R}^{3 \times 1}$) represents a trainable embedding, and $\textbf{r}$ ($\textbf{r} \in \mathbb{R}^{h_xw_x \times h_xw_x \times 3}$) denotes the relative positional embeddings computed by the distance of every two pixels. $\text{Norm}(\cdot)$ normalizes the correlation of the $i$-th pixel and all pixels. Note that, the gating parameter $z$ is trainable and $\sigma(\cdot)$ refers to the sigmoid function, which can dynamically balance the contributions of feature correlation and positional embeddings.}
Last, we use the attention map $\textbf{M}_U$ to enhance $\textbf{X}_I$ and thus yield the locality-aware features, i.e., $\overline{\textbf{X}}_{I\vert U} =  \textbf{M}_U \otimes \textbf{X}_U$.

\subsubsection{Multi-Context Fusion}

% Analogy to the vision system of human being, the contexts of different scales may offer various types of fine-grained details to reason about the content of the local patch. 
% Inspired by the human vision system, the contexts of different scales may provide complementary information to reason the content of local patches. 
For the ultra-high resolution geospatial images that often contain a large number of objects with large size variations, the contexts of different scales may be attributed to the segmentation of the objects with various granularity. 
% But, geospatial images often contain a large number of objects with large contrast in their sizes. 
Therefore, properly combining different contextual information can be complementary for extracting semantics and removing artifacts. 

In particular, given the local patch $I$, 
% we have several corresponding context regions, $U_t$ ($t = [1, \cdots, T]$) and pass them into each stream of our local segmentation model to obtain locality-ware features $\overline{\textbf{X}}^{U_t}$. 
we pass the local patch $I$, the medium context $M$, and the large context $L$ into the network streams of our local segmentation model (see Fig.~\ref{fig:framework}).
Through feature extraction and LCF modules, we can obtain locality-ware features $\overline{\textbf{X}}_{I\vert I}$, $\overline{\textbf{X}}_{I\vert M}$, and $\overline{\textbf{X}}_{I\vert L}$.
% to facilitate the segmentation for each patch, 
% Specifically, we assume that there are $T$ corresponding context regions that may affect the local segmentation. As shown in the left figure of Fig.~\ref{fig:local_framework}, we have three types of extracted features from large context, medium context, and local context (or local patch) in practice. 
% The routine we choose context regions follows the formulation in Sec.~\ref{sec:choose_context}.
% In our practice, $T$ is equal to $3$, and there are three contextual features including local features, medium context features, and large context features. 
% ($\overline{\textbf{X}}_k^t = \mathcal{F}_{LCC}(I_k, U^t_k)$).
% , each of which has a unique dimension, i.e., $(w^{ti}_u, h^{ti}_u) \neq (w^{tj}_u, h^{tj}_u), \forall ti \neq tj$. 
To effectively exploit and combine the locality-aware features that stem from various contexts, as shown in left part of Fig.~\ref{fig:framework}, we integrate a multi-context fusion (MCF) block into our local segmentation model.
% , which adaptively estimates the spatial weight maps for fusing the contextual information in a balanced way. 
% To accomplish this, we first calculate the features enhanced by contexts, i.e., $\overline{\textbf{X}}_k^t = \mathcal{F}_{lcr}(I_k, U_k^t)$, respectively, via the local-context relevance module. 
% a novel network module $\mathcal{F}_{est}$ estimates the weight maps $\{\textbf{H}^t\}$ corresponded to the features $\{\overline{\textbf{X}}^t\}$, respectively.
Specifically, all the locality-aware features are first concatenated and passed through a \textit{squeeze-and-split} structure, which compresses and entangles the multi-scale features before predicting the normalized weights for the locality-aware features. The squeeze-and-split structure
squeezes the concated features via a convolutional layer with the kernel size $1 \times 1$, which blends the locality-aware features. Then, the squeezed features are reconstructed to the original dimension via another $1 \times 1$ convolutional layer and passed through a softmax to obtain three normalized weight maps $\{\textbf{H}_U\}$ ($\textbf{H}_U \in \mathbb{R}^{c \times h_x \times w_x}$, $U \in \{I,M,L\}$) which balance the contributions of each term. Hence, the entire procedure is:
\begin{align}
\{\textbf{H}_I,\textbf{H}_M,\textbf{H}_L\} = \mathcal{F}_{est}(\text{Concat}(\overline{\textbf{X}}_{I\vert I},\overline{\textbf{X}}_{I\vert M},\overline{\textbf{X}}_{I\vert L})),
\end{align}
where $\mathcal{F}_{est}$ refers to the network module for predicting the weights of locality-aware features. 
% \wx{[*** describe the network structure***] 
    % $\mathcal{F}_{est}$ consists of ..., which is appended with a softmax layer to normalize the attention map. }
Formally, complementing the locality-aware features from varied contexts can be expressed below:
% \begin{align}
%     {\overline{\textbf{X}}} &= \sum_{t=1}^T \textbf{H}^t \odot \overline{\textbf{X}}^{U_t}, \quad s.t.  \sum_{t=1}^T \textbf{H}^t = \textbf{1}
% \end{align}
\begin{align}
    {\overline{\textbf{X}}_I} &= \textbf{H}_I \odot \overline{\textbf{X}}_{I\vert I} + \textbf{H}_M \odot \overline{\textbf{X}}_{I\vert M} + \textbf{H}_L \odot \overline{\textbf{X}}_{I\vert L}, \nonumber \\
    & s.t.  \quad \textbf{H}_I+\textbf{H}_M+\textbf{H}_L = \textbf{1},
\end{align}
where $\odot$ refers to the element-wise multiplication and $\textbf{1}$ represents the matrix where all elements are normalized to be 1. 
% Note that, the elements of three weight maps along each channel are summed to be 1. 
% Last, the fused features will be joined with the features of the local patch via a skip connection to form a residual structure, and then used to compute the segmentation mask of the patch via several upsampling layers in the decoder.
In this way, our feature fusion scheme is able to take advantage of the complementary information from different contexts. 

In overall, the entire process of LCF module can be summarized as below:
\begin{align}
    \overline{\textbf{X}}_I = \mathcal{F}_{LCF}(\textbf{X}_I, \textbf{X}_M, \textbf{X}_L).\label{eq:lcf}
\end{align}

% the loss function of our local segmentation model is: 
% \begin{align}
%     \mathcal{L}_{seg} = - y(1-p)^\gamma log(p) - (1-y)p^\gamma log(1-p)
% \end{align}

\begin{figure}[t]
    \centering
    \footnotesize
	\begin{tabular}{c@{}c@{}c}
	\includegraphics[width=0.14\textwidth,height=2.3cm]
        {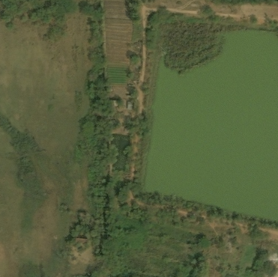}&
        \includegraphics[width=0.14\textwidth,height=2.3cm]
        {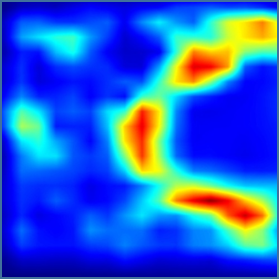}&
		\includegraphics[width=0.14\textwidth,height=2.3cm]
        {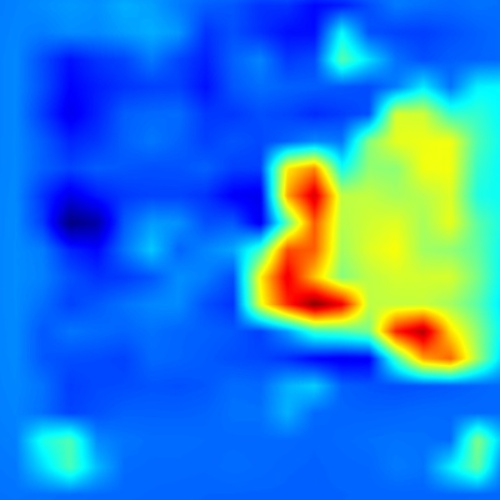}\\
        Local Patch & Before ALE & After ALE \\
        \includegraphics[width=0.14\textwidth,height=2.3cm]
        {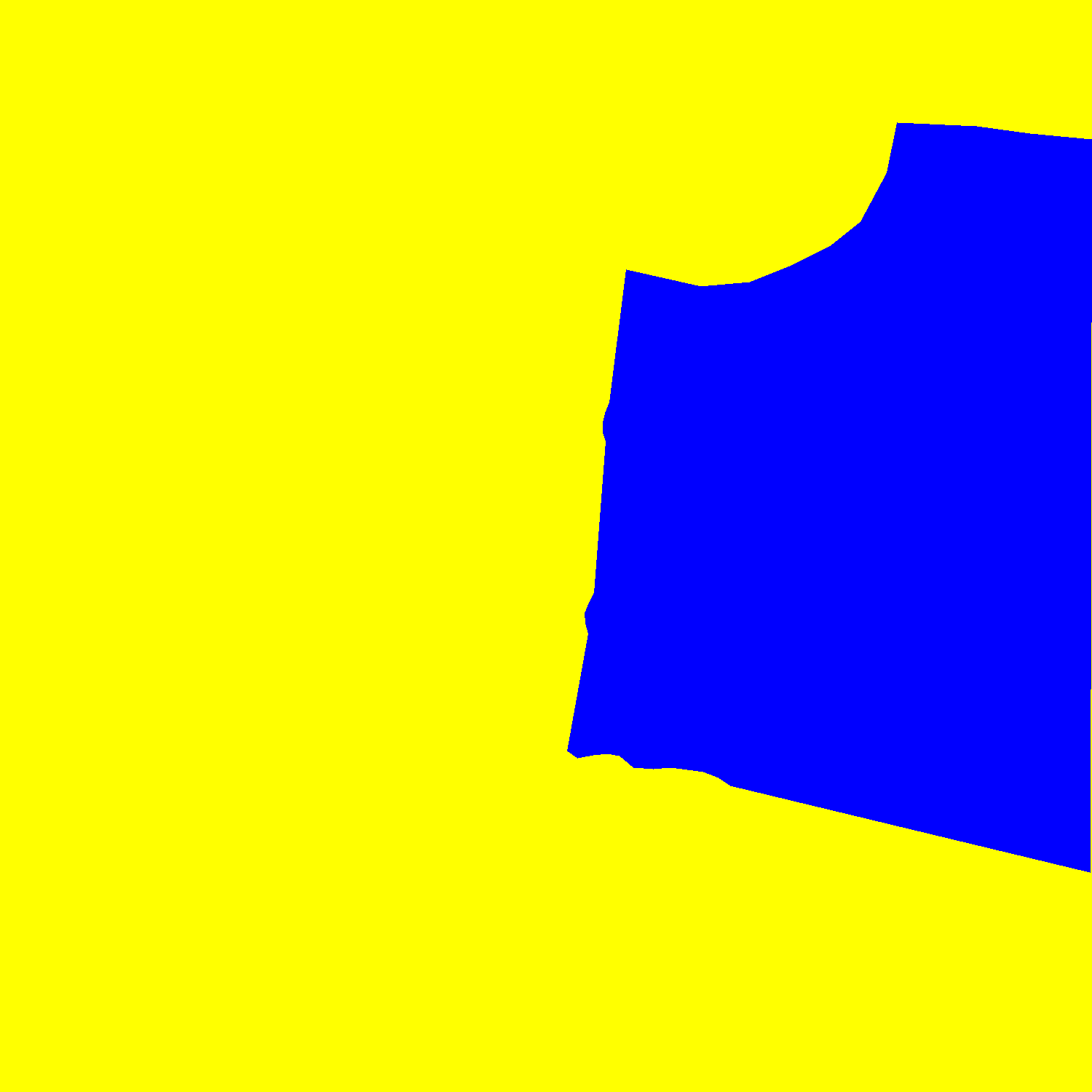}&
		\includegraphics[width=0.14\textwidth,height=2.3cm]
        {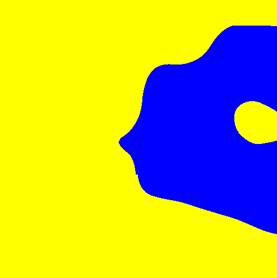}&
        \includegraphics[width=0.14\textwidth,height=2.3cm]
        {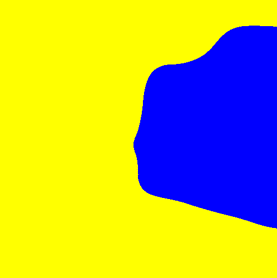}\\
        Ground Truth & w/o ALE & w/ ALE\\
	\end{tabular}
	\caption{\wxb{We illustrate a simple example that 
	the locality-aware features only may cause errors due to redundant contextual information. Concretely, the images on the first row imply that the locality-aware features (i.e., "Before ALE") falsely focus the boundary of ROI, whereas ALE manages to fix the error and produce better results. }}
	\label{fig:ALE}
\end{figure}

% \subsection{Collaborative Local Representation Optimization}

\subsection{Alternating Local Enhancement}

As mentioned above, our proposed LCF is able to utilize the contextual information to complement the local features, but its non-local correlation scheme inevitably introduces redundant information, especially from the region outside the local patch, and thus results in erroneous estimation. 
% \wx{Besides, LCF essentially performs the fusion of the contextual features, in which the contextual features are employed as static features in the training process. Thus, it results in that the common erroneous estimation from all contexts can hardly be fixed.}
\wxb{
% Yet, LCF essentially performs the fusion of the contextual features, in which the contextual features are treated as static features. 
% Thus, it results in erroneous estimation due to fusion features affected by redundant information. 
We show a simple example illustrating it in Fig.~\ref{fig:ALE}, where the estimation of the water region is misled by complex contexts so that the locality-aware features (i.e., the middle figure on the top row of Fig.~\ref{fig:ALE}) falsely focus on the boundary of the region.}

% the contextual information is not fully exploited to infer the local semantics. 
\wxb{To remedy the shortcomings, one straightforward solution is to merge the local features from different contexts while excluding the features outside the region-of-interest. In practice, as shown in Fig.~\ref{fig:framework}, the center regions of the context features that correspond to the local patch are cropped and upsampled into the same dimension before fusing them. However, upsampling features may cause spatial misalignment and thus degrade the performance. 
Therefore, we propose the alternating local enhancement (ALE) module, in which 
% As shown in Fig.~\ref{fig:framework}, the center regions of the context features that correspond to the local patch are cropped and upsampled into the same dimension as the local features before being delivered into ALE for feature enhancement. In ALE, 
the features from different contexts will be utilized to optimize each other, so as to leverage their correlation to alleviate the negative impact of redundant contexts and compensate the limitations of LCF.
We will elaborate the ALE in the following. } 
% To remedy the shortcomings, we propose the collaborative local representation optimization module (CLROM). We elaborate the CLROM in the following.  

\wxr{\subsubsection{Alternating Enhancement for Local Features}}

% In practice, 
To prevent from bringing in redundant information outside the local region of interest, we need to concentrate on the common region from local patch and contexts.
% First, to avoid the redundant information in the non-local regions, we crop the local area from the contextual features and upsample it to align with the features of local patch. 
As show in Fig.~\ref{fig:framework}, we have the features of the local patch $\textbf{X}_I$, as well as the features of the medium and large contexts ${\textbf{X}}_{M}$ and ${\textbf{X}}_{L}$. Then, we crop their center regions that align with the local patch $I$ and resize them to the same dimension as $\textbf{X}_I$, which are denoted as $\tilde{\textbf{X}}_{M}$ and $\tilde{\textbf{X}}_{L}$, respectively. Hence, both of them involve the contextual information of varied scales without introducing too much noise.

Next, we want to employ them to jointly optimize each other and thus to yield optimal local features for compensating the outcome of LCF. To accomplish this purpose, we can treat $\{\textbf{X}_I, \tilde{\textbf{X}}_M, \tilde{\textbf{X}}_L\}$ equally, so any two of them can be utilized as \textit{reference} to enhance the other one (denoted as \textit{target}). Hence, we
propose the alternating local enhancement algorithm as explained below (Alg.~\ref{alg:op}).

\begin{algorithm}
    \renewcommand{\algorithmicrequire}{\textbf{Input:}}
	\renewcommand{\algorithmicensure}{\textbf{Output:}}
    \caption{\wxr{Alternating Local Enhancement}}\label{alg:op}

    \begin{algorithmic}[1]
    
        \Require 
        {Local features $\textbf{X}_{I}$, cropped-upscaled medium contextual features $\tilde{\textbf{X}}_{M}$, cropped-upscaled large contextual features $\tilde{\textbf{X}}_{L}$, local feature interaction function $f(\cdot;\cdot)$ that rely on reference features to enhance target features based on their relevance} 
        \Ensure 
        {Enhanced local features $\hat{\textbf{X}}_{I}$} 
        \State Apply $\{\tilde{\textbf{X}}_M, \tilde{\textbf{X}}_L\}$ as references to enhance $\textbf{X}_I$:
        $\tilde{\textbf{X}}_{I} = \mathop{\arg\min}_{\textbf{X}_{I}} f(\textbf{X}_{I} ; \{\tilde{\textbf{X}}_{M}, \tilde{\textbf{X}}_{L}\}) \label{eq:optim1}$\\
        Apply $\{\textbf{X}_I, \tilde{\textbf{X}}_L\}$ as references to enhance $\tilde{\textbf{X}}_M$: 
        $\hat{\textbf{X}}_{M} = \mathop{\arg\min}_{\tilde{\textbf{X}}_{M}} f(\tilde{\textbf{X}}_{M} ; \{\textbf{X}_{I}, \tilde{\textbf{X}}_{L}\}) \label{eq:optim2}$\\
        Apply $\{\textbf{X}_I, \tilde{\textbf{X}}_M\}$ as references to enhance $\tilde{\textbf{X}}_L$:  
        $\hat{\textbf{X}}_{L} = \mathop{\arg\min}_{\tilde{\textbf{X}}_{L}} f(\tilde{\textbf{X}}_{L} ; \{\textbf{X}_{I}, \tilde{\textbf{X}}_{M} \}) \label{eq:optim3}$\\
        Obtain $\hat{\textbf{X}}_I$ based on the previously enhanced features:  
        $\hat{\textbf{X}}_{I} = \mathop{\arg\min}_{\tilde{\textbf{X}}_{I}} f(\tilde{\textbf{X}}_{I} ; \{\hat{\textbf{X}}_{M}, \hat{\textbf{X}}_{L}\}) \label{eq:optim4}$
         
    \end{algorithmic}
\end{algorithm}

In our proposed scheme, we treat $\{\tilde{\textbf{X}}_M, \tilde{\textbf{X}}_L\}$ as references to optimize $\textbf{X}_I$ and thus obtain the enhanced features $\tilde{\textbf{X}}_I$ (L1 in Alg.~\ref{alg:op}). Similarly, we optimize $\tilde{\textbf{X}}_M$ and $\tilde{\textbf{X}}_L$ in the meantime (L2 and L3 in Alg.~\ref{alg:op}). Finally, the optimized features $\hat{\textbf{X}}_M$ and $\hat{\textbf{X}}_L$ are jointly applied to enhance the local features $\tilde{\textbf{X}}_I$ and thus yield $\hat{\textbf{X}}_I$ (L4 in Alg.~\ref{alg:op}).
Corresponding to the algorithm, we design the network structure as shown on the right of Fig. \ref{fig:local_framework}. Actually, our proposed ALE can also be flexibly transformed into different structures with varied depths and topology, as described in Fig.~\ref{fig:ALEs}.

As the core of the structure, $f(\cdot;\cdot)$ is a specific function block, dubbed local feature interaction (LFI) block, which leverages reference features to refine the target features. 
% For instance, in Eq.~\ref{eq:optim1}, $\textbf{X}_{I} \text{ and } \tilde{\textbf{X}}_{M}$ serve as the reference patches, while $\tilde{\textbf{X}}_{L}$ is the target patch.
\wxr{For instance, on L1 of Alg.~\ref{alg:op}, $\tilde{\textbf{X}}_{M} \text{ and } \tilde{\textbf{X}}_{L}$ serve as the reference features with $\textbf{X}_{I}$ as the target features.}
We will elaborate the LFI block in the following.

\subsubsection{Local Feature Interaction}

% \begin{figure}
%     \centering
%     \includegraphics[width=0.5\textwidth]{imgs/ev.png}
%     \caption{ev...}
%     \label{fig:ev}
% \end{figure}

As mentioned, LFI serves as a function $f$ that aims to utilize two reference patches to strengthen the target patch. Given the spatially-aligned features of the reference patches and target patch cropped from different contexts, they contain the semantic information in different granularity. To avoid introducing redundant information, we confine the pixel-level feature enhancement within small neighboring region. Hence, we design a novel spatial attention model to discover the local relation around each pixel. \wxb{Note that, the naive feature concatenation is able to accomplish the purpose of feature interactions. However, as the reference features and the target features are not perfectly aligned, it may easily cause performance degradation. 
% In addition, it remains a static feature fusion while the reference and target features are updated each other dynamically and check each other in our LFI.
}

% (Motivation, i.e. why)
% \textcolor{blue}{From the pixel level, the locality-aware contextual correlation module uses all pixels in the context to reconstruct local patch, which may contain redundant information. Therefore, we proposed the local-context interaction module to focus on enhancing the feature of the patch by using the information around the patch,  which contains the restricted context information. We use the locality-confined feature $\textbf{X}_k^c$ cropped from contextual feature and its location is corresponding to the local feature.}

% (comparing to local context)
% \textcolor{blue}{Compared with the local context, the local context uses all the pixels of the context, while $\textbf{X}_k^c$ only uses the features of the neighboring areas of the pixels, which restricted contextual information and reduces redundant information.}

% (Locality-Confined Attention)
% \textcolor{blue}{
Formally, the reference feature patches can be denoted as $\{\textbf{X}_{ref}^{(1)}, \textbf{X}_{ref}^{(2)}\}$ and the target feature patches are denoted as $\textbf{X}_{tar}$. 
In specific, the target features $\textbf{X}_{tar}$ and the reference features $\textbf{X}_{ref}^{(i)}$ ($i=\{1,2\}$) are tokenized via $1\times 1$ convolutional layers. 
The features of the reference patches are first concatenated as $\textbf{X}_{ref} = \text{Concat}\{\textbf{X}_{ref}^{(1)}, \textbf{X}_{ref}^{(2)}\}$.
Then, 
% To accomplish the locality, 
a 2D kernel $\kappa$ with the size $\omega \times \omega$ can be adopted to convolve the features $\textbf{X}_{ref}$ and derive an attention map $E$ ($E \in \mathbb{R}^{H\times W \times C}$) to strengthen the target features $\textbf{X}_{tar}$. 
% . Thus, by convolving the locality-confined features $\textbf{X}_k^c$ with the kernel $\kappa$, we can derive an attention map $E$ ($E \in \mathbb{R}^{H\times W \times C}$).
% }

\begin{figure}[t]
    \centering
    \footnotesize
	\begin{tabular}{c@{}c@{}c@{}c}
        \includegraphics[width=0.12\textwidth,height=1.8cm]
        {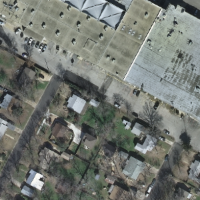}&
        \includegraphics[width=0.12\textwidth,height=1.8cm]
        {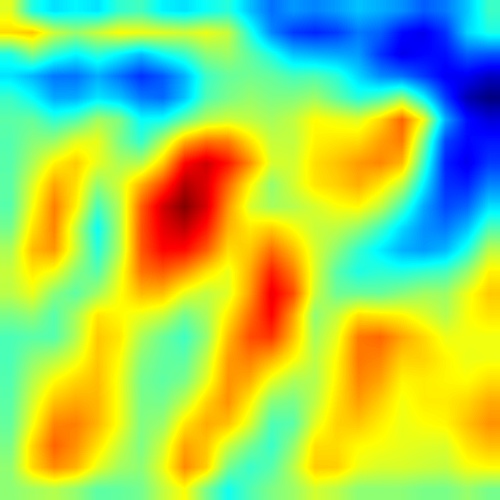}&
        \includegraphics[width=0.12\textwidth,height=1.8cm]
        {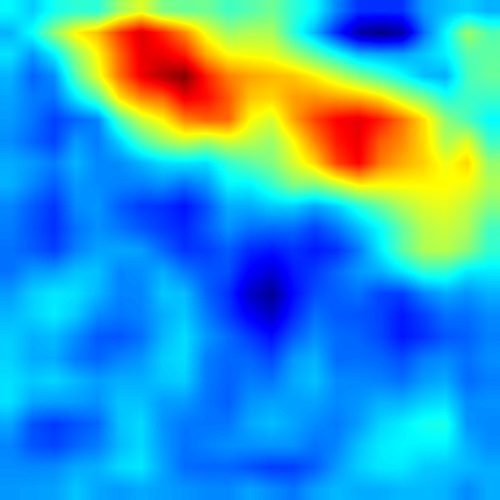}&
		\includegraphics[width=0.12\textwidth,height=1.8cm]
        {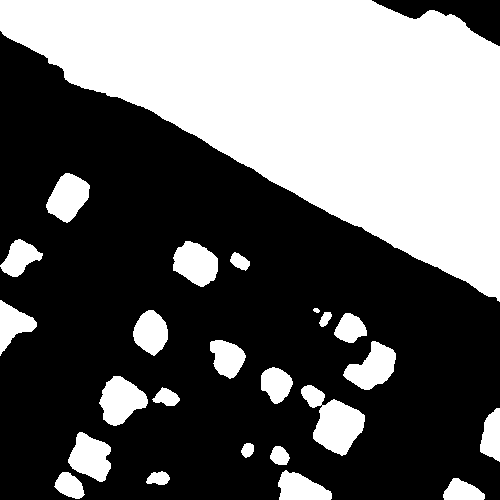}\\
        \includegraphics[width=0.12\textwidth,height=1.8cm]
        {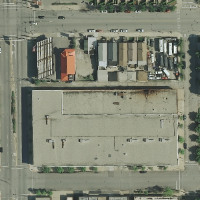}&
        \includegraphics[width=0.12\textwidth,height=1.8cm]
        {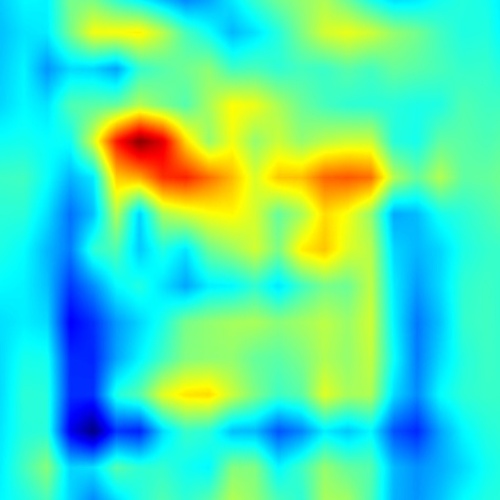}&
        \includegraphics[width=0.12\textwidth,height=1.8cm]
        {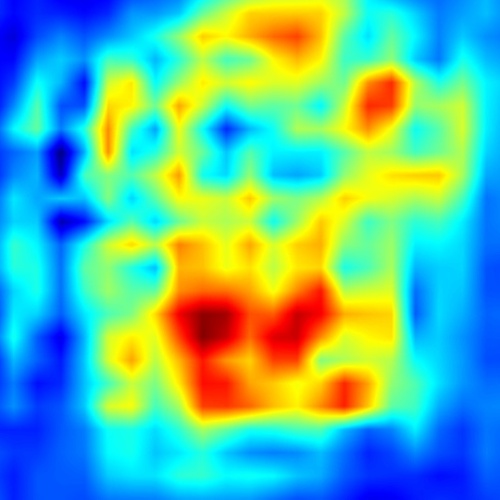}&
		\includegraphics[width=0.12\textwidth,height=1.8cm]
        {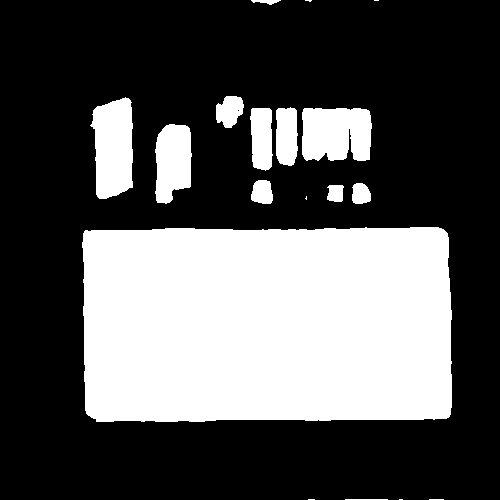}\\
        Local Patch & Small Window & Large Window & Ours \\
	\end{tabular}
	\caption{\wxb{We illustrate examples of LFI with small and large window sizes. As observed, the kernels with small window size encouarge to attend on the dense and small buildings, while the kernels with large window size are inclined to capture the large buildings.}}
	\label{fig:window}
\end{figure}

Unlike the standard spatial attention module, as illustrated in the network structure on the right part of Fig.~\ref{fig:local_framework}, we adopt kernels with different window sizes, \wxb{to involve feature information of varied granularity. In Fig.~\ref{fig:window}, we visualize two examples embodying the attention regions of different kernels.}
% to involve rich contextual information of varied scales. 
To do so, we slice the features $\textbf{X}_{ref}$ along channels into groups and then convolve the features of each group with a distinct 2D kernel via depthwise convolution, denoted as \textit{GroupDepthConv}, which can be expressed as below.
\begin{align}
    E = & \text{GroupDepthConv}_{i = \{1, \cdots, G\}}(\textbf{X}_{ref}; \{c_i\}, \{\kappa_i\}),\nonumber \\
    = & \oplus_{i=1}^G \text{DepthConv}(\textbf{X}_{ref}[c_{i-1}+1:c_i]; \kappa_i),
\end{align}
where $\oplus$ denotes the channel-wise concatenation and $G$ is the number of groups. $c_i$ represents the channel index ($c_i=0$ when $i=0$ and $c_i \in \mathbb{N}^{+}$ otherwise). $\textbf{X}_{ref}[c_{i-1}+1:c_i]$ implies the sliced feature maps of the $i$-th group, ranging from the $(c_{i-1}+1)$-th channel to $c_i$-th channel of $\textbf{X}_{ref}$. $\kappa_i$ refers to the 2D kernel with the size of $\omega_i \times \omega_i$. 
Last, we leverage the estimated attention map $E$ to enhance the target features via a residual structure, i.e.:
\begin{align}
    \overline{\textbf{X}}_{tar} &= \textbf{X}_{tar} + E \odot \textbf{X}_{tar},
\end{align}
where $\overline{\textbf{X}}_{tar}$ is the enhanced target features, and $\odot$ denotes element-wise multiplication. Hence, the computation of LFI can be expressed as $\overline{\textbf{X}}_{tar} = f(\textbf{X}_{tar}; \textbf{X}_{ref})$.

\subsection{Decoder and Loss Function}

As shown in Fig.~\ref{fig:framework}, the results of LCF (Eq.~\ref{eq:lcf}) and ALE (Eq.~\ref{eq:optim1}, \ref{eq:optim2}, and \ref{eq:optim3}) are combined with the local features, \textit{i.e.,} $\textbf{X}_I + \overline{\textbf{X}}_I+\hat{\textbf{X}}_I$, to yield the final optimal local features. \wxb{To sum up, $\textbf{X}_I$ retains most of the original local features extracted from the encoder. $\overline{\textbf{X}}_I$ supplies the contextual information with multi-context fusion to enhance $\textbf{X}_I$, while $\hat{\textbf{X}}_I$ restricts the negative impact of the redundant information of $\overline{\textbf{X}}_I$ via alternating local enhancement. Three of them complement each other to form the optimal features by a simple addition operation.}
Finally, the optimal features are then delivered into decoder to obtain context-aware local semantic masks, \wxb{which consists of three deconvolution layers as FCN \cite{long2015fully}.}
For supervision, we apply the focal loss \cite{Lin_2017_ICCV} as the objective function.

\section{Experimental Results}
\label{sec:exp}

In this section, we demonstrate the comprehensive experimental results over public benchmarks. We thoroughly compare our model against the state-of-the-art methods to show the segmentation quality and conduct the ablation study to evaluate the capability of our model.

\subsection{Datasets}

\textbf{DeepGlobe \cite{demir2018deepglobe}.} This dataset contains 803 ultra-high resolution images ($2448\times 2448$ pixels). Following \cite{chen2019collaborative}, we split images into training, validation and testing sets with 454, 207, and 142 images respectively. The dense annotation contains seven classes of landscape regions, including cyan represents "urban", yellow represents "agriculture", purple represents "rangeland", green represents "forest", blue represents "water", white represents "barren", where one class out of seven called "unknown" region is not considered in the challenge.

\noindent \textbf{Inria Aerial \cite{maggiori2017can}.} This dataset covers diverse urban landscapes, ranging from dense metropolitan districts to alpine resorts. It provides 180 images (from five cities) of $5000\times 5000$ pixels, each annotated with a binary mask for building/non-building areas. Unlike DeepGlobe, it splits the training/test sets by city. We follow the protocol as \cite{chen2019collaborative} by splitting images into training, validation and testing sets with 126, 27, and 27 images, respectively.

\subsection{Implementation Details}

\textbf{Settings for contexts.} \wxb{In practice, we apply three contexts in our model as mentioned above, which are denoted as local, medium, and large contexts.} The sizes of contexts differ for two benchmarks. We evaluate the performance of different context settings in Sec.~\ref{sec:abla}.

\noindent \textbf{Training details.} 
We implement our framework using Pytorch on a computer with a single NVIDIA GTX 1080Ti GPU. In particular, we adopt VGG16 \cite{Simonyan15} \wxb{and Segformer~\cite{xie2021segformer}} as our backbone and our baseline model is similar to FCN-8s \cite{long2015fully}. All the input images (i.e., local patches) are normalized to $508 \times 508$ and the output size is $508 \times 508$, which follows the setting of \cite{chen2019collaborative} in order to trade-off performance and efficiency.
When merging local results into a high-resolution one, we let neighboring patches have a $120 \times 508$ overlapping region to avoid boundary vanishing. 

During training our local segmentation model, we adopt the Adam optimizer and a mini-batch size of 6 by gradient accumulation. The initial learning rate is set to $5\times 10^{-5}$ and it is decayed by a poly learning rate policy where the initial learning rate is multiplied by $(1-\frac{iter}{total\_iter})^{0.9}$ after each iteration. In practice, it takes 50 epochs to converge our model. Besides, as our baseline and comparison method, FCN-8s\cite{long2015fully} also follows the training strategy above.

\noindent \textbf{Inference details.}
\lqq{During inference, we adopt the test-time augmentation technique (TTA) with rotation and flip. For a fair comparison, all the comparison models apply TTA in the following experiments.}

\subsection{Comparison with State-of-the-arts}

For evaluation, we compare our approach against U-Net \cite{ronneberger2015u}, ICNet \cite{zhao2018icnet}, PSPNet \cite{zhao2017pyramid}, SegNet \cite{badrinarayanan2017segnet}, DeepLab v3+ \cite{chen2018encoder}, FCN-8s \cite{long2015fully}, CascadePSP \cite{cheng2020cascadepsp}, GLNet  \cite{chen2019collaborative}, \wxb{Segformer~\cite{xie2021segformer}, and LCC \cite{li2021contexts}} over the benchmarks DeepGlobe and Inria Aerial, in terms of mIOU(\%), F1(\%), and Accuracy(\%). Their results are depicted in Table \ref{tab:table_deepglobe} and Table \ref{tab:table_inria}, in which we follow most of the quantitative results provided by \cite{chen2019collaborative}.
Most of these methods are not designed for ultra-high resolution images (denoted as \textit{Generic Model} in Table \ref{tab:table_deepglobe}), so there are two ways to train these models: 1) training from the local patches and merging local results; and 2) training from the downscaled global images. 
% They require the input images to be rescaled to the resolution of $508 \times 508$ before passing into the network. 
\begin{table}
    \centering
    \setlength{\tabcolsep}{1.8pt}
    \footnotesize
    \caption{Comparison with state-of-the-arts on DeepGlobe. }
    \begin{tabular}{cccccccc}
        \toprule
        \multirow{2}{*}{\textit{Generic Model}} & \multirow{2}{*}{\textit{Backbone}}  & \multicolumn{3}{c}{\textit{Local Inference}} & \multicolumn{3}{c}{\textit{Global Inference}} \\
        \cmidrule{3-8}
        % & \multicolumn{2}{c}{mIOU (\%)}\\
        & & mIOU & F1 & Acc. & mIOU & F1 & Acc.\\
        \midrule
        U-Net \cite{ronneberger2015u} & - & 37.3 & - & - & 38.4 & - & - \\
        ICNet \cite{zhao2018icnet} & ResNet50 & 35.5 & - & - & 40.2 & - & - \\
        PSPNet \cite{zhao2017pyramid} & ResNet50 & 53.3 & - & - & 56.6 & - & - \\
        DeepLab v3+ \cite{chen2018encoder} & ResNet50 & 63.1 & - & - & 63.5 & - & - \\
        % FCN-8s & 70.1?71.84 & 5227?1963\\
        GLNet* \cite{chen2019collaborative} & ResNet50 & 57.3 & 64.6 & 72.2 & 66.4 & 79.5 &85.8 \\
        SegNet \cite{badrinarayanan2017segnet} & VGG16 & 60.8 & - & - & 61.2 & - & - \\
        FCN-8s \cite{long2015fully} & VGG16 & 71.8 & 82.6 & 87.6 & 68.8 & 79.8 & 86.2\\
        Segformer\cite{xie2021segformer} & MiT-B2 & 74.4 & 84.6 & 88.5 & 54.6 & 68.6 & 78.3 \\
        Segformer\cite{xie2021segformer} & MiT-B5 & 73.6 & 84.1 & 88.2 & 70.3 & 81.4 & 86.8 \\
        \midrule
        \textit{High-Res Model} & \textit{Backbone} & \multicolumn{2}{c}{mIOU} & \multicolumn{2}{c}{F1} & \multicolumn{2}{c}{Acc.}\\
        \midrule
        CascadePSP \cite{cheng2020cascadepsp} & ResNet50  & \multicolumn{2}{c}{68.5} & \multicolumn{2}{c}{79.7} & \multicolumn{2}{c}{85.6} \\
        GLNet \cite{chen2019collaborative} & ResNet50 & \multicolumn{2}{c}{71.6} & \multicolumn{2}{c}{83.2} & \multicolumn{2}{c}{88.0}\\
        % MagNet \cite{huynh2021progressive} & \multicolumn{2}{c}{73.0} & \multicolumn{2}{c}{82.8} & \multicolumn{2}{c}{87.8}\\
        LCC \cite{li2021contexts} & VGG16 & \multicolumn{2}{c}{73.5} &\multicolumn{2}{c}{83.8}& \multicolumn{2}{c}{88.3}\\
        Ours & VGG16 & \multicolumn{2}{c}{73.9} &\multicolumn{2}{c}{84.1}& \multicolumn{2}{c}{\textbf{88.5}}\\
        % Segformer\cite{xie2021segformer}  & MiT-B2 & \multicolumn{2}{c}{74.4} & \multicolumn{2}{c}{84.6} & \multicolumn{2}{c}{\textbf{88.5}}\\
        Ours & MiT-B2 & \multicolumn{2}{c}{\textbf{74.8}} & \multicolumn{2}{c}{\textbf{84.9}} & \multicolumn{2}{c}{88.4}\\
        % Segformer\cite{xie2021segformer} & MiT-B5 & \multicolumn{2}{c}{73.6} & \multicolumn{2}{c}{84.1} & \multicolumn{2}{c}{88.2}\\
        Ours & MiT-B5 & \multicolumn{2}{c}{73.7} & \multicolumn{2}{c}{84.2} & \multicolumn{2}{c}{88.4}\\
        \bottomrule
    \end{tabular}
    \label{tab:table_deepglobe}
\end{table}

\begin{table}
    \centering
    \setlength{\tabcolsep}{7pt}
    \footnotesize
    \caption{Comparison with state-of-the-arts on Inria Aerial.}
    \begin{tabular}{ccccc}
        \toprule
        \textit{Model} & \textit{Backbone} & mIOU & F1 & Acc. \\
        \midrule
        ICNet \cite{zhao2018icnet} & ResNet50 & 31.1 & - & - \\
        DeepLab v3+ \cite{chen2018encoder} & ResNet50 & 55.9 & - & - \\
        CascadePSP \cite{cheng2020cascadepsp} & ResNet50 & 69.4 & 81.8 & 93.2\\
        GLNet \cite{chen2019collaborative} & ResNet50 & 71.2 & 82.4 & 93.8 \\
        FCN-8s \cite{long2015fully} & VGG16 & 69.1 & 81.7 & 93.6 \\
        LCC \cite{li2021contexts} & VGG16 & 73.7 & 84.1 & 94.6\\
        Segformer \cite{xie2021segformer} & MiT-B2 & 76.4 & 86.6 & 95.2\\
        Segformer \cite{xie2021segformer} & MiT-B5 & 77.9 & 87.5 & 95.5\\
        \midrule
        Ours & VGG16 & 74.6 & 85.4 & 94.8\\
        Ours & MiT-B2 & 77.4 & 87.3 & 95.4\\
        Ours & MiT-B5 & \textbf{78.7} & \textbf{88.1} & \textbf{95.7}\\
        \bottomrule
    \end{tabular}
    \label{tab:table_inria}
\end{table}

\begin{figure*}[t]
    \centering
    \footnotesize
	\begin{tabular}{c@{}c@{}c@{}c@{}c@{}c@{}c}
        
        \includegraphics[width=0.14\textwidth,height=2.2cm]
        {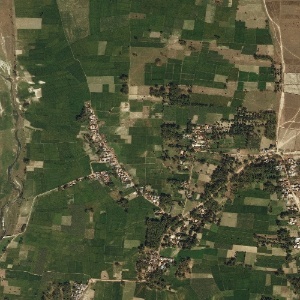} &
		\includegraphics[width=0.14\textwidth,height=2.2cm]
        {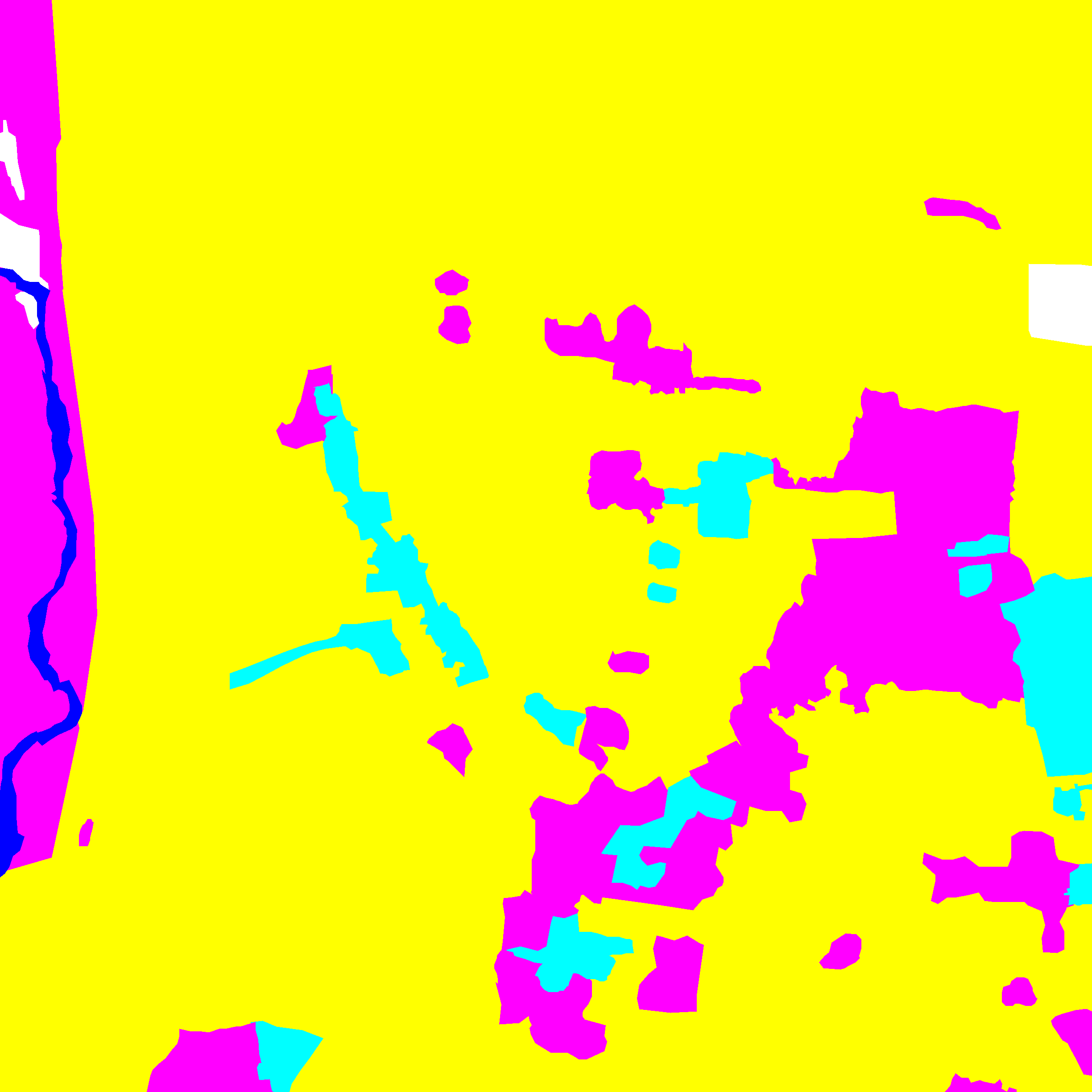}&
		\includegraphics[width=0.14\textwidth,height=2.2cm]
        {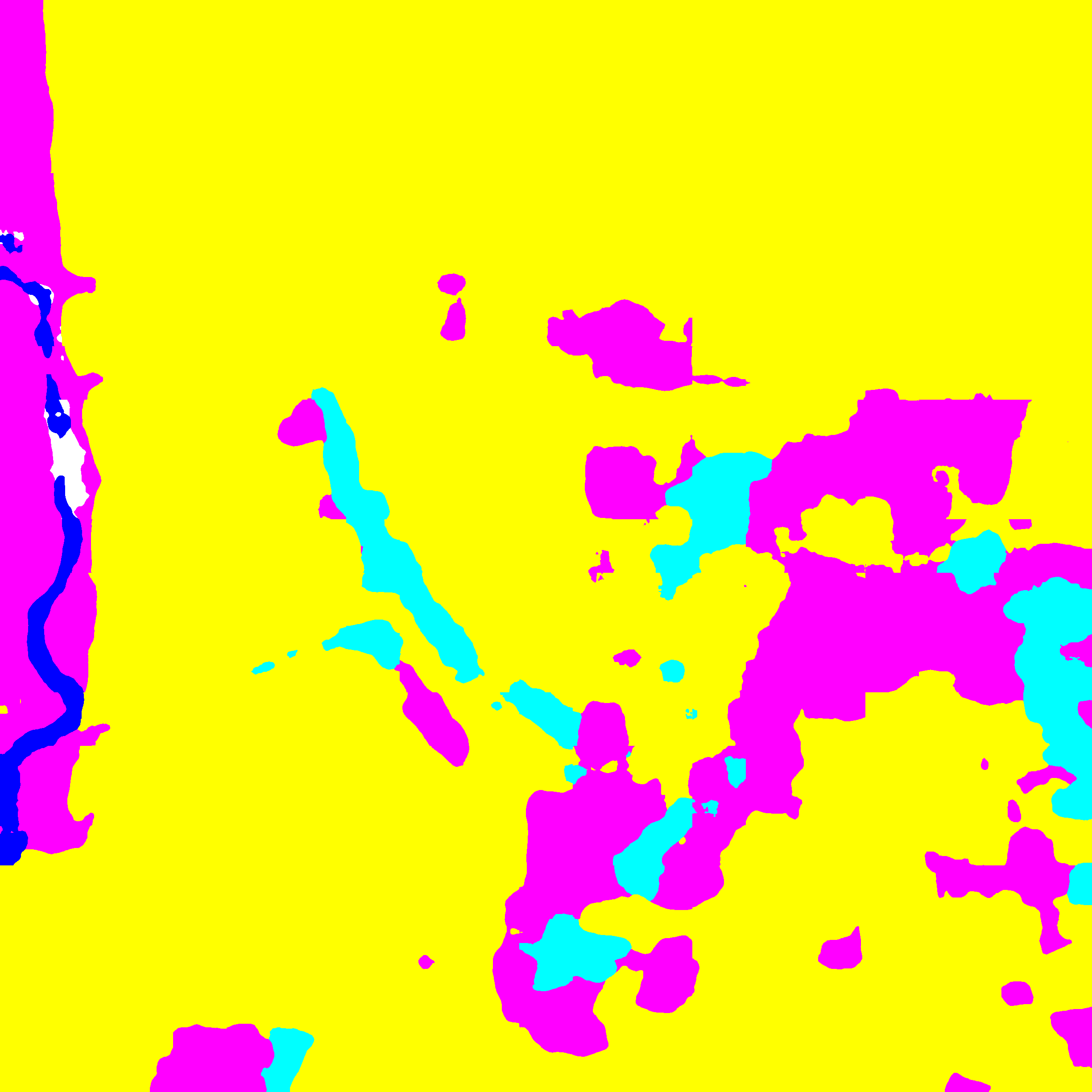}&
        \includegraphics[width=0.14\textwidth,height=2.2cm]
        {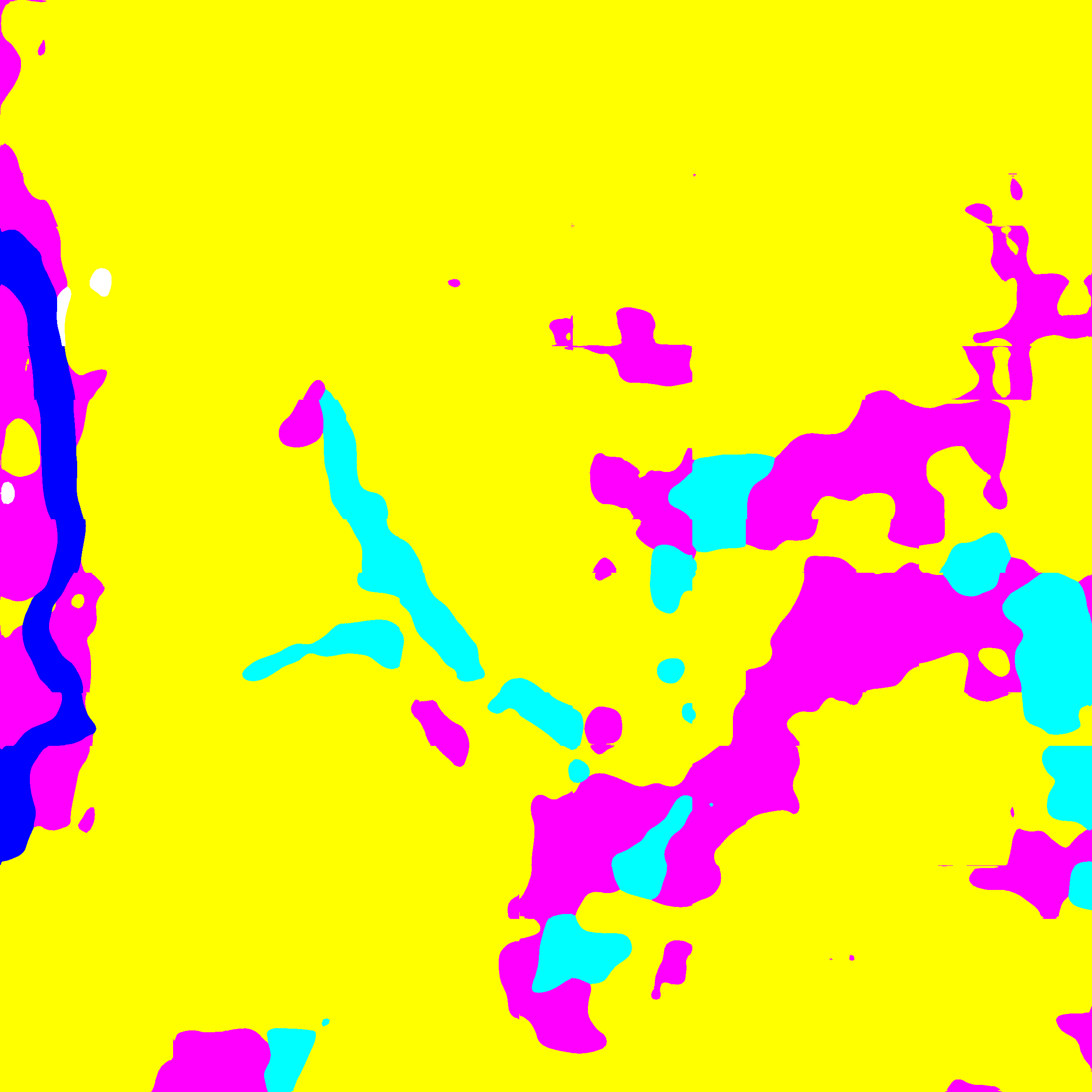}&
        \includegraphics[width=0.14\textwidth,height=2.2cm]
        {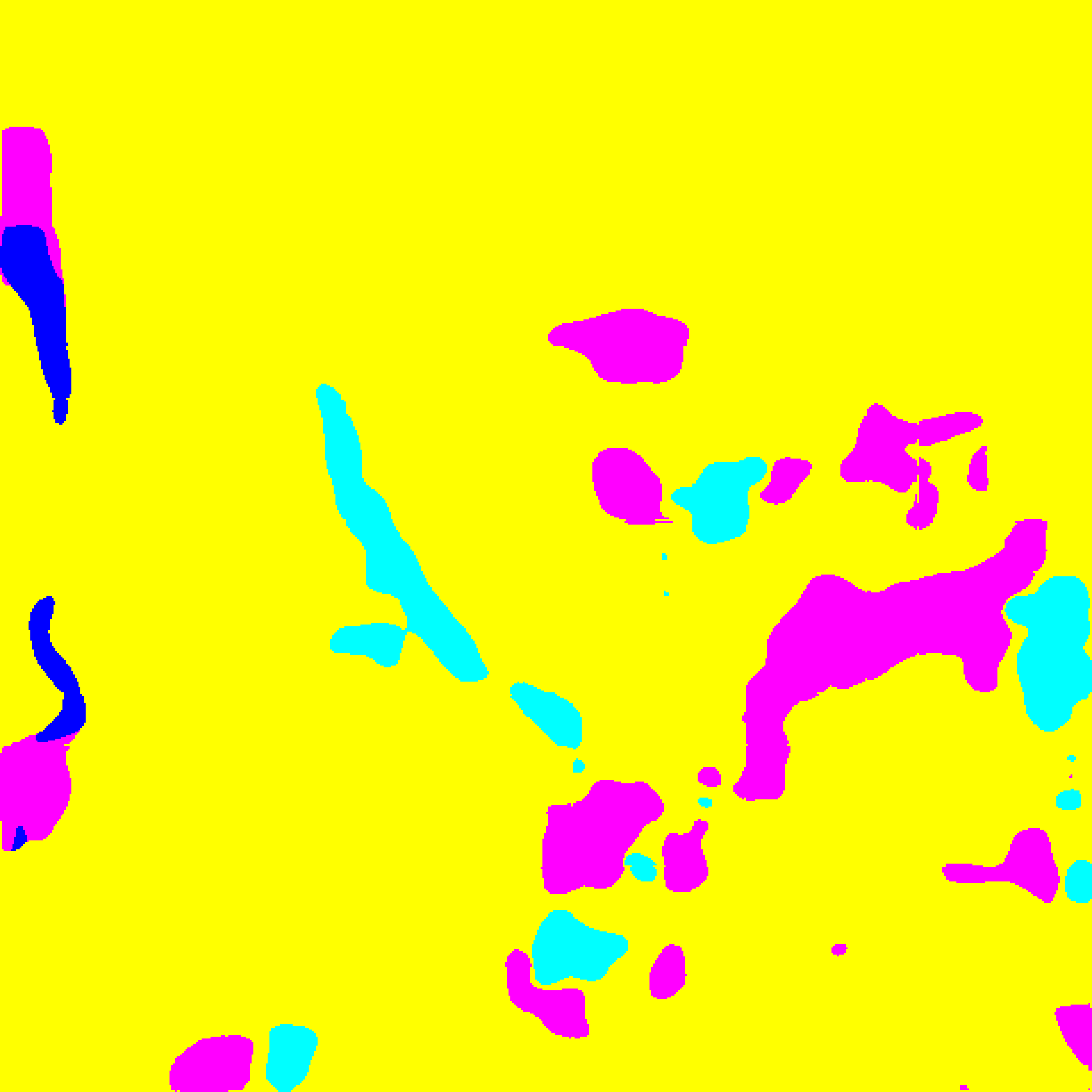}&
        \includegraphics[width=0.14\textwidth,height=2.2cm]
        {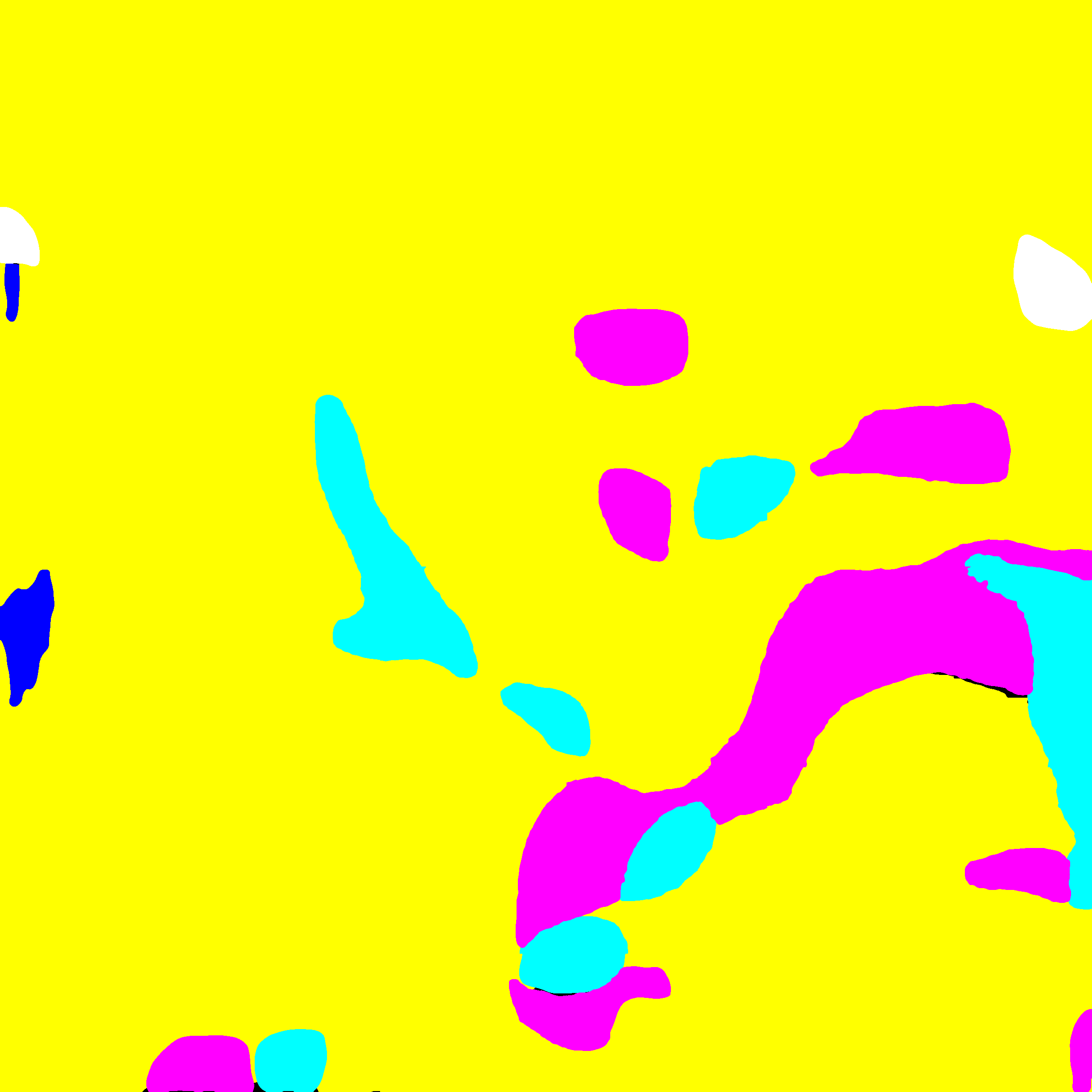}&
		\includegraphics[width=0.14\textwidth,height=2.2cm]
        {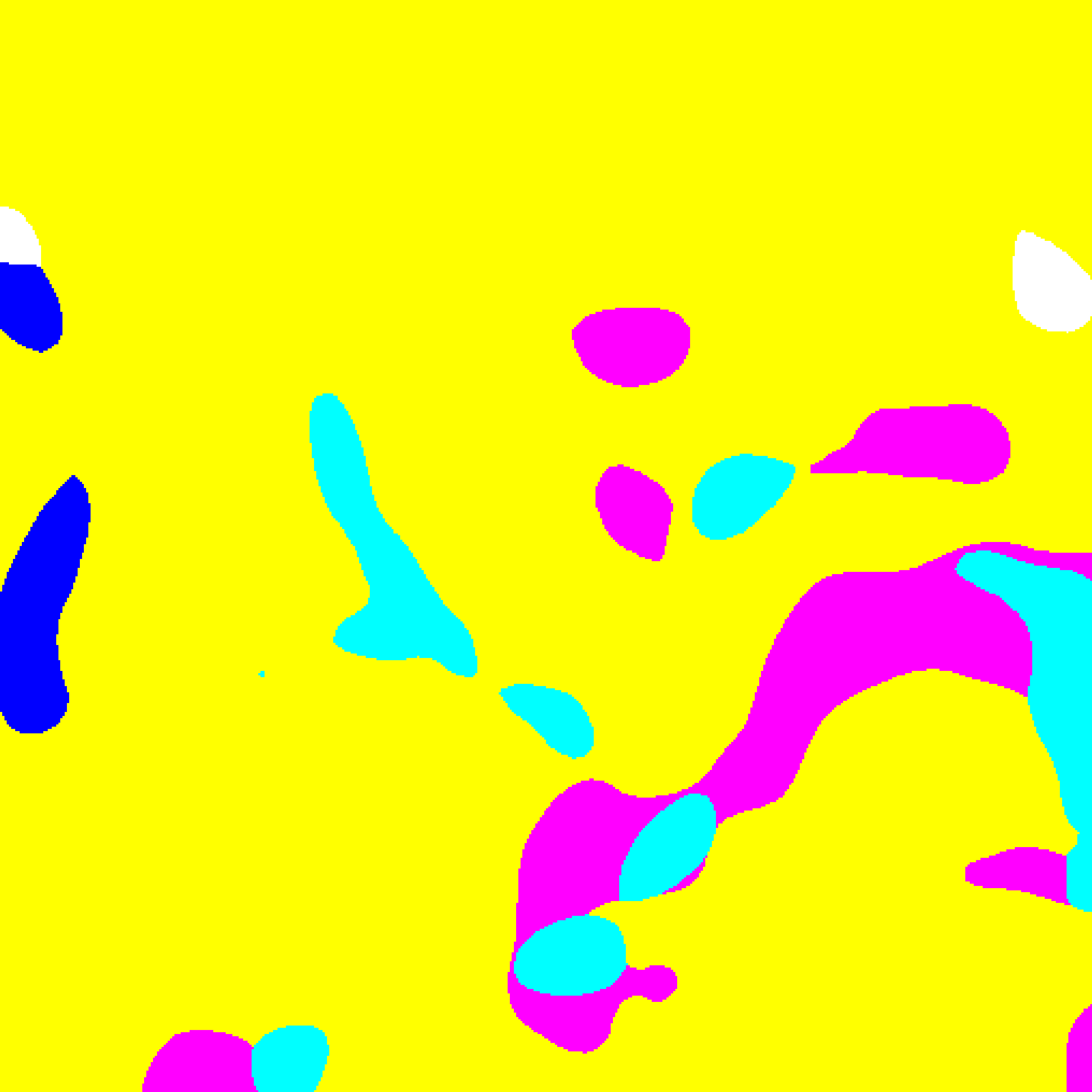}\\
        
       \includegraphics[width=0.14\textwidth,height=2.2cm]
        {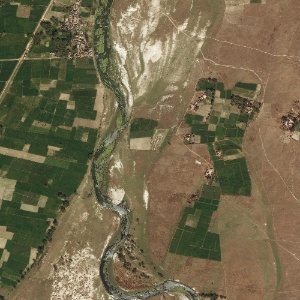} &
		\includegraphics[width=0.14\textwidth,height=2.2cm]
        {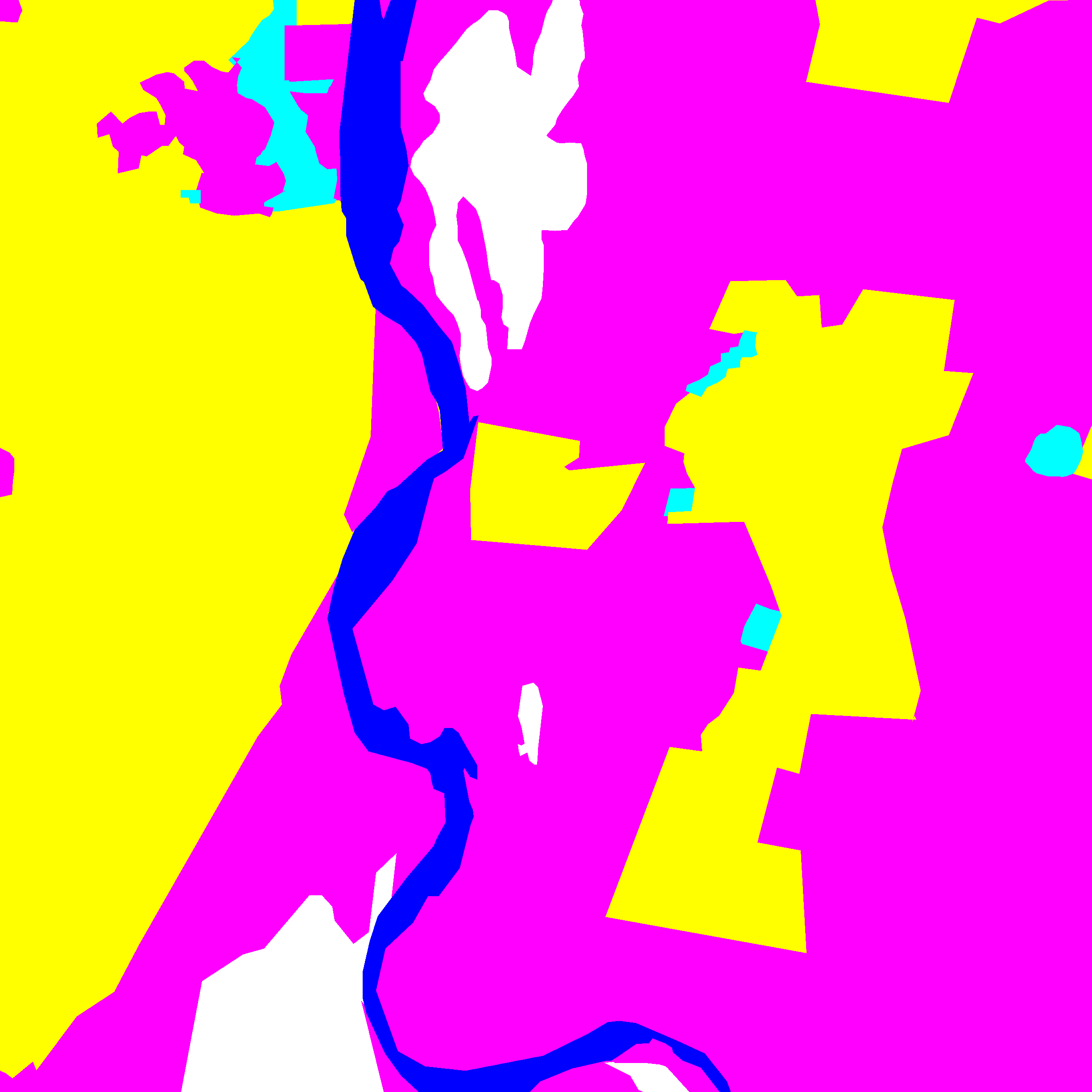}&
		\includegraphics[width=0.14\textwidth,height=2.2cm]
        {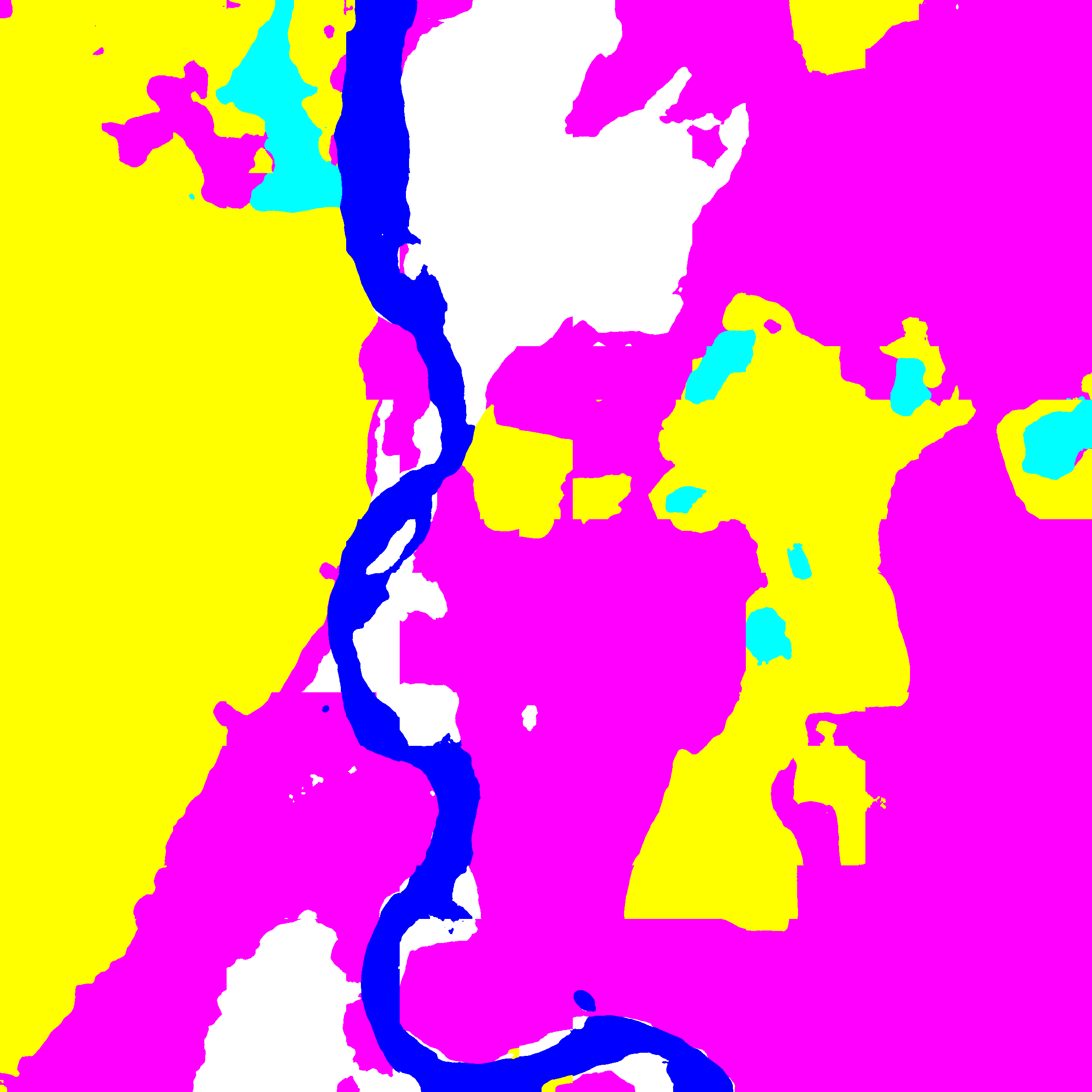}&
        \includegraphics[width=0.14\textwidth,height=2.2cm]
        {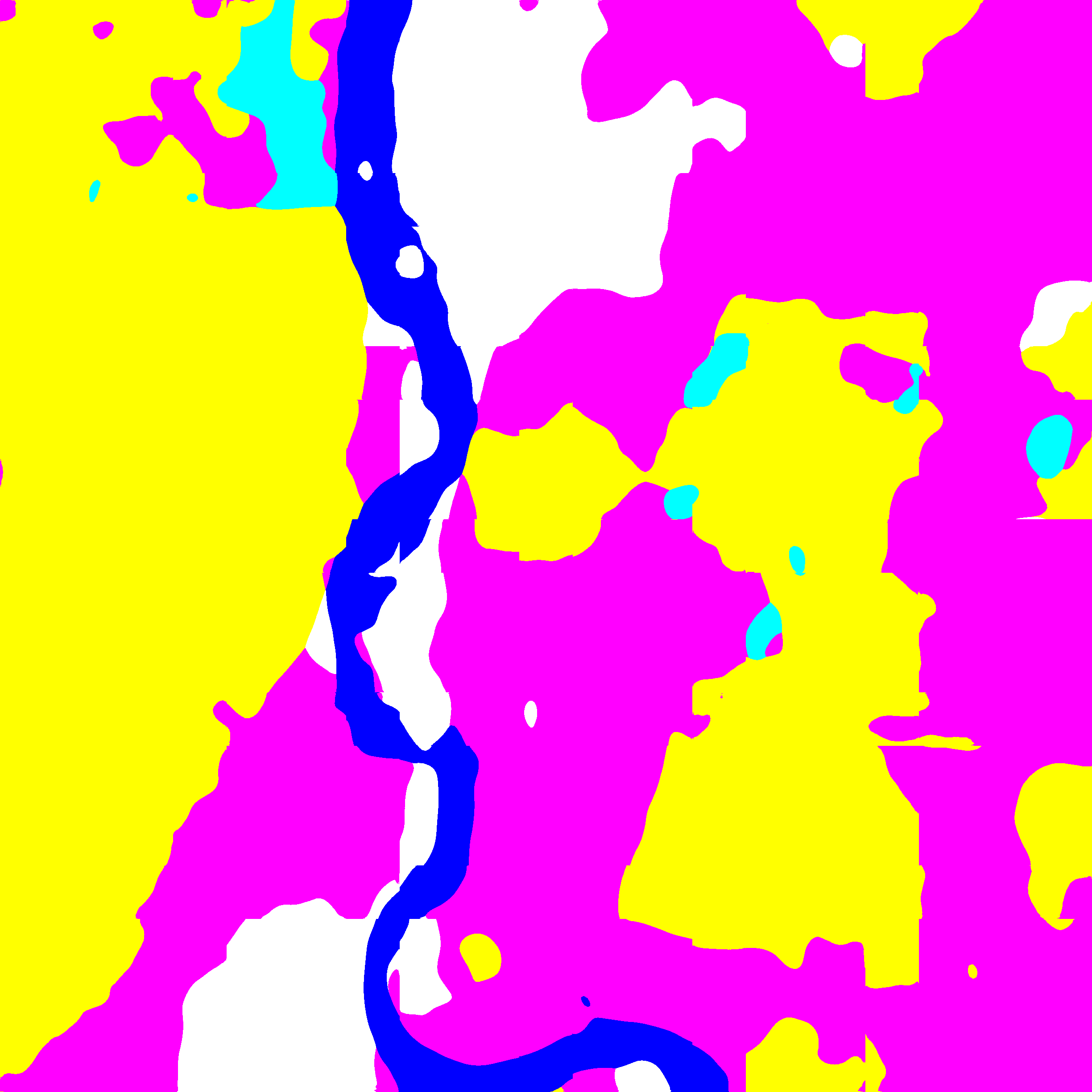}&
        \includegraphics[width=0.14\textwidth,height=2.2cm]
        {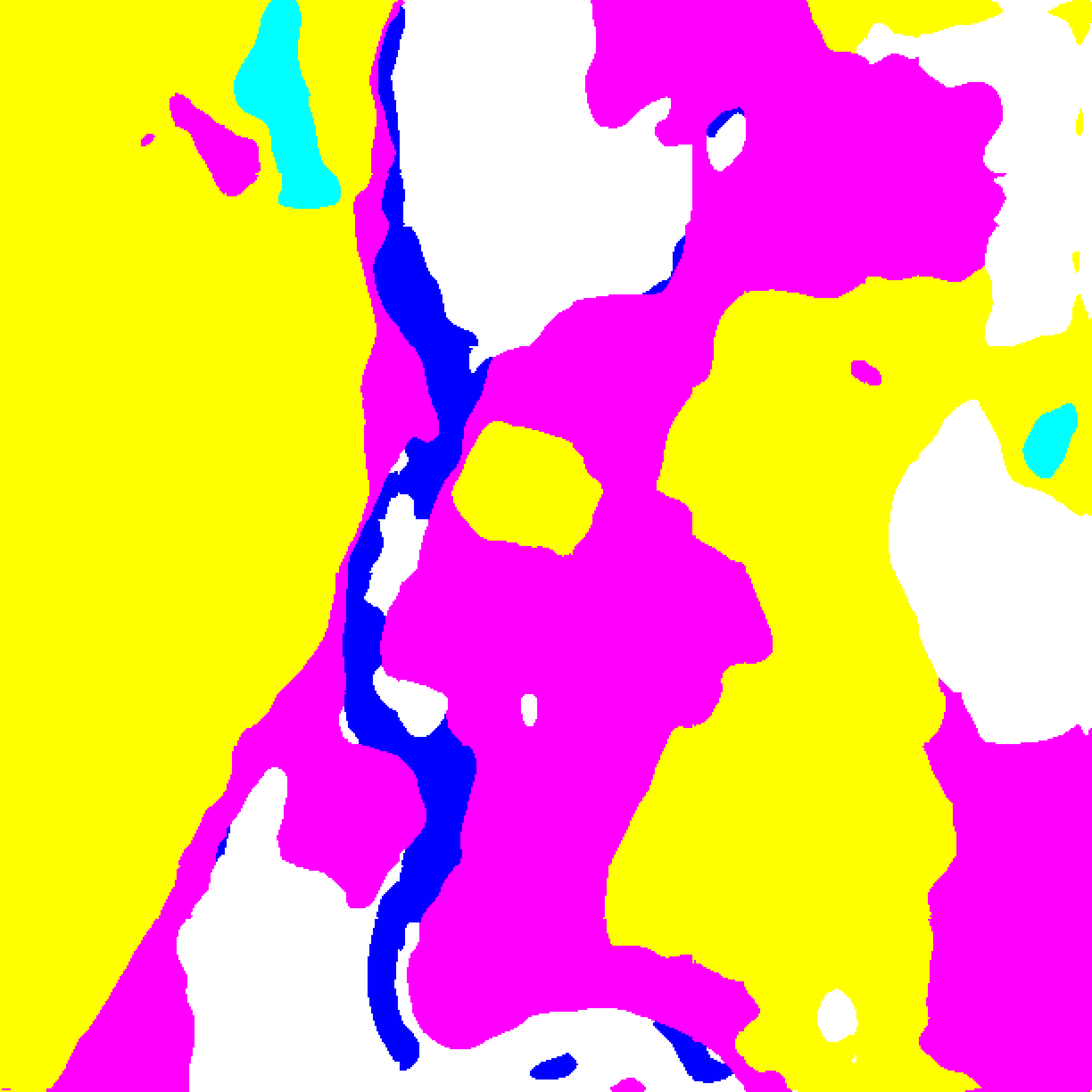}&
        \includegraphics[width=0.14\textwidth,height=2.2cm]
        {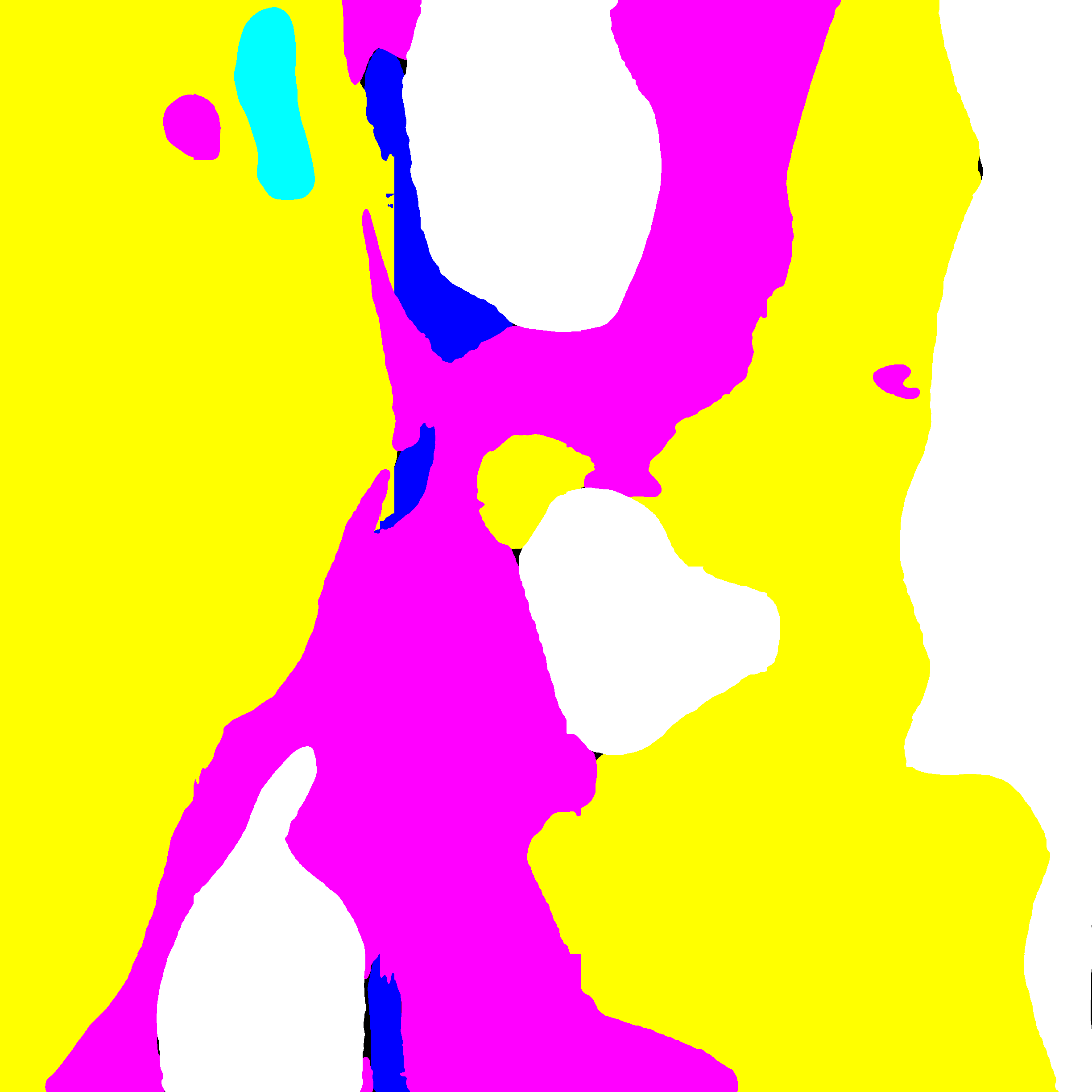}&
		\includegraphics[width=0.14\textwidth,height=2.2cm]
        {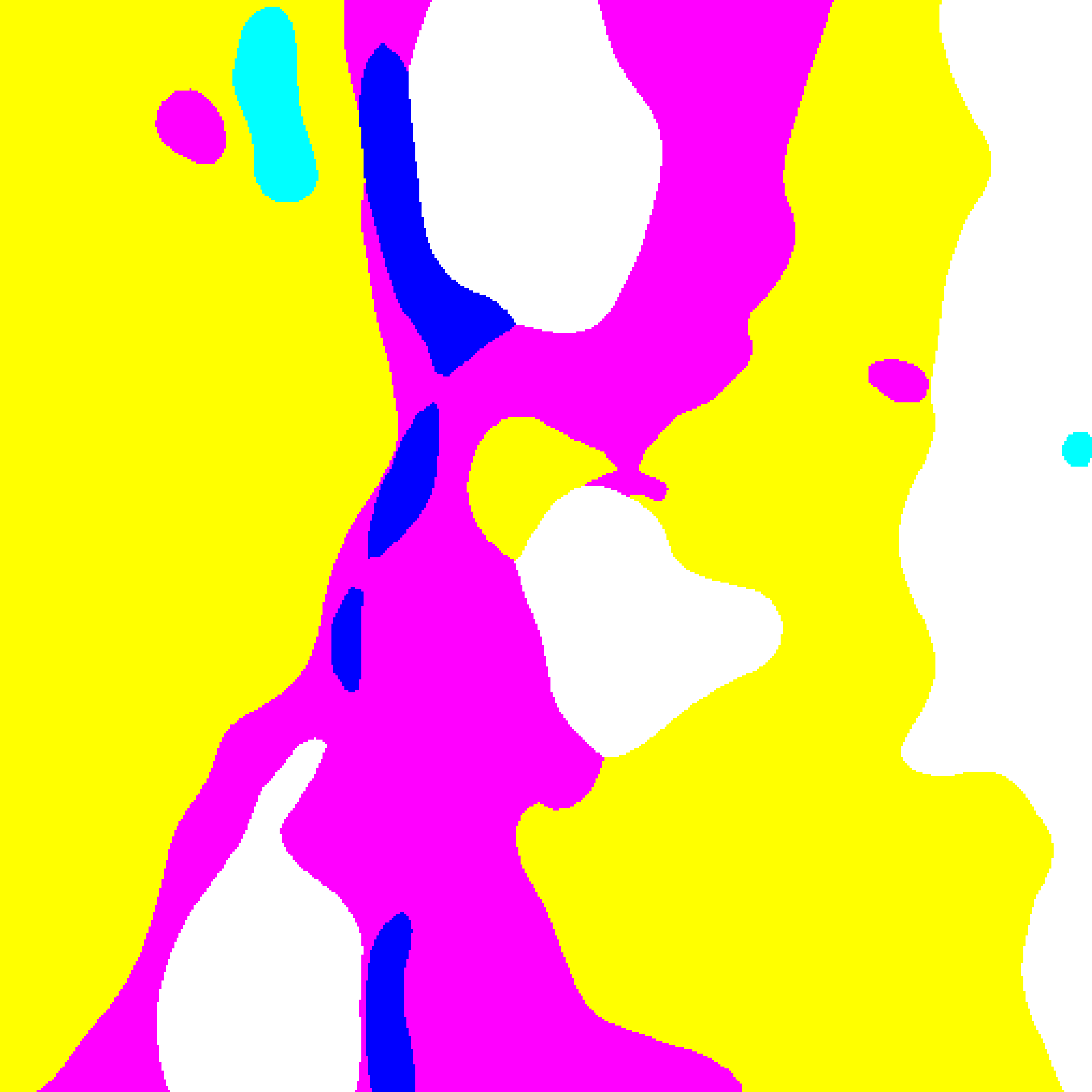}\\
                
        \includegraphics[width=0.14\textwidth,height=2.2cm]
        {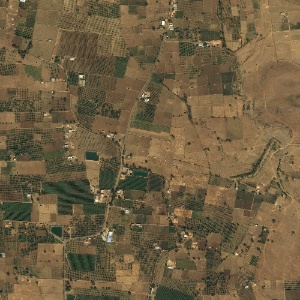} &
		\includegraphics[width=0.14\textwidth,height=2.2cm]
        {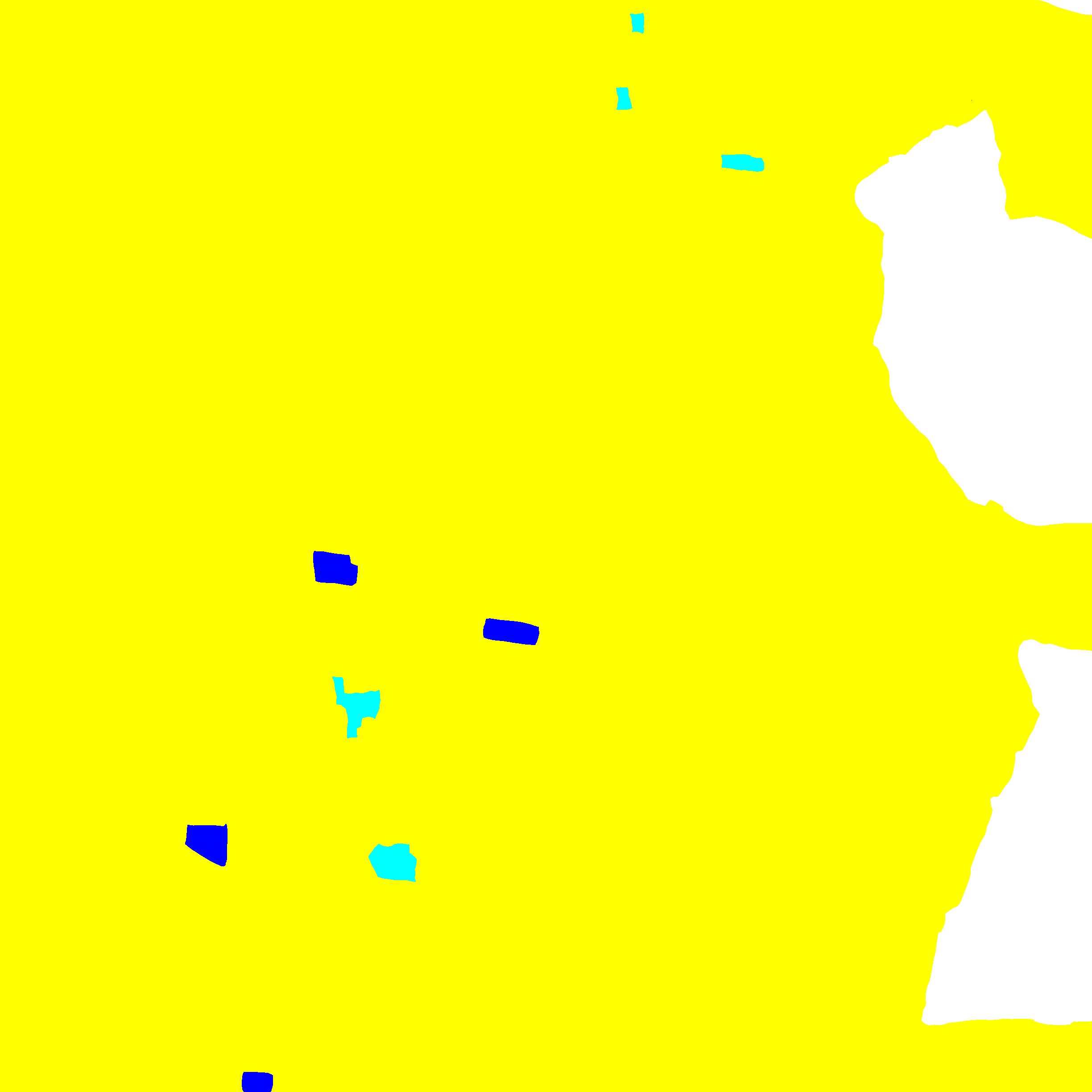}&
		\includegraphics[width=0.14\textwidth,height=2.2cm]
        {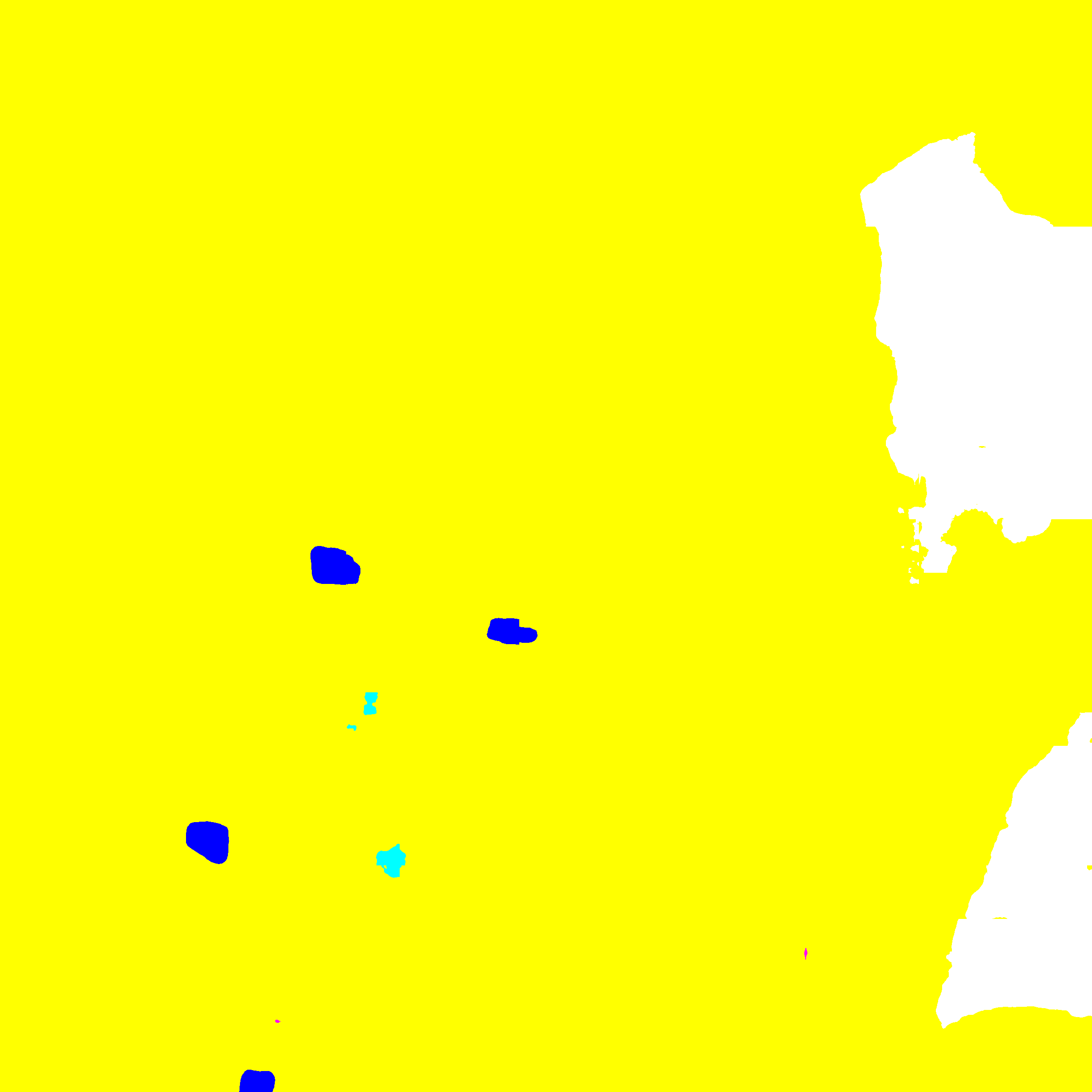}&
		\includegraphics[width=0.14\textwidth,height=2.2cm]
        {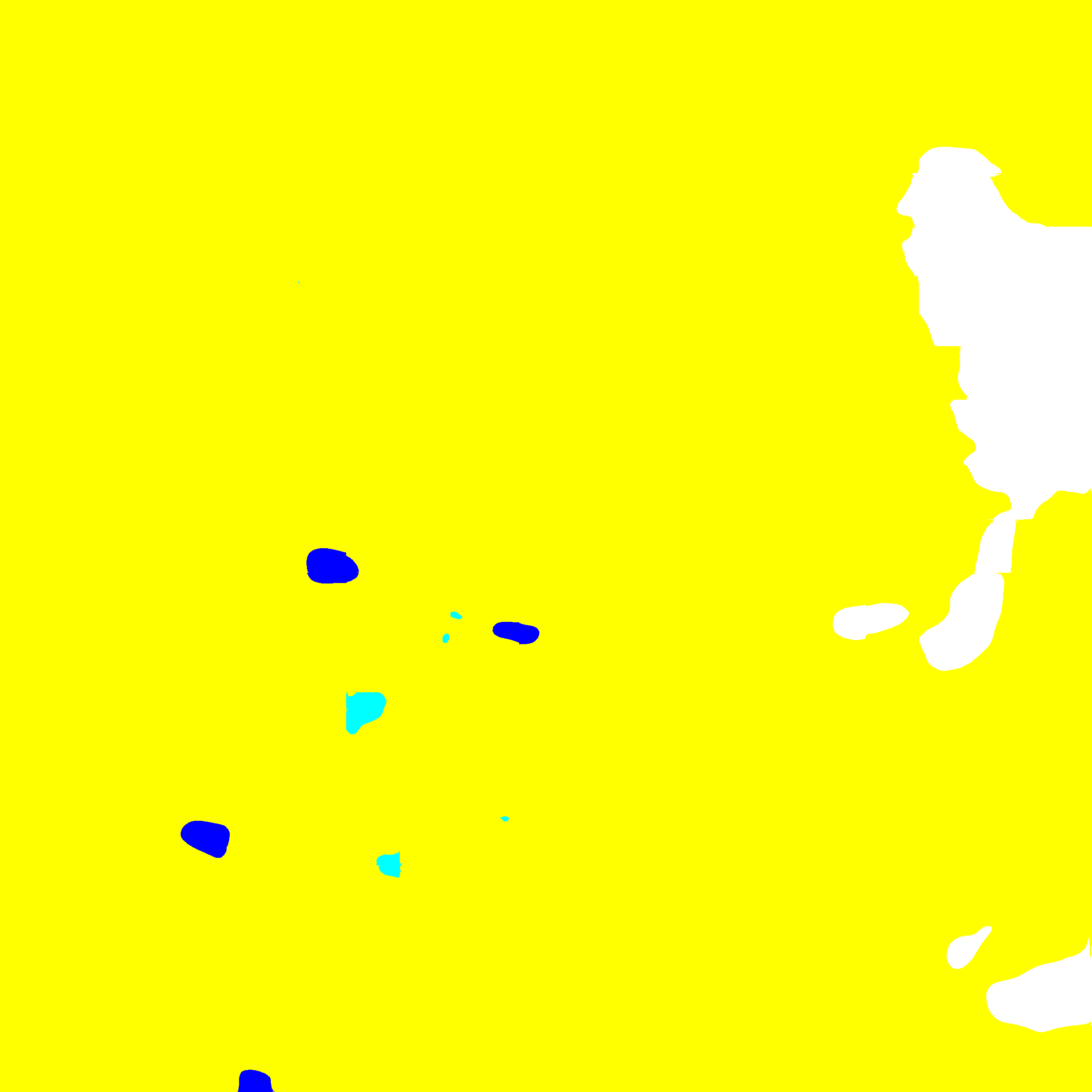}&
        \includegraphics[width=0.14\textwidth,height=2.2cm]
        {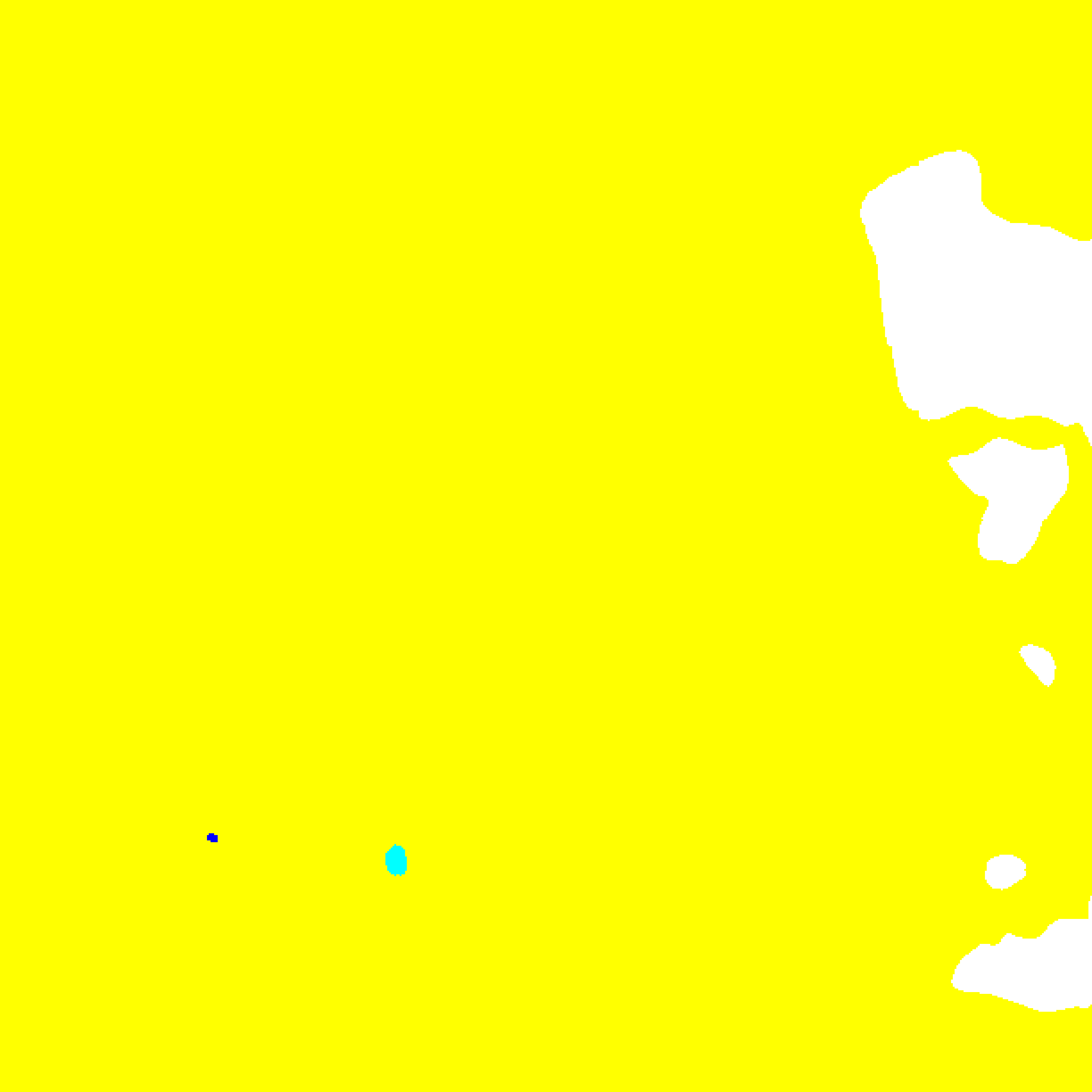}&
        \includegraphics[width=0.14\textwidth,height=2.2cm]
        {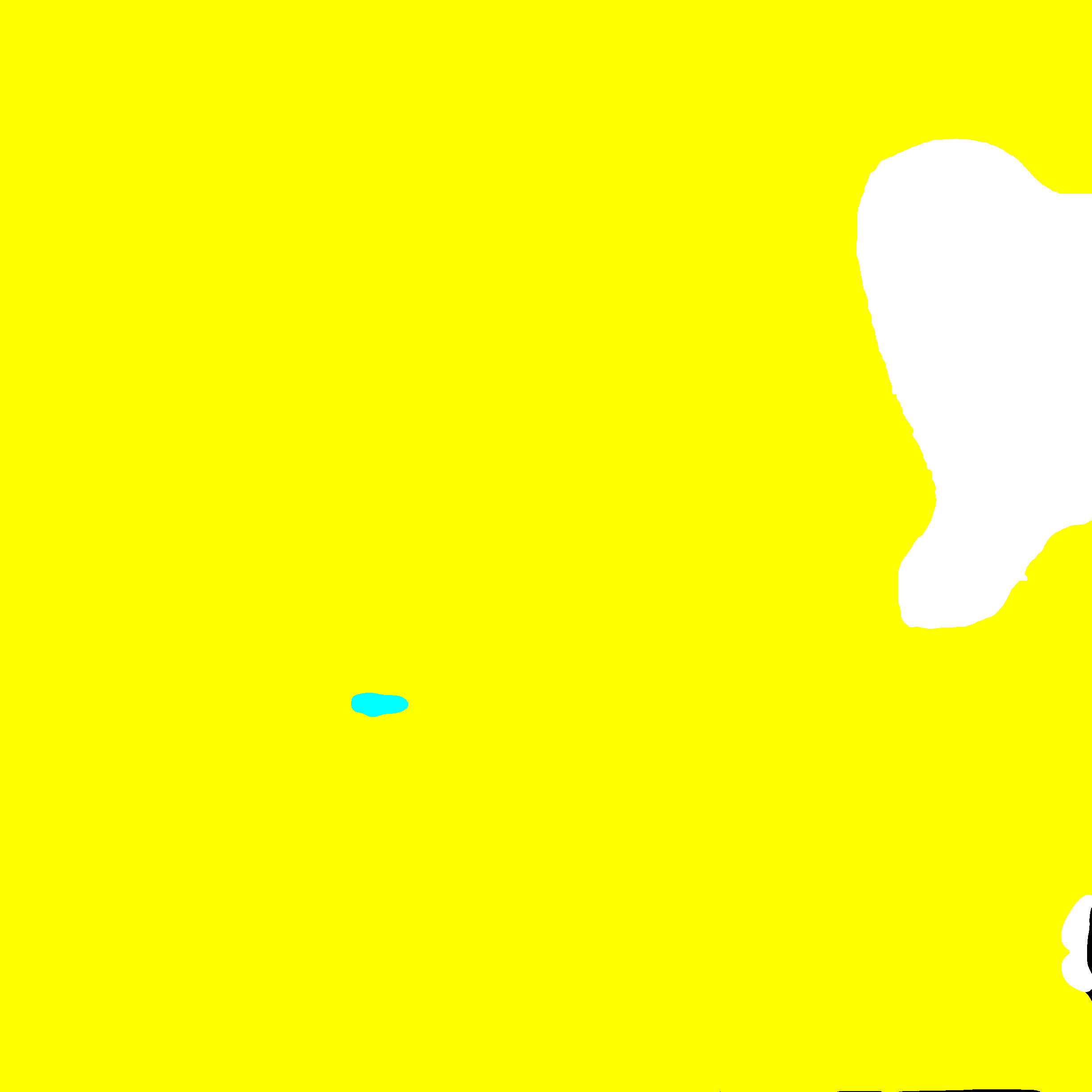}&
		\includegraphics[width=0.14\textwidth,height=2.2cm]
        {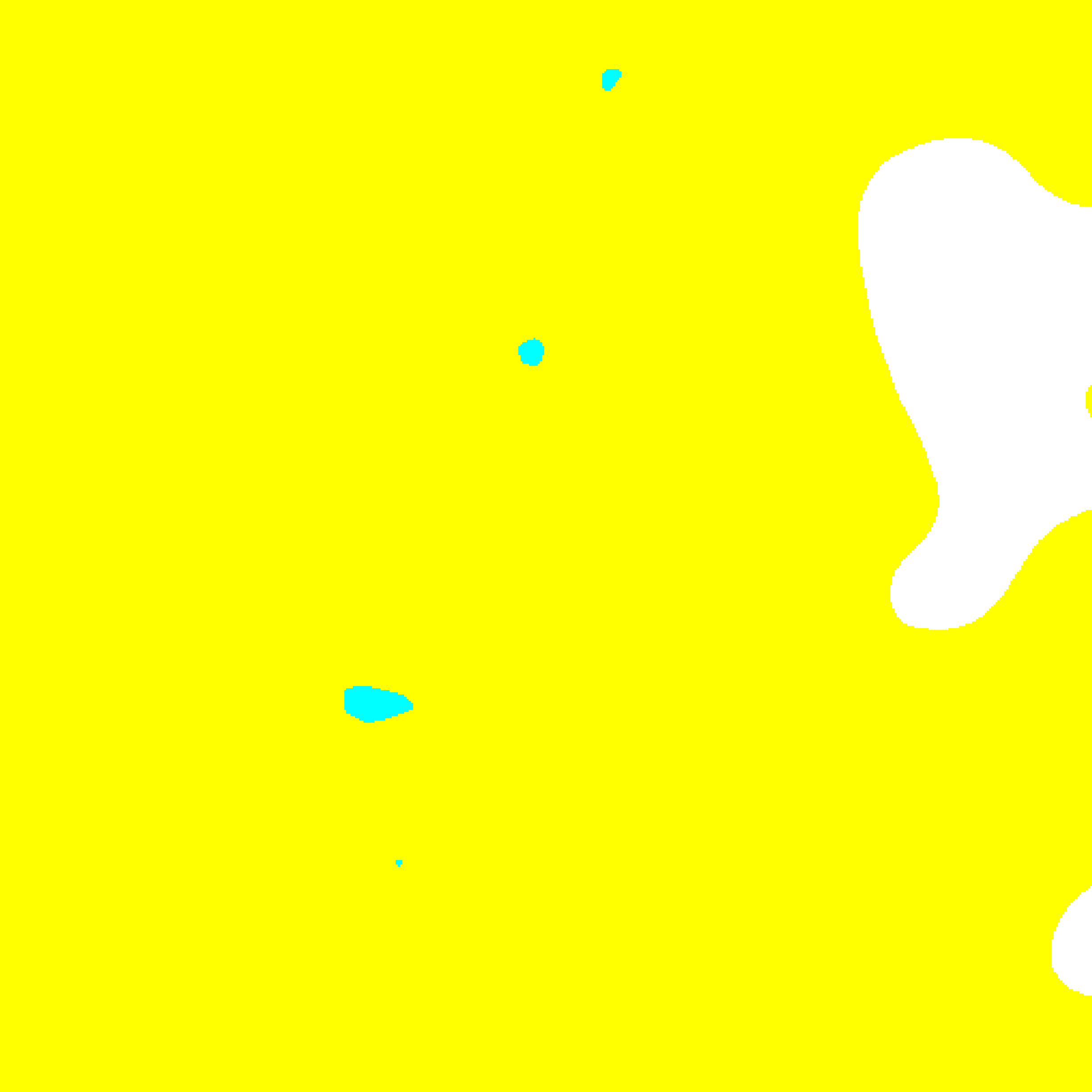}\\
        
        \includegraphics[width=0.14\textwidth,height=2.2cm]
        {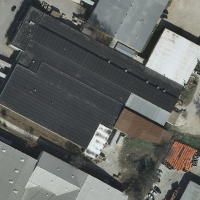} &
		\includegraphics[width=0.14\textwidth,height=2.2cm]
        {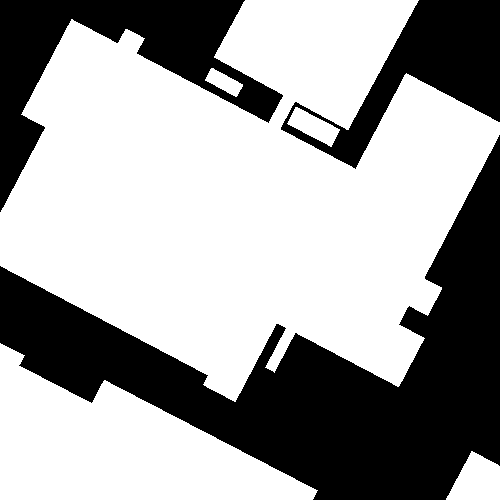}&
		\includegraphics[width=0.14\textwidth,height=2.2cm]
        {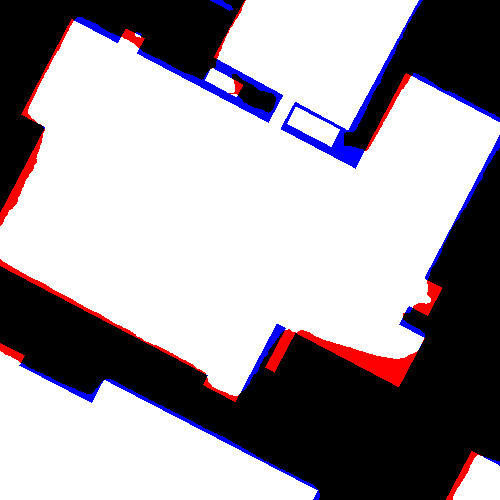}&
        \includegraphics[width=0.14\textwidth,height=2.2cm]
        {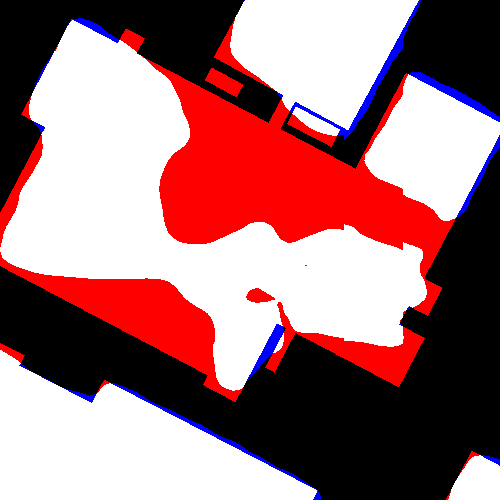}&
        \includegraphics[width=0.14\textwidth,height=2.2cm]
        {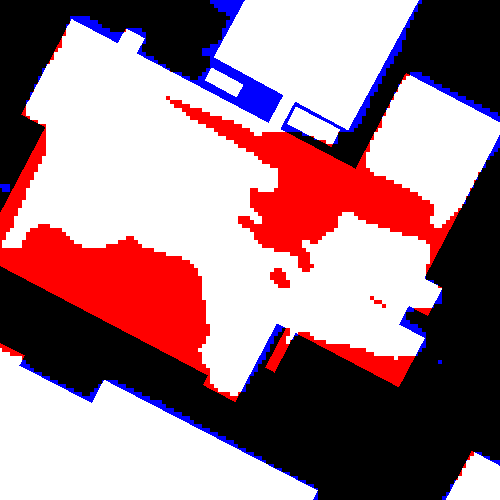}&
        \includegraphics[width=0.14\textwidth,height=2.2cm]
        {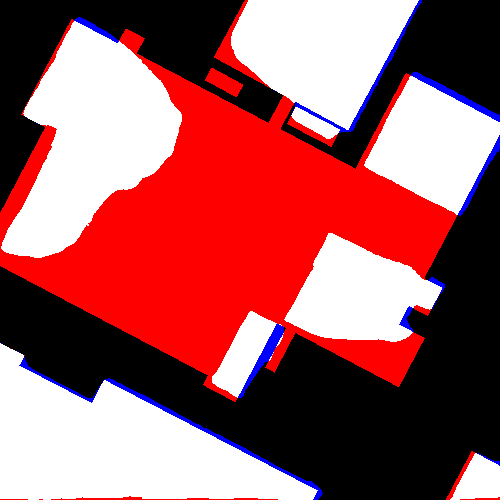}&
		\includegraphics[width=0.14\textwidth,height=2.2cm]
        {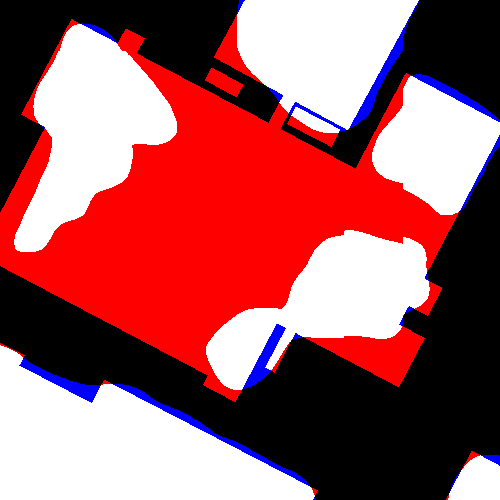}\\
        
        \includegraphics[width=0.14\textwidth,height=2.2cm]
        {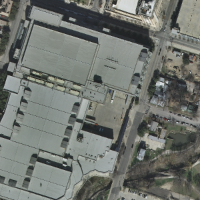} &
		\includegraphics[width=0.14\textwidth,height=2.2cm]
        {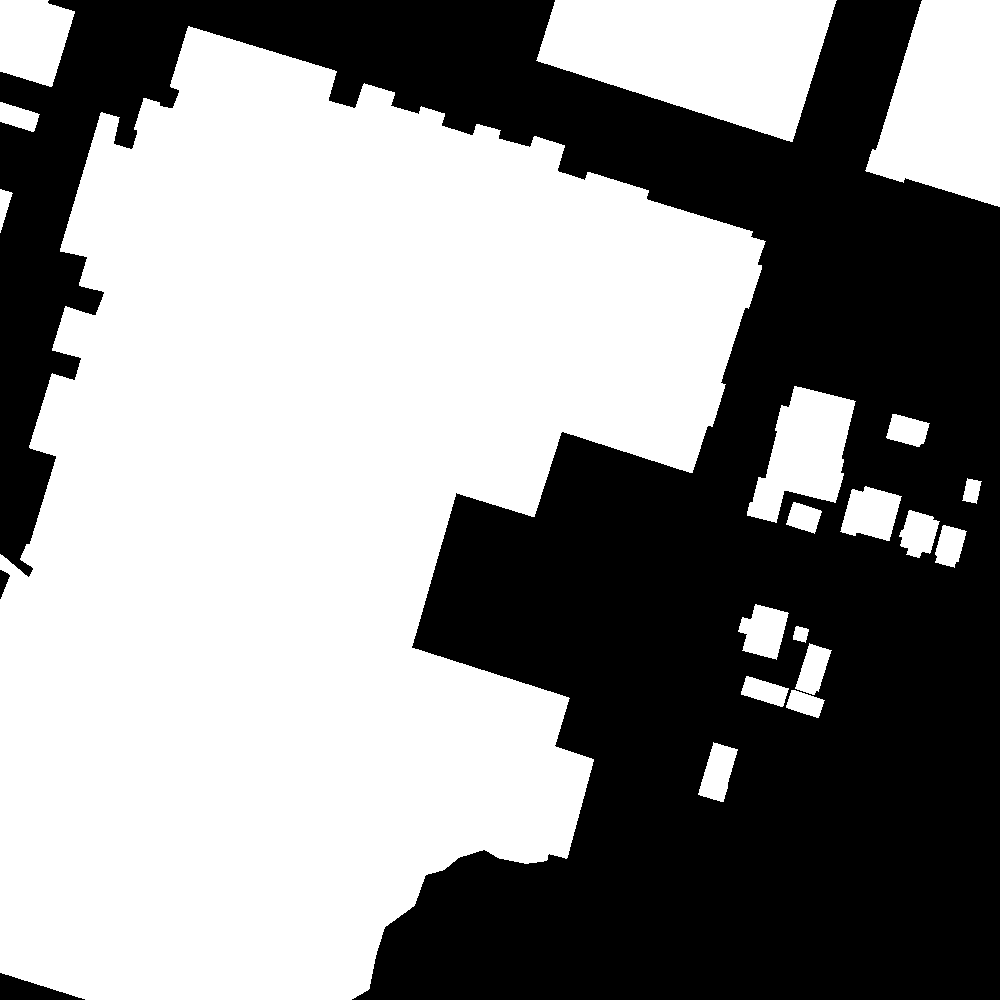}&
		\includegraphics[width=0.14\textwidth,height=2.2cm]
        {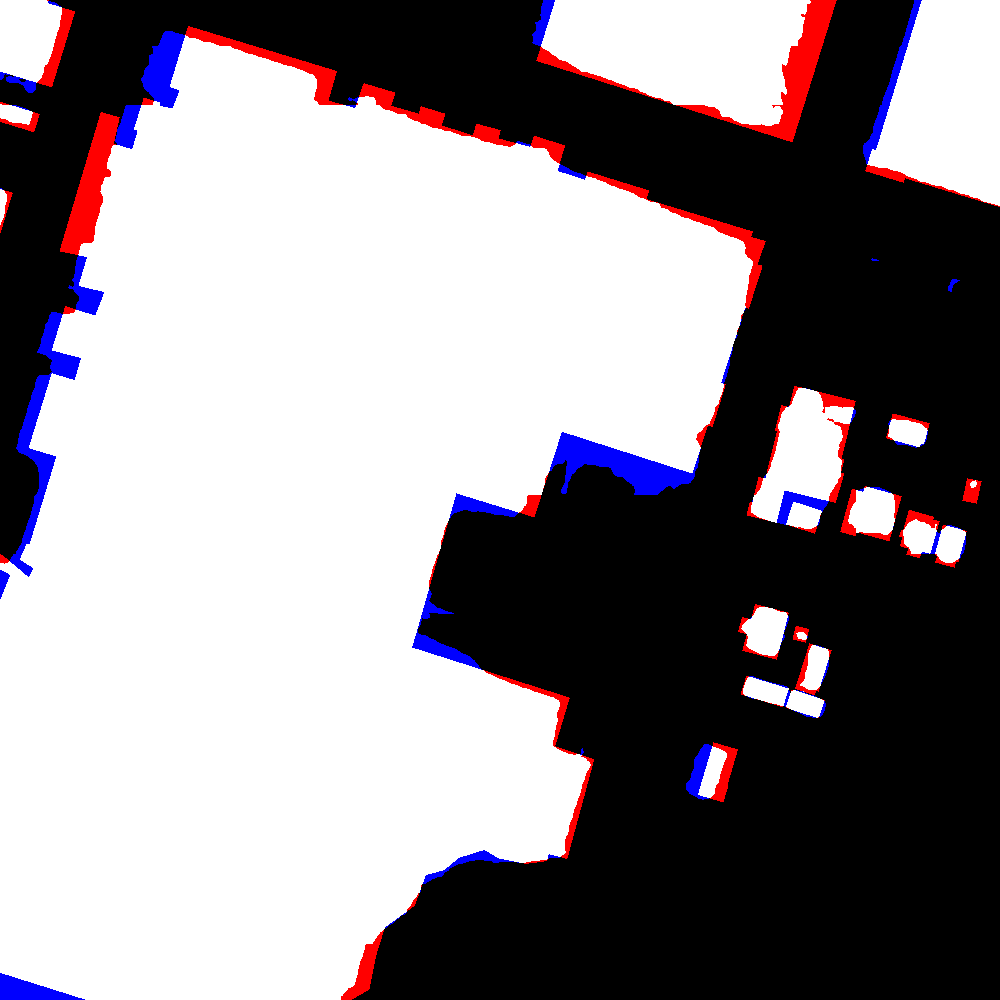}&
		\includegraphics[width=0.14\textwidth,height=2.2cm]
        {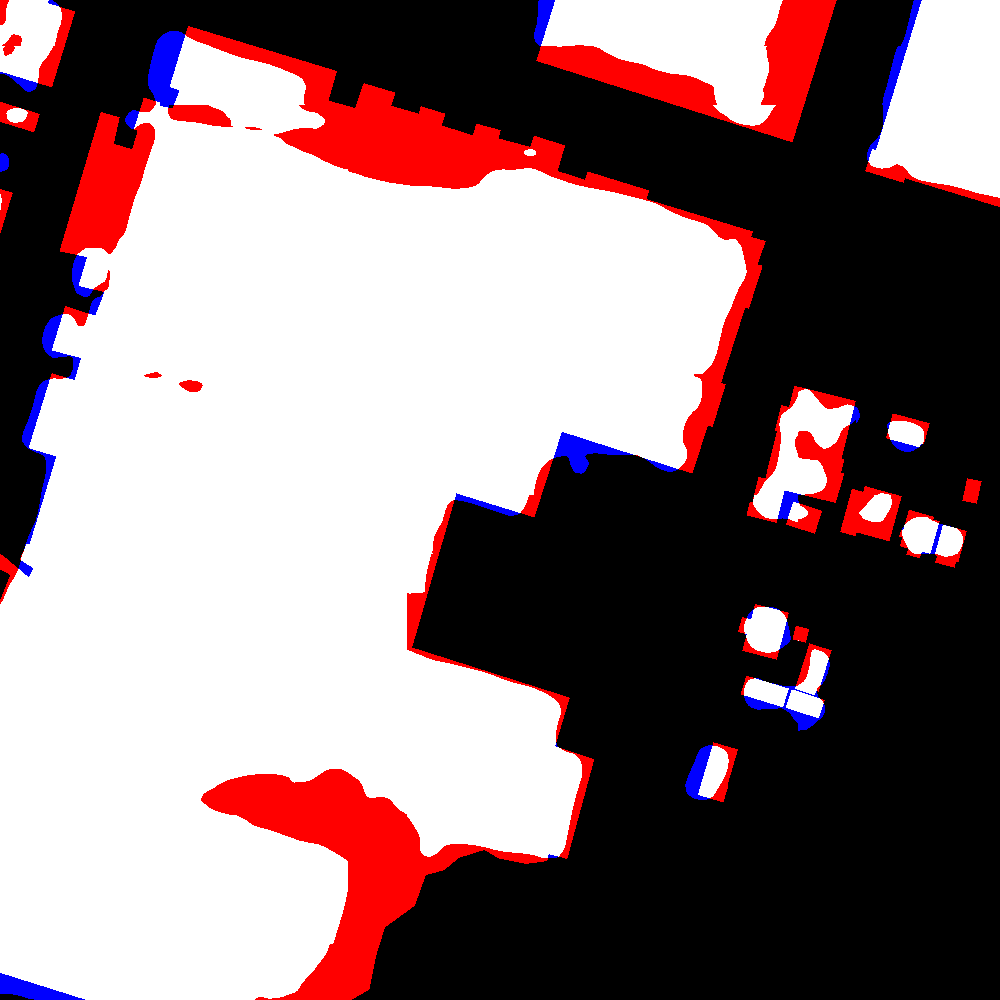}&
        \includegraphics[width=0.14\textwidth,height=2.2cm]
        {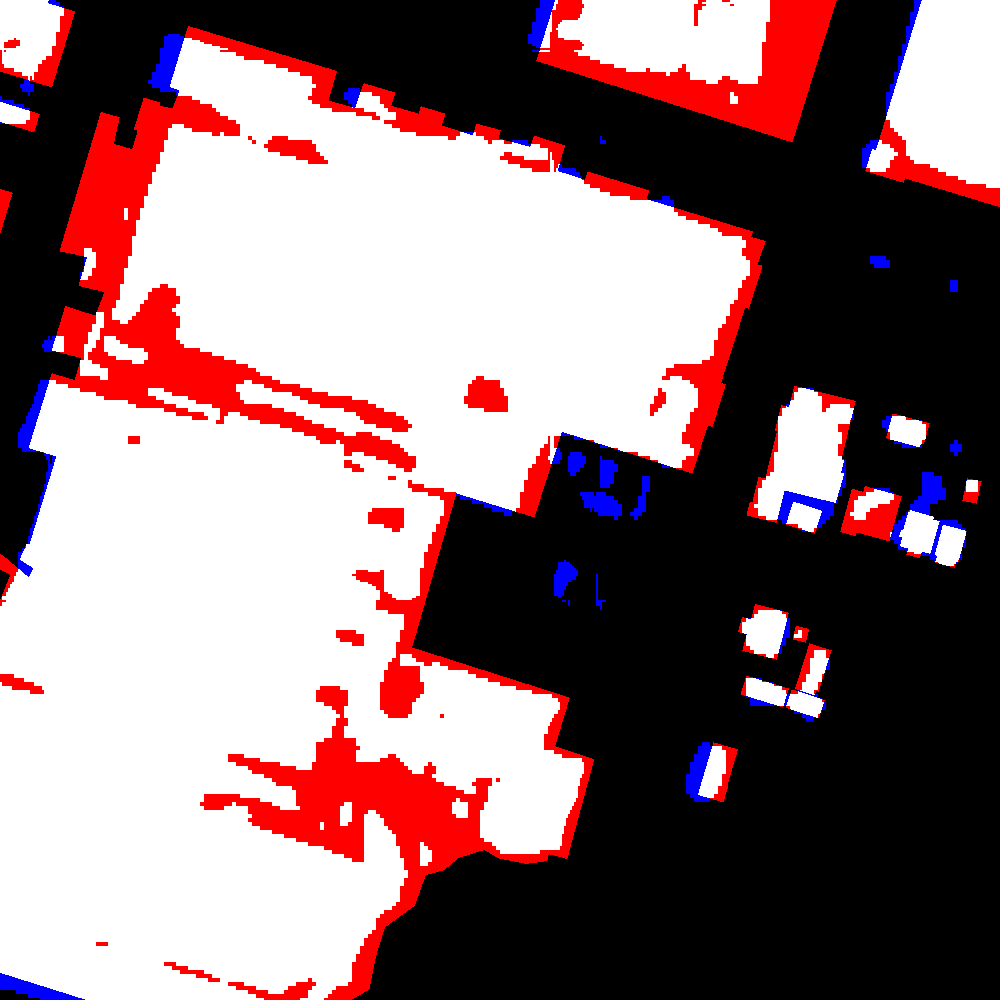}&
        \includegraphics[width=0.14\textwidth,height=2.2cm]
        {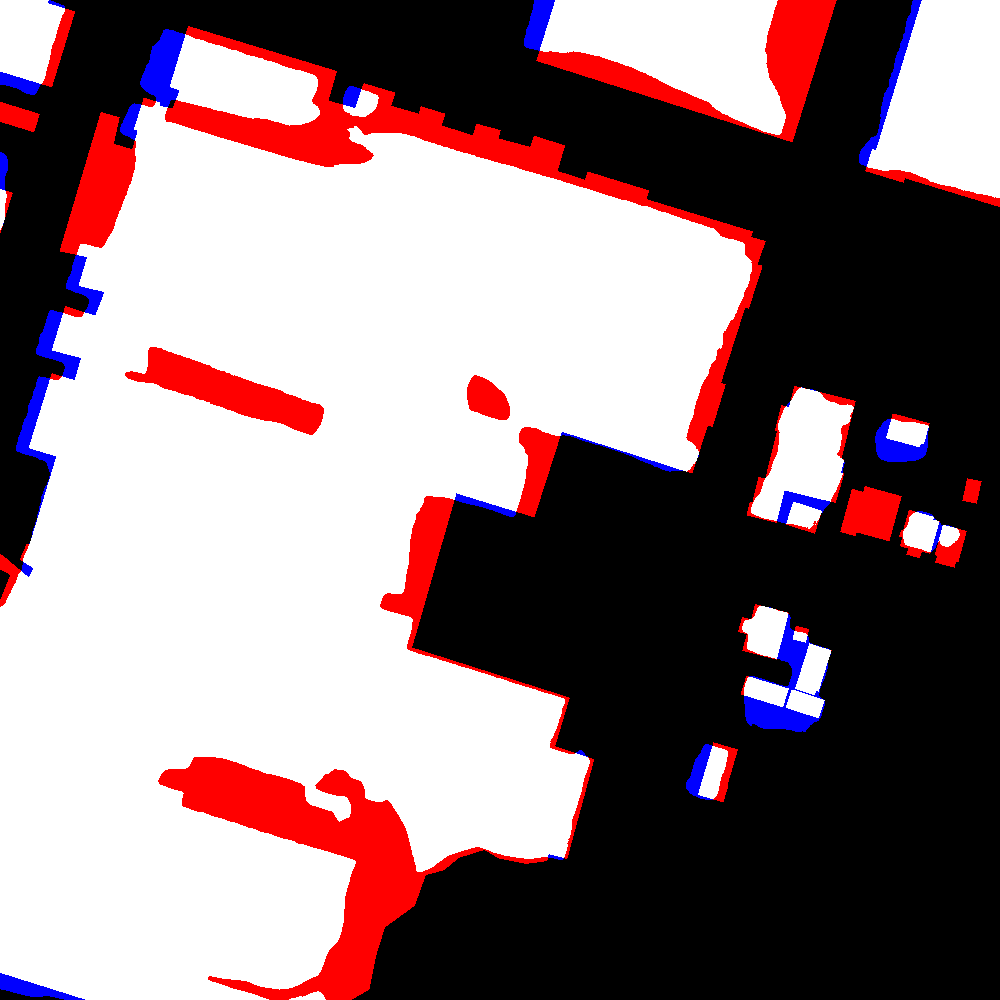}&
		\includegraphics[width=0.14\textwidth,height=2.2cm]
        {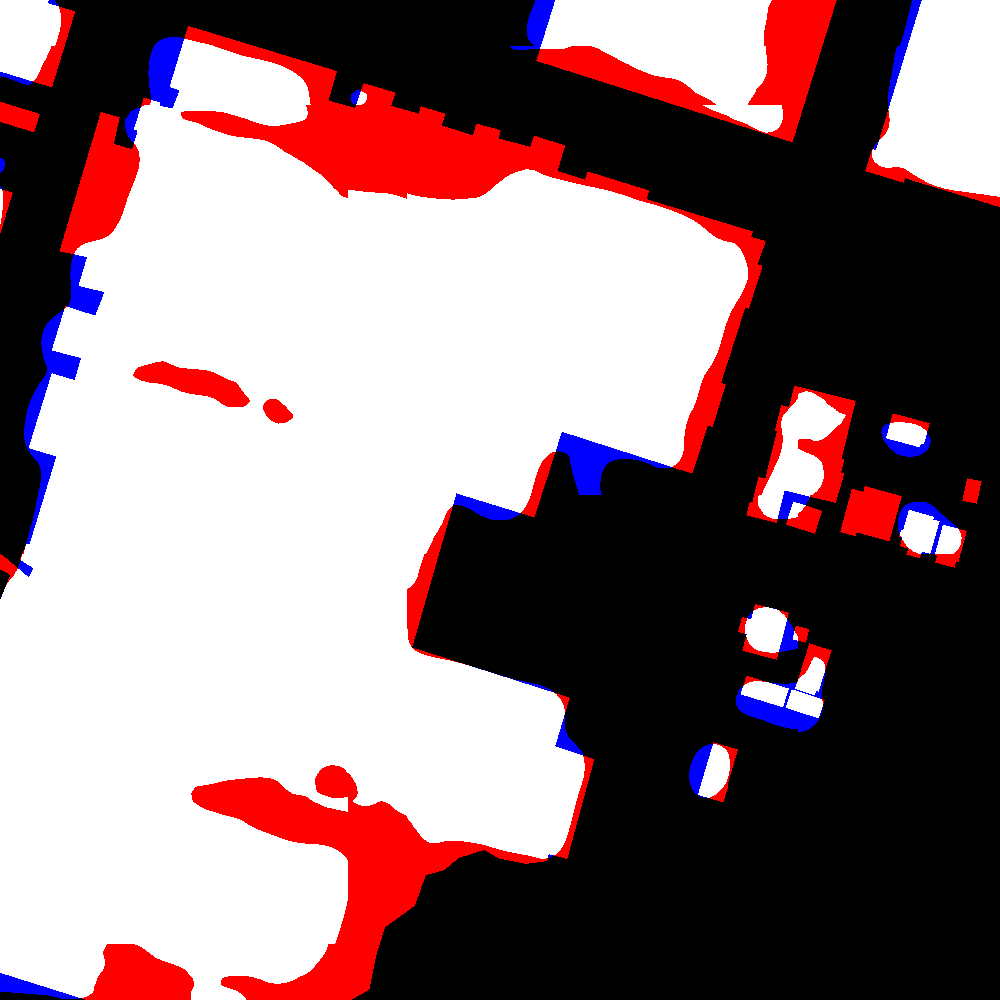}\\
        
        \includegraphics[width=0.14\textwidth,height=2.2cm]
        {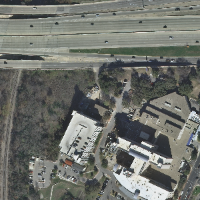} &
		\includegraphics[width=0.14\textwidth,height=2.2cm]
        {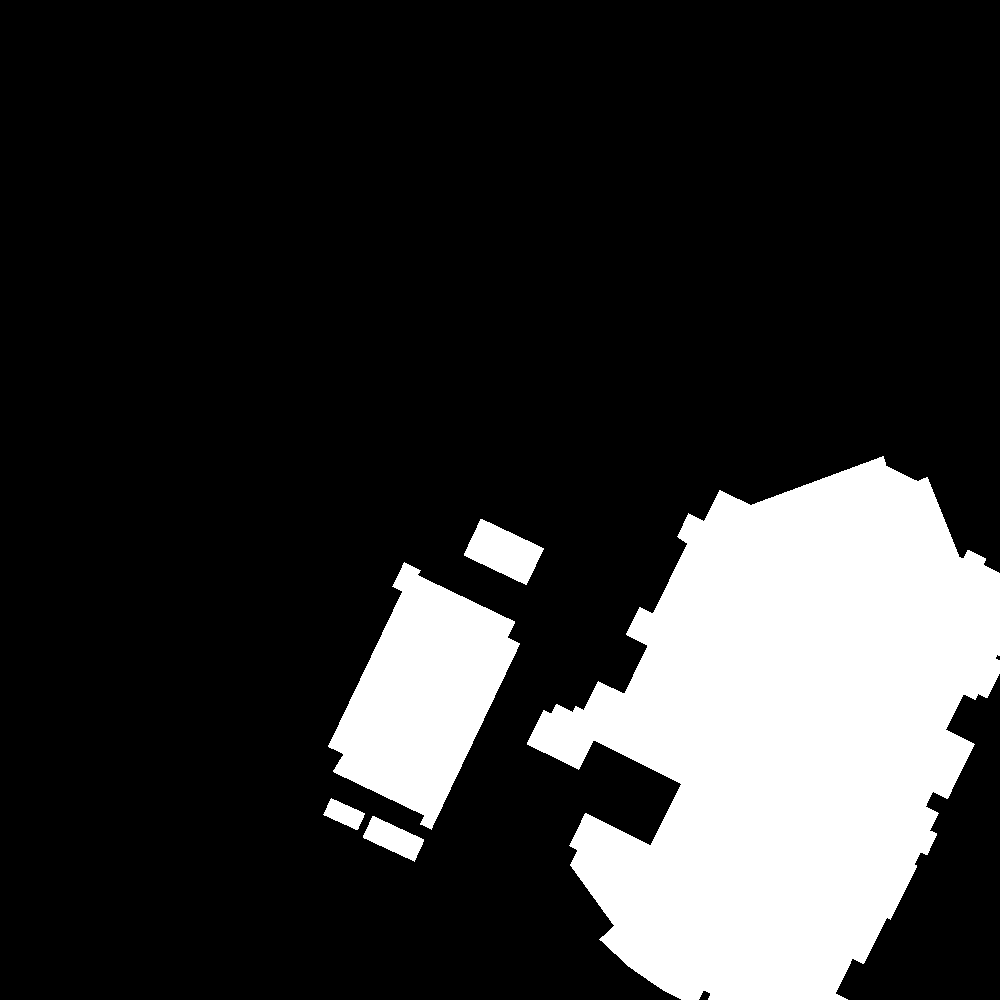}&
		\includegraphics[width=0.14\textwidth,height=2.2cm]
        {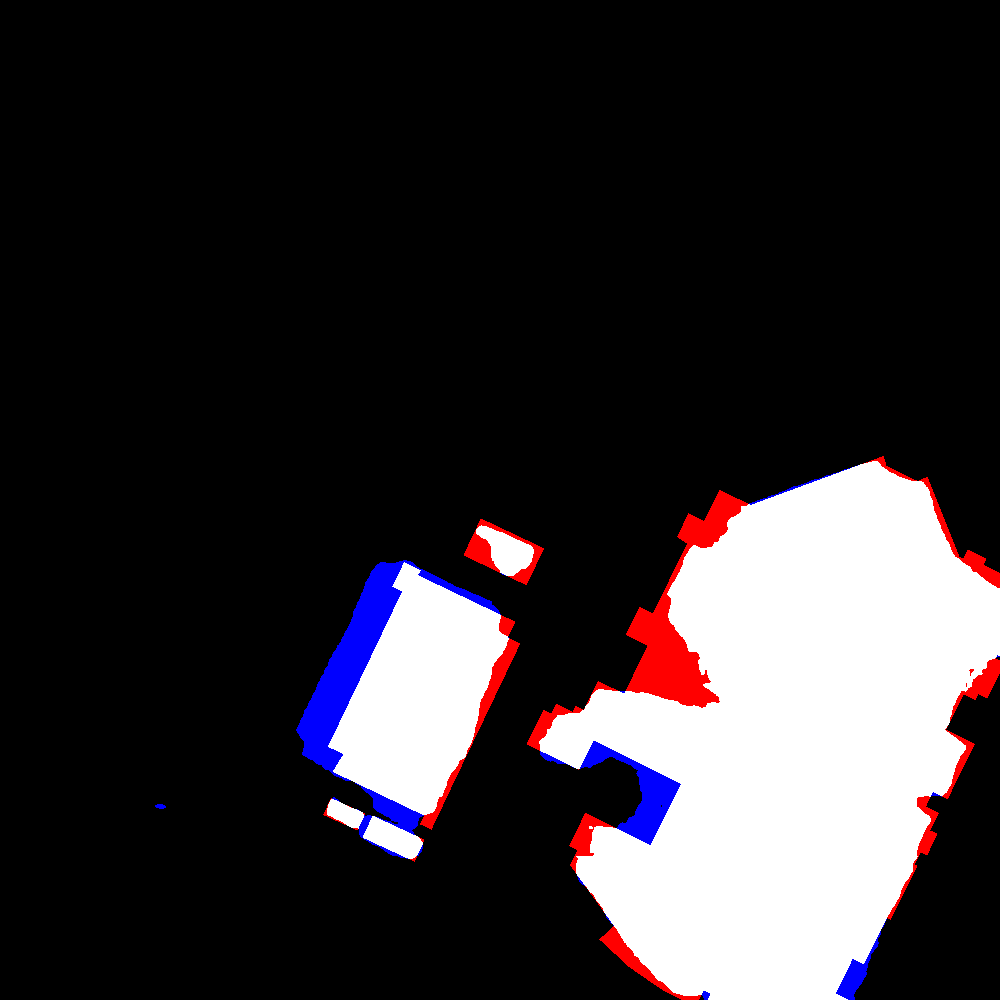}&
		\includegraphics[width=0.14\textwidth,height=2.2cm]
        {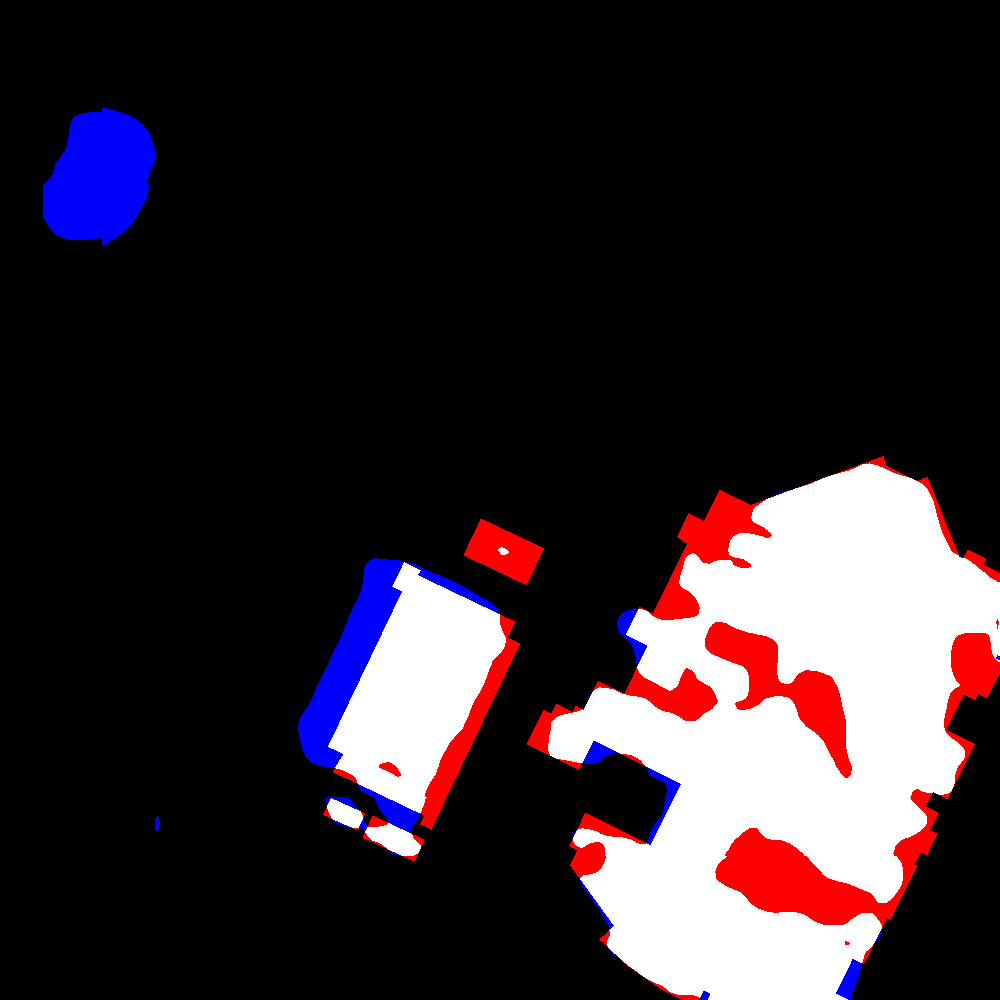}&
        \includegraphics[width=0.14\textwidth,height=2.2cm]
        {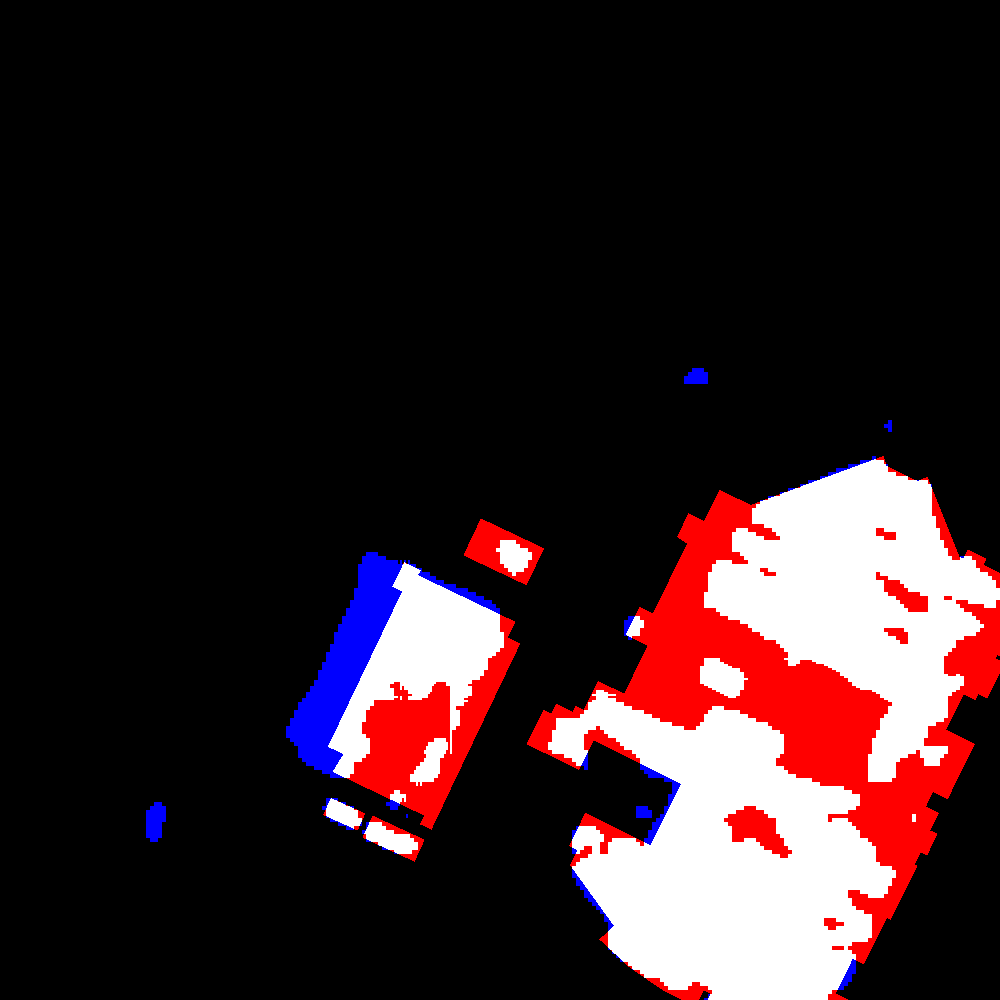}&
        \includegraphics[width=0.14\textwidth,height=2.2cm]
        {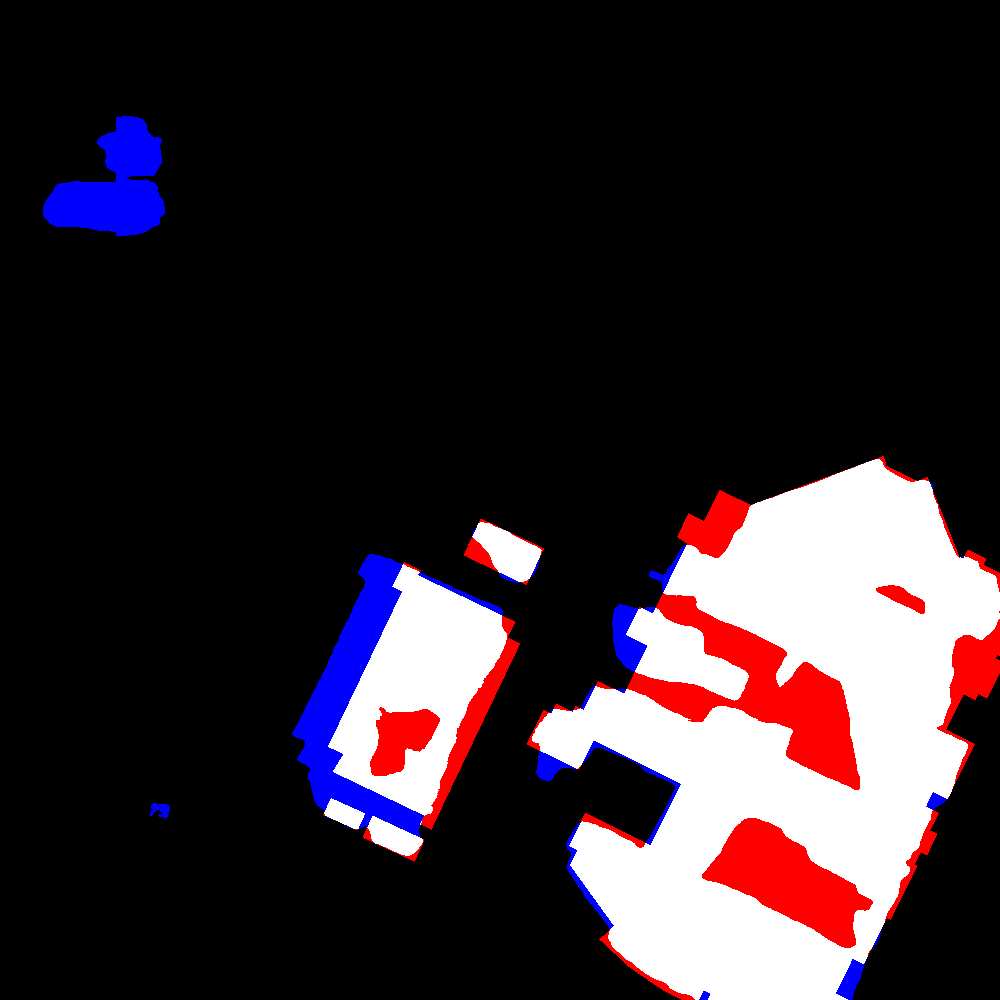}&
		\includegraphics[width=0.14\textwidth,height=2.2cm]
        {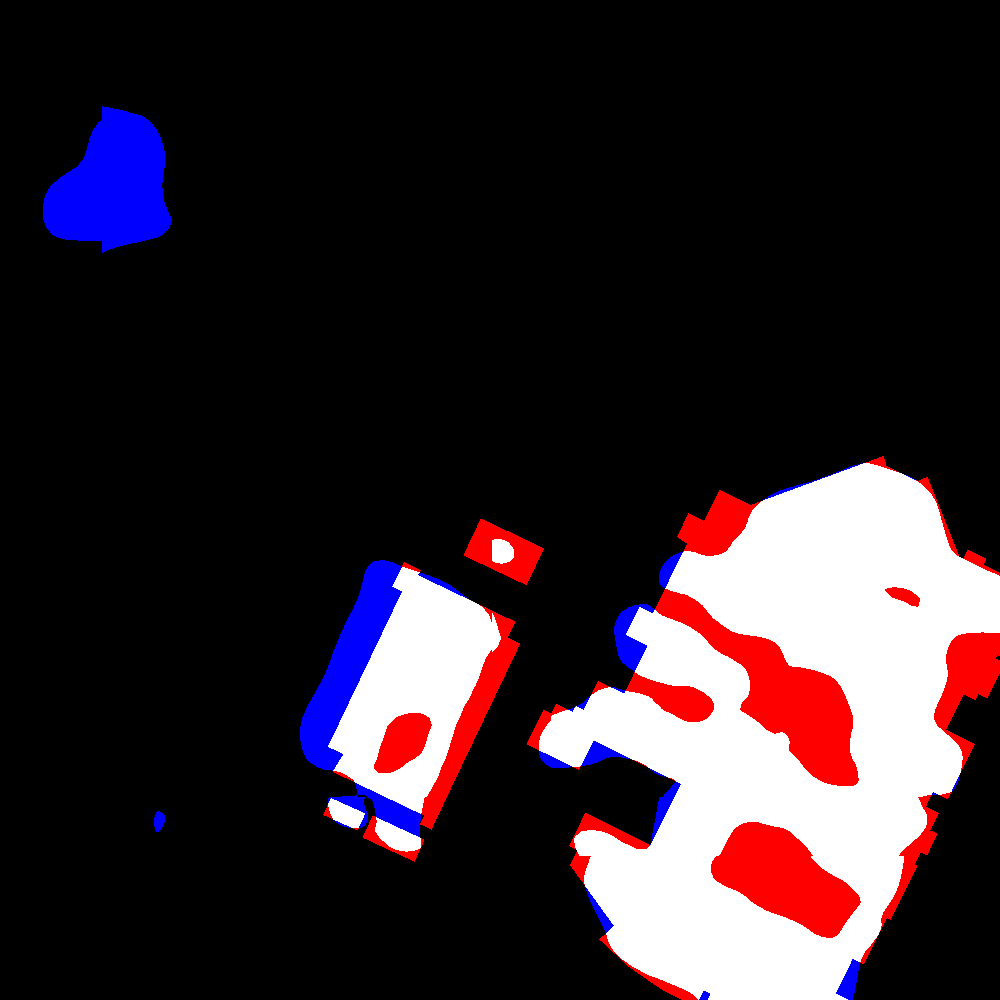}\\

		Input Image & Ground-truth & Ours & LCC \cite{li2021contexts} & GLNet \cite{chen2019collaborative} & CascadePSP \cite{cheng2020cascadepsp} & FCN-8s \cite{long2015fully}
	\end{tabular}
	\caption{We illustrate several examples of semantic segmentation in ultra-high resolution images, comparing with the state-of-the-arts. \wxb{In the first three rows of the figures, we show comparison results from DeepGlobe, in which the masks with varied colors represent different semantic regions. Particularly,
	cyan represents ``urban", yellow represents ``agriculture", purple represents ``rangeland", green represents ``forest", blue represents ``water", and white represents ``barren". In the last three rows of the figures, we show representative results from Inria Aerial, where the black regions refer to the background while the white regions represent the extracted buildings. Besides, we highlight the pixels which are the discrepancy between our estimation and ground truth. The blue and red pixels represent False Negative and False Positive, respectively.}}
	\label{fig:deepglobe}
\end{figure*}

Thus, we provide their corresponding metrics on DeepGlobe as \textit{Local Inference} and \textit{Global Inference} in Table~\ref{tab:table_deepglobe}. Specifically,
\wxb{CascadePSP, GLNet, and LCC} are specifically tailored for the task of ultra-high resolution image semantic segmentation (denoted as \textit{High-Res Model} in Table \ref{tab:table_deepglobe}). 
In particular, CascadePSP mainly aims to handle natural images, while GLNet can be applied for geospatial images.
Note that, in Table~\ref{tab:table_deepglobe}, GLNet* refers to the model without its global-local feature sharing module. \wxb{Moreover, LCC is our preliminary version, which takes advantage of locality-aware context fusion and a contextual semantics refinement network to process local image patches.}
% The segmentation performance is measured using the metric Mean Intersection-Over-Union (mIOU). 
All of the results are obtained following the same training and testing protocol.
Besides, since the original paper of CascadePSP has not reported the results on these two datasets, we train their model following the same protocol in our experiments as well. CascadePSP requires a pretrained model to provide rough global results. 

\begin{table}[tbp]
    \centering
    \footnotesize
    \setlength{\tabcolsep}{1.5pt}
    \caption{\lqq{IoU of each class on DeepGlobe. We use VGG-16 as backbone for the first two rows and MiT-B2 for the last two rows. The classes of ``rangeland" and ``barren" are smaller than other classes.} } 
    \begin{tabular}{c|cccccc|c}
        \toprule
        Model & urban & agri. & \underline{rang.} & forest & water & \underline{barren} & Mean \\
        \midrule
        FCN-8s \cite{long2015fully} & 79.2 & 87.5 & 40.1 & 77.7 & 82.1 & 64.4 & 71.8 \\
        Ours  & 80.5 & 88.4 & 43.8 & 79.2 & 84.4 & 67.3 & 73.9 \\
        \midrule
        Segformer \cite{xie2021segformer} & 80.2 & 87.9 & 47.1 & 79.5 & 84.8 & 67.0 & 74.4 \\
        Ours   & 79.9 & 87.7 & 48.2 & 78.9 & 85.0 & 68.8 & 74.8 \\
        \bottomrule
    \end{tabular}
    \label{tab:Per}
\end{table}

\begin{table}
    \centering
    \setlength{\tabcolsep}{4pt}
    \footnotesize
    \caption{Efficacy of different contexts and gated position embeddings (GPE).}
    \begin{tabular}{ccc|c|cc}
        \toprule
         & \textbf{Context} &  &  & \multicolumn{2}{c}{\textbf{mIOU}} \\
        \midrule
        Local & Medium & Large & GPE & \textit{DeepGlobe} & \textit{Inria Aerial} \\
        \midrule
          &   &  &  &71.84 & 69.08\\
        \checkmark &   &  &  & 72.12 & 72.50 \\
        & \checkmark &  & & 72.67 & 72.48\\
         &   & \checkmark & & 72.67 & 72.46\\
         &  \checkmark & \checkmark &  & 73.12 & 73.18 \\
         \checkmark & \checkmark& \checkmark &  & \textbf{73.22}  & \textbf{73.53} \\
         \midrule
         & \checkmark &  & \checkmark & 73.33 & 72.73\\
         &  \checkmark & \checkmark & \checkmark & 73.37 & 73.00 \\
        \checkmark & \checkmark& \checkmark & \checkmark & \textbf{73.47} & \textbf{73.50} \\
        
        \bottomrule
    \end{tabular}
    \label{tab:abla1}
\end{table}

As observed in Table \ref{tab:table_deepglobe} and \ref{tab:table_inria}, amongst all comparison methods, our model achieves the state-of-the-art performance comparing to the competing methods in the respective datasets. \wxb{For CNN-based backbones, our model outperforms LCC by innovating our framework, which fully demonstrates superiority of alternating local enhancement module. In addition, as the SOTA transformer based model, Segformer also achieves excellent results and is superior to our preliminary version, which only aggregates context within the local patches by using the powerful self-attention mechanism. However, Segformer does not take advantage of the contextual information outside the local patches. To prove the validity of our proposed modules for Segformer, the locality-aware context fusion module and alternating local enhancement module are incorporated with Segformer and it achieves even better performance. As a conclusion, our model is effective and achieves the optimal performance for both CNN-based and transformer backbone in the public datasets.}
\lqq{Our task often suffers from severe class imbalance problem, e.g., the category ``agriculture" occupies much more area (i.e. pixels) than ``rangeland". The pixel-wise metrics accuracy can hardly reflect how the models handle this problem. Instead, mIOU and F1 measure the average segmentation quality of each category. In Table.~\ref{tab:Per}, we compare our model against baselines for each class. The classes of ``rangeland" and ``barren" occupy much less pixels than the other classes, so their IoUs are generally lower than the other classes. Observed from the results, our model is able to elevate the performance on these two small classes. For VGG-16 backbone, our model achieves 3.7\% and 3.9\% improvements, while our transformer-based model also gains 1.1\% and 1.8\% boost.
% Thus, the major improvements on mIOU (more than $\sim2\%$) and F1 ($\sim1.5\%$) indicate the effectiveness of our model.
}

In addition, we show several qualitative comparison results in Fig. \ref{fig:deepglobe}. As observed, our model is able to identify strip-shaped regions (e.g., river) and \wxb{large regions (e.g., agriculture)}, which is benefited from the correlation between local and contexts. \wxb{Compared to the preliminary versions, it is clear that our results are more detailed and much closer to the ground truth in both large and small regions. Moreover, for both small and large buildings, our predictions are more finely segmented and less noisy than our preliminary version and other competing methods.}

% \subsection{Visualization of Contextual Attention}

\begin{figure}[t]
    \centering
    \tiny
	\begin{tabular}{c@{}c@{}c@{}c@{}c}
        
		\includegraphics[width=0.095\textwidth,height=1.5cm]
        {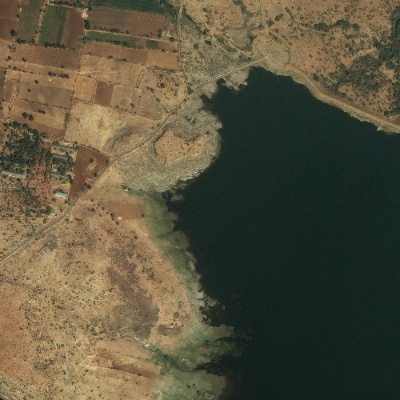} &
		\includegraphics[width=0.095\textwidth,height=1.5cm]
        {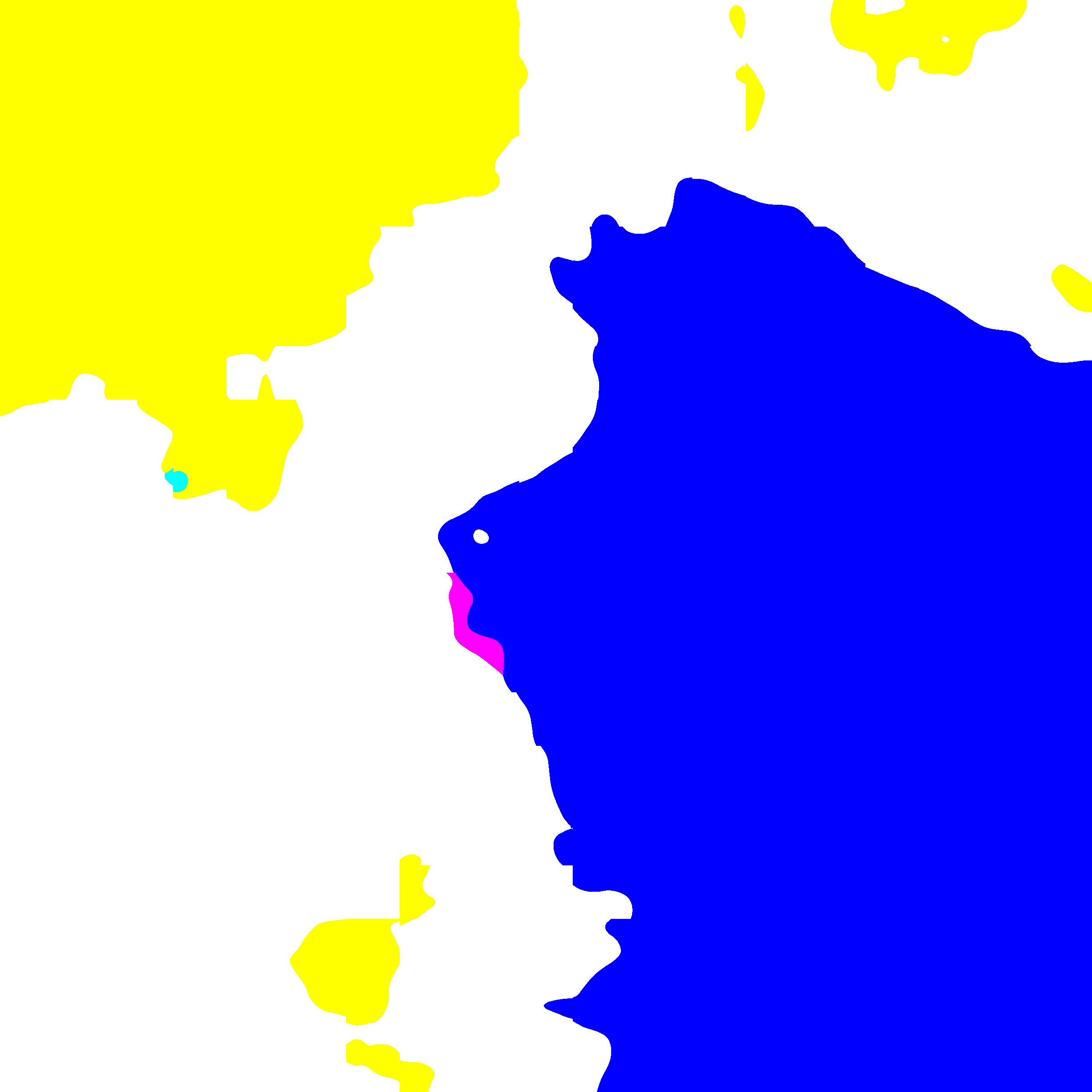}&
        \includegraphics[width=0.095\textwidth,height=1.5cm]
        {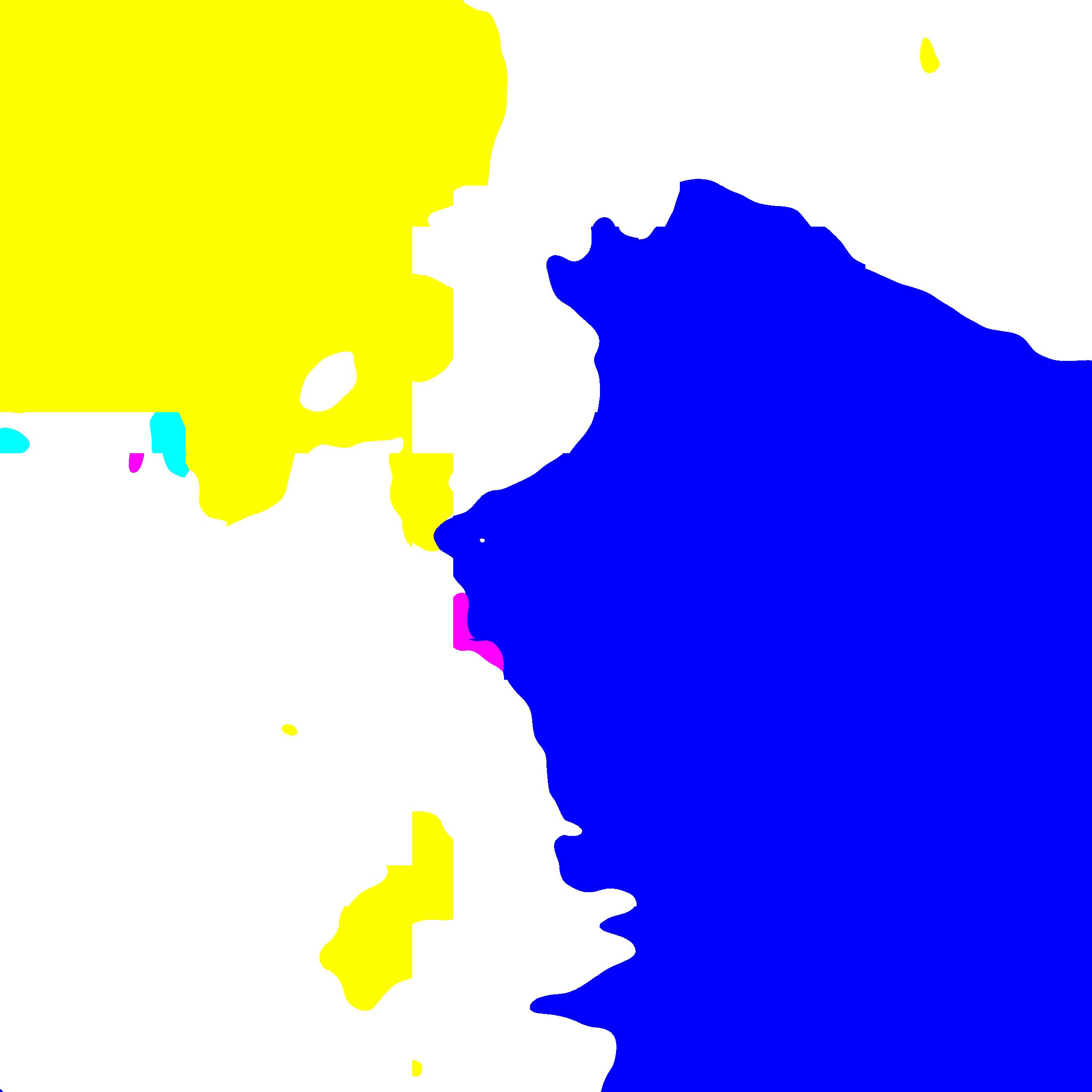}&
        \includegraphics[width=0.095\textwidth,height=1.5cm]
        {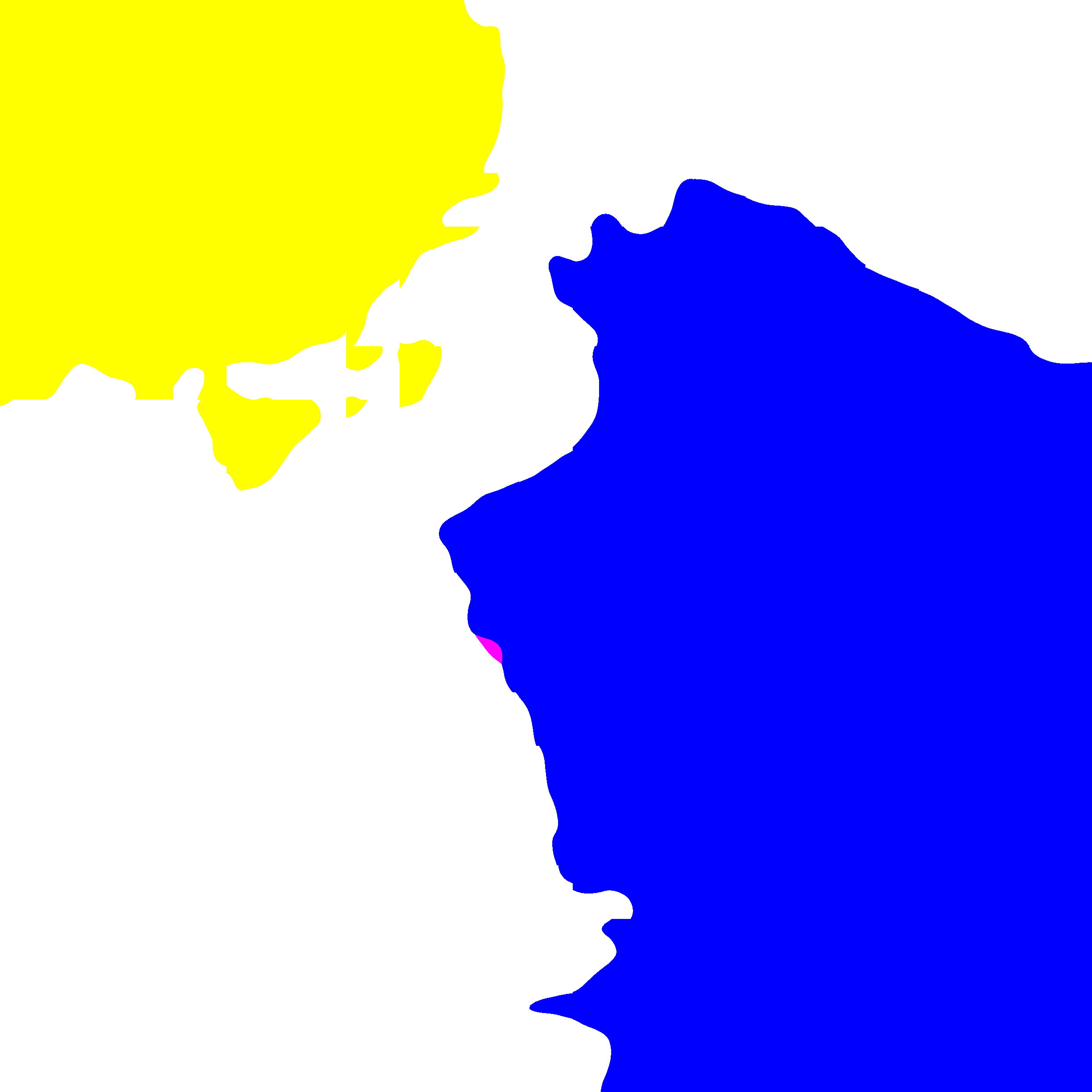}&
        \includegraphics[width=0.095\textwidth,height=1.5cm]
        {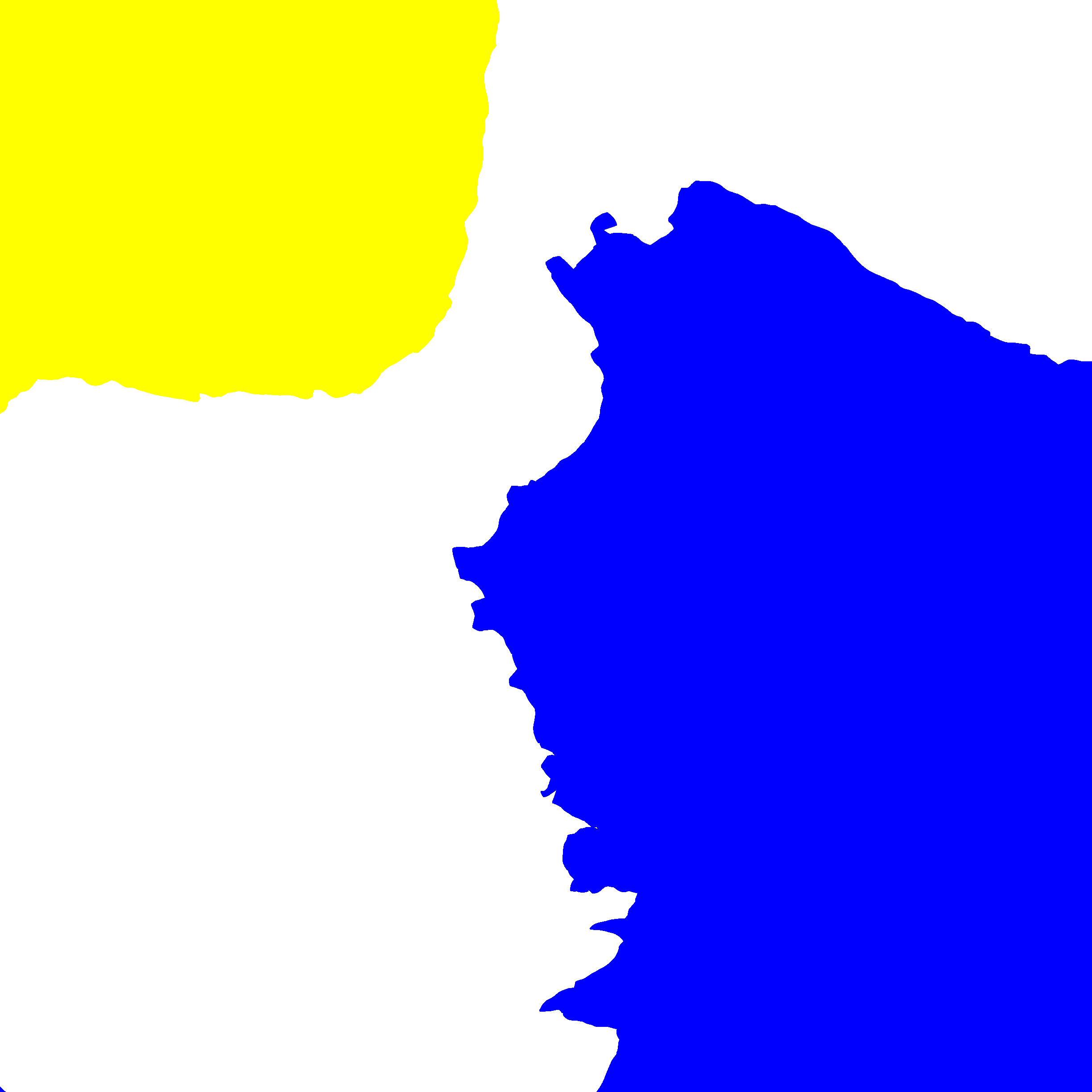}\\
        
		\includegraphics[width=0.095\textwidth,height=1.5cm]
        {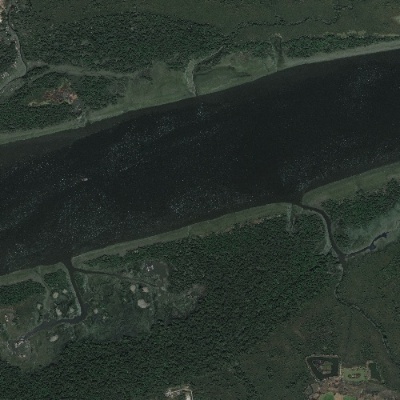} &
		\includegraphics[width=0.095\textwidth,height=1.5cm]
        {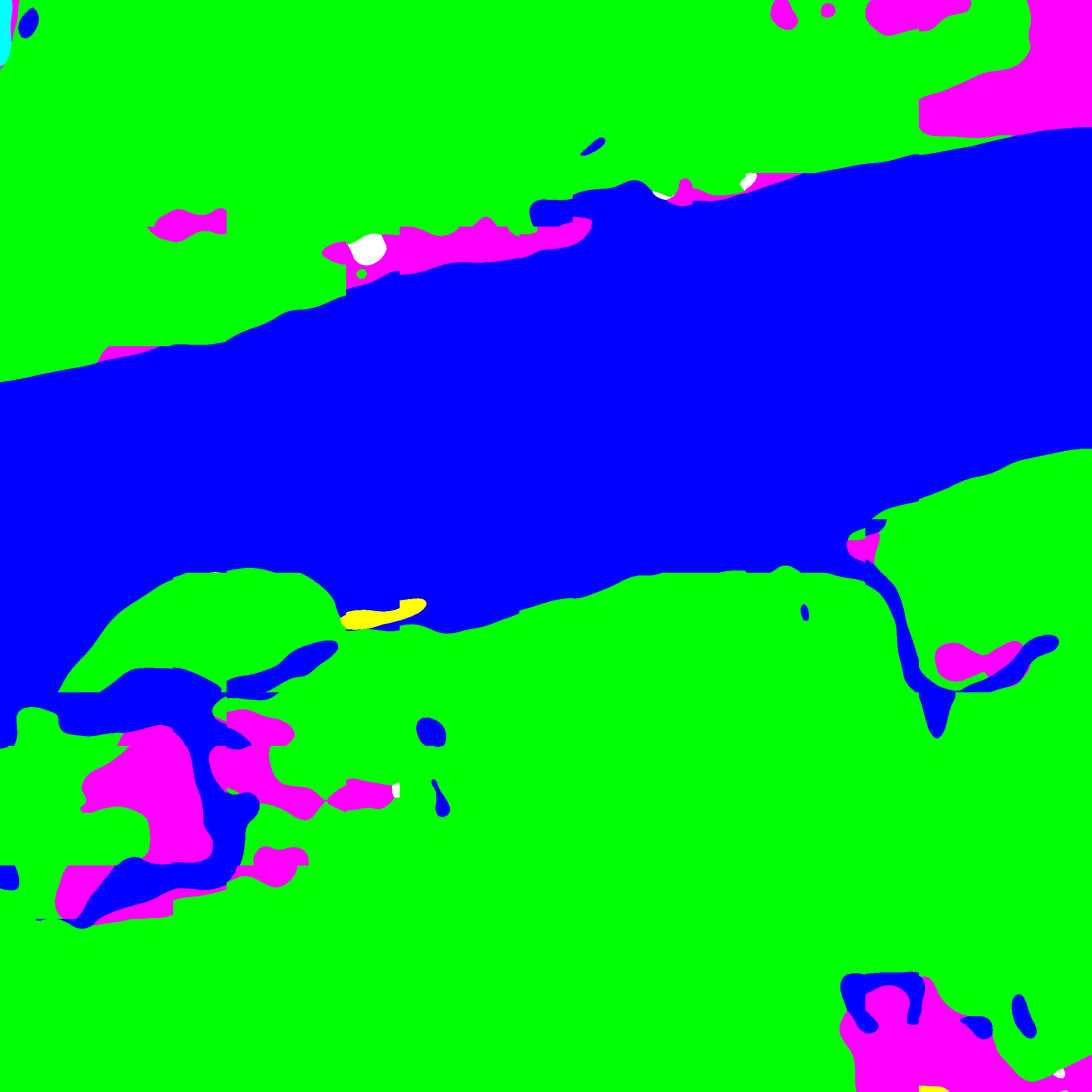}&
        \includegraphics[width=0.095\textwidth,height=1.5cm]
        {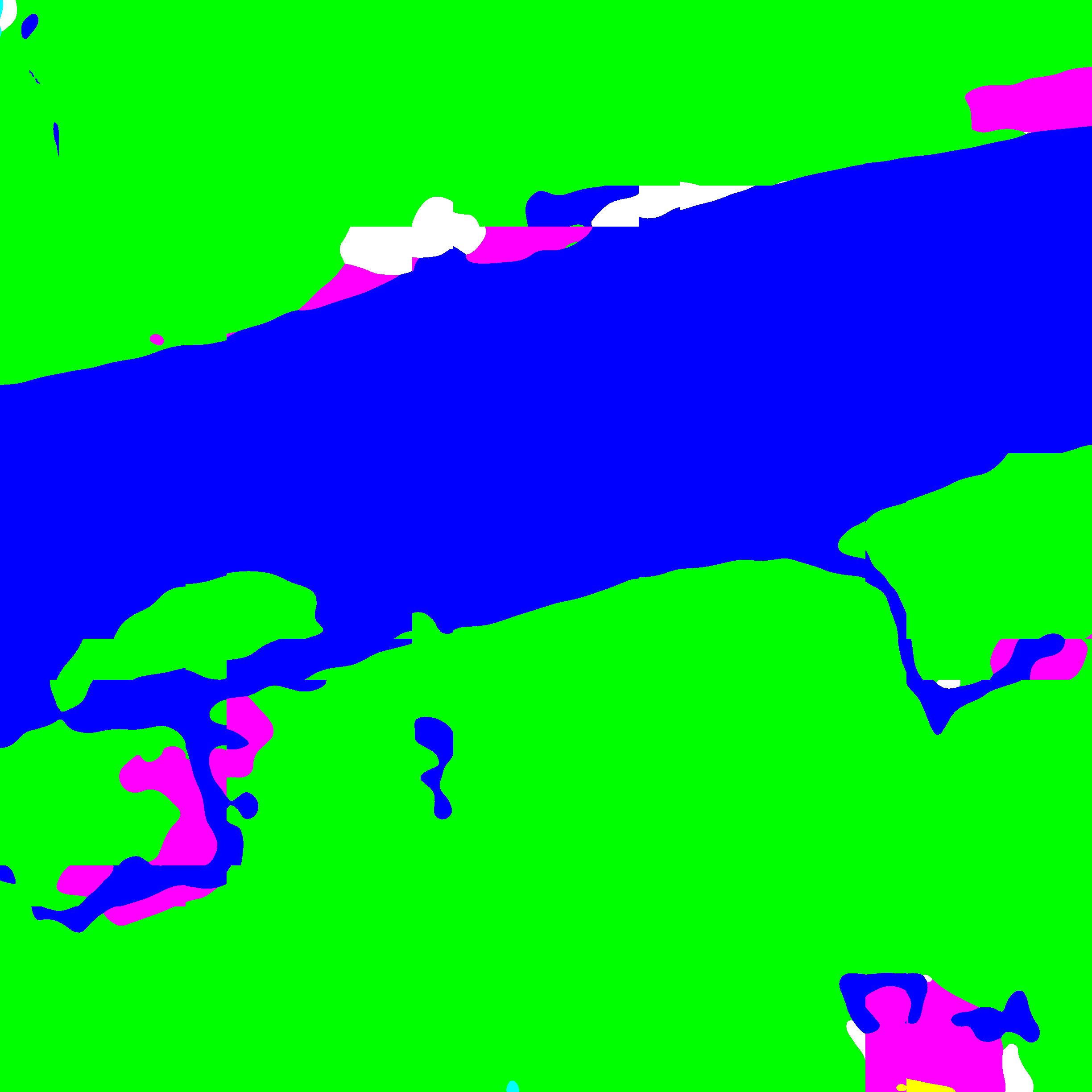}&
        \includegraphics[width=0.095\textwidth,height=1.5cm]
        {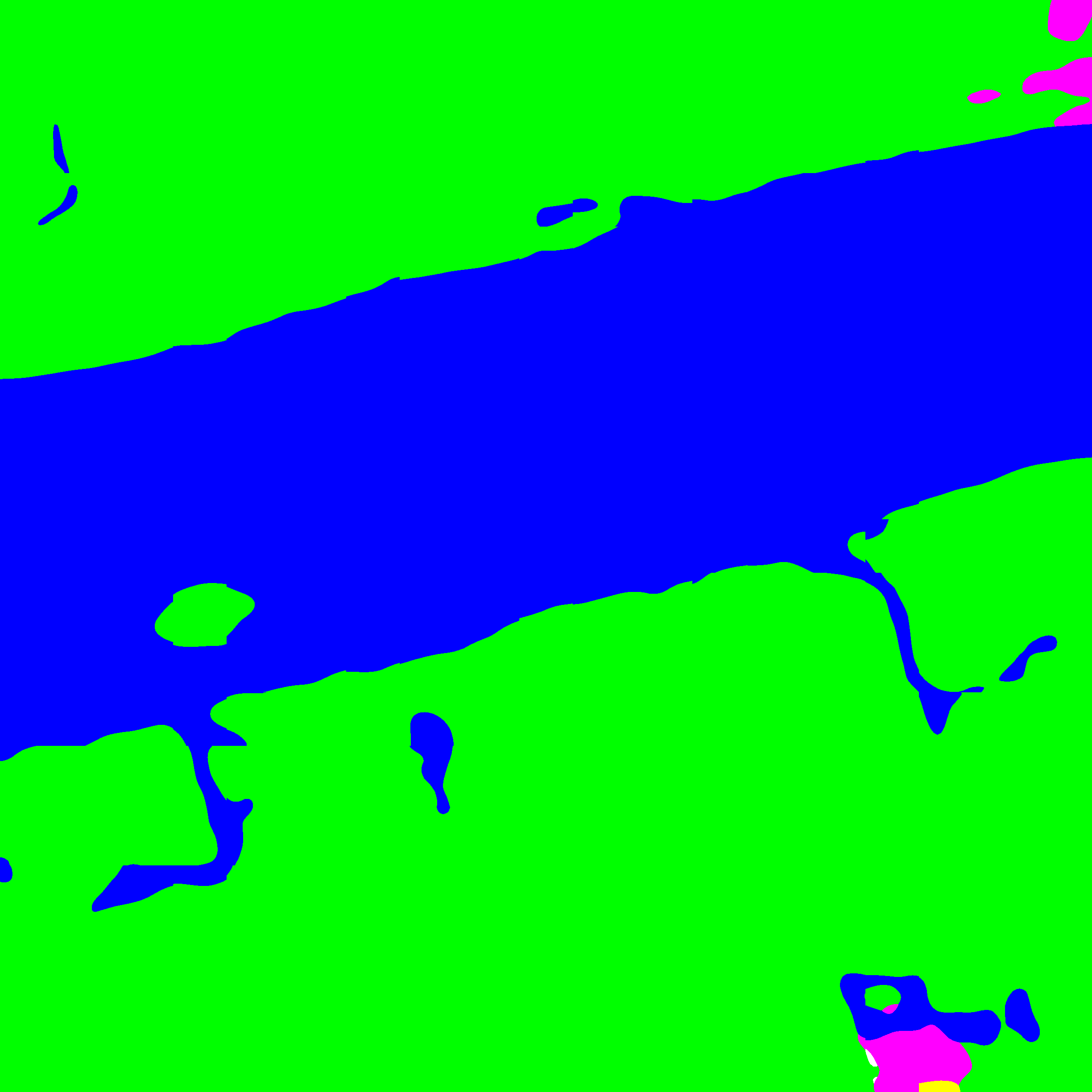}&
        \includegraphics[width=0.095\textwidth,height=1.5cm]
        {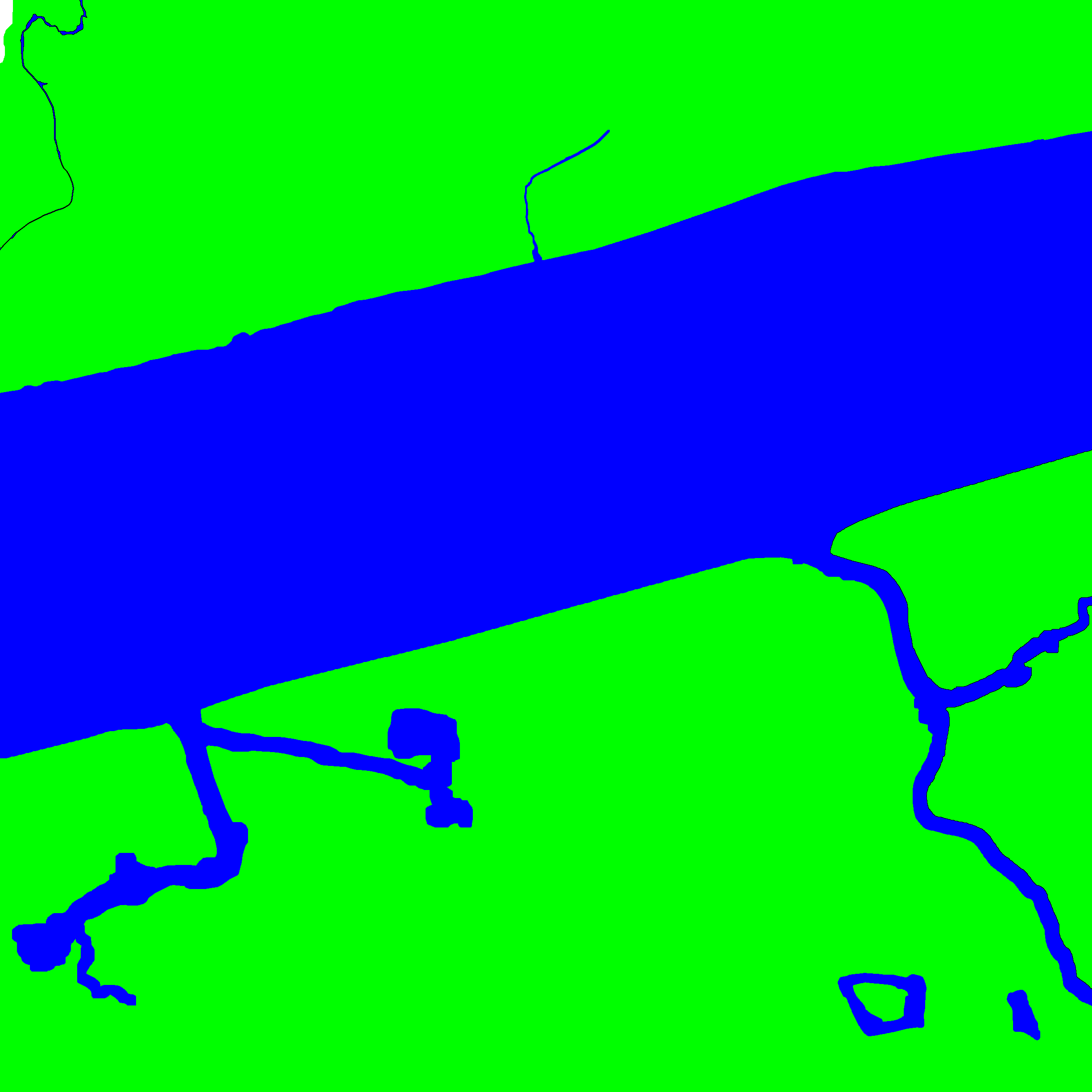}\\

        \includegraphics[width=0.095\textwidth,height=1.5cm]
        {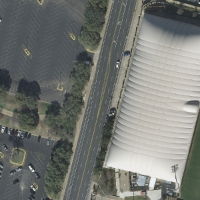} &
		\includegraphics[width=0.095\textwidth,height=1.5cm]
        {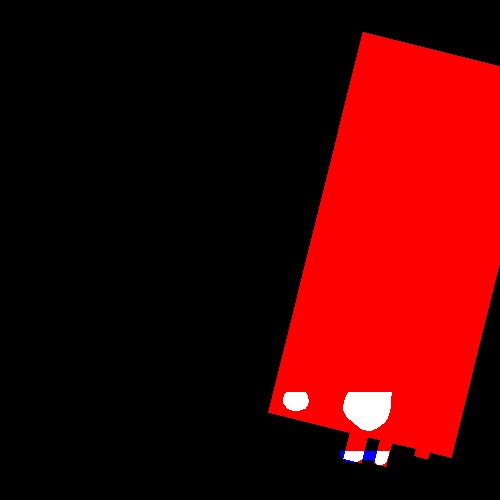}&
        \includegraphics[width=0.095\textwidth,height=1.5cm]
        {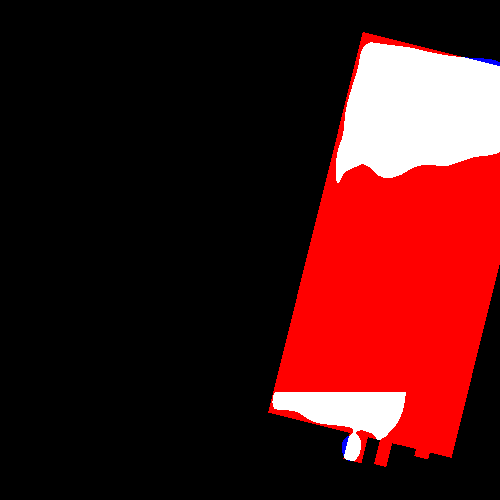}&
        \includegraphics[width=0.095\textwidth,height=1.5cm]
        {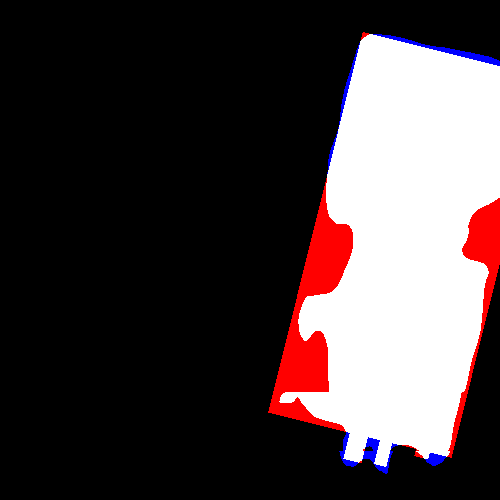}&
        \includegraphics[width=0.095\textwidth,height=1.5cm]
        {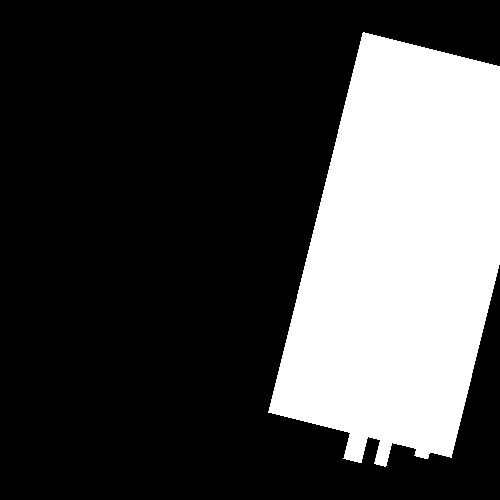}\\
        
%         \includegraphics[width=0.095\textwidth,height=1.5cm]
%         {imgs/ablation/IA/inp_chicago14_mask.png} &
% 		\includegraphics[width=0.095\textwidth,height=1.5cm]
%         {imgs/ablation/IA/fcn_local_chicago14_mask.png}&
%         \includegraphics[width=0.095\textwidth,height=1.5cm]
%         {imgs/ablation/IA/PE/chicago14.png}&
%         \includegraphics[width=0.095\textwidth,height=1.5cm]
%         {imgs/ablation/IA/prediction73_chicago14_mask.png}&
%         \includegraphics[width=0.095\textwidth,height=1.5cm]
%         {imgs/ablation/IA/gt_chicago14_mask.png}\\

        \includegraphics[width=0.095\textwidth,height=1.5cm]
        {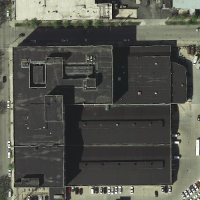} &
		\includegraphics[width=0.095\textwidth,height=1.5cm]
        {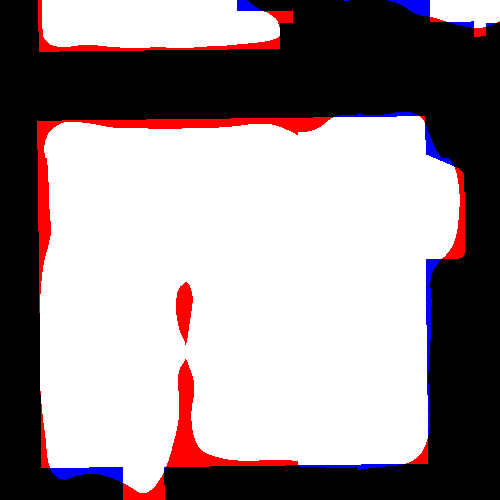}&
        \includegraphics[width=0.095\textwidth,height=1.5cm]
        {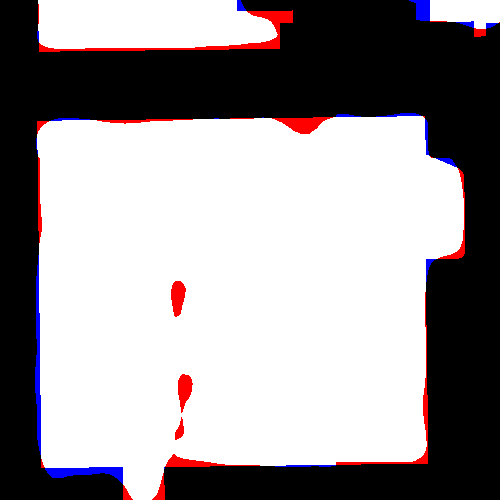}&
        \includegraphics[width=0.095\textwidth,height=1.5cm]
        {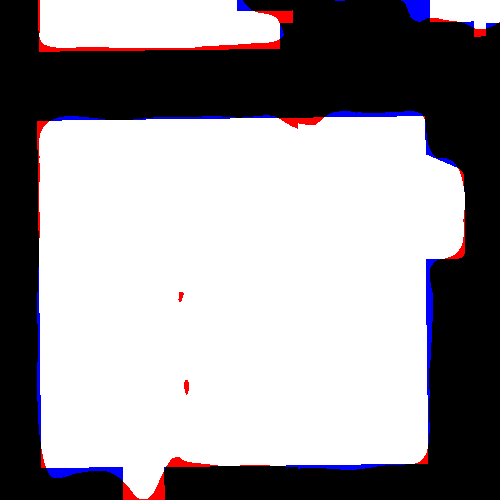}&
        \includegraphics[width=0.095\textwidth,height=1.5cm]
        {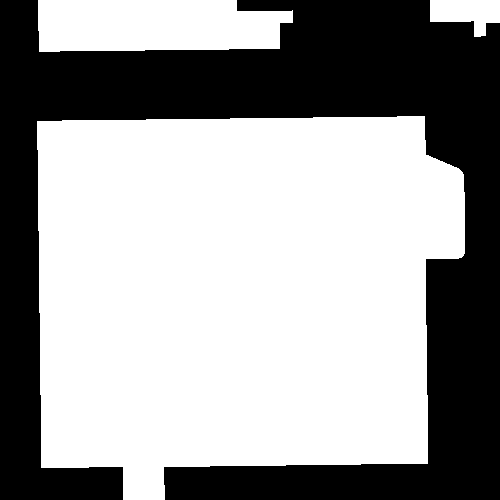}\\

		Input Image &  w/o Contexts & w/o GPE &  w/ GPE & Ground truth
	\end{tabular}
	\caption{Examples show the efficacy of contexts and gated positional embeddings (GPE). Please zoom-in for better view.}
% 	\vspace{-.4cm}
	\label{fig:abla_context}
\end{figure}

\begin{figure}[t]
    \centering
    \footnotesize
	\begin{tabular}{c@{}c@{}c}
        
        \includegraphics[width=0.15\textwidth,height=2.4cm]
        {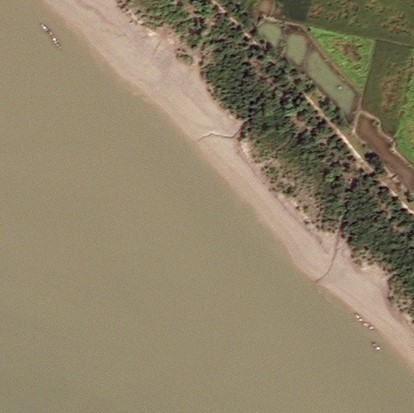} &
		\includegraphics[width=0.15\textwidth,height=2.4cm]
        {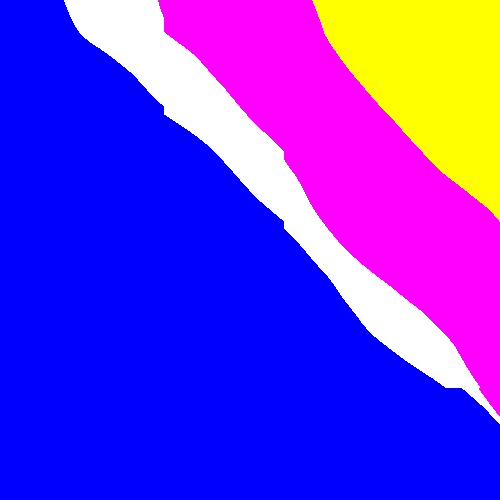}&
		\includegraphics[width=0.15\textwidth,height=2.4cm]
        {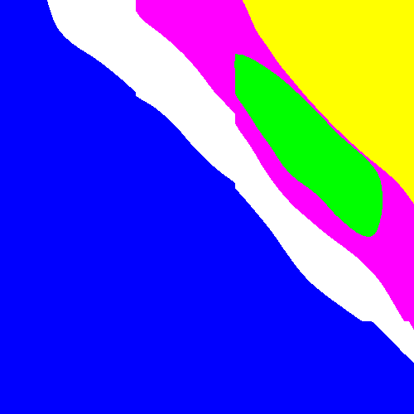}\\
        Local Patch & w/ Contexts & w/o Contexts \\
        \includegraphics[width=0.15\textwidth,height=2.4cm]
        {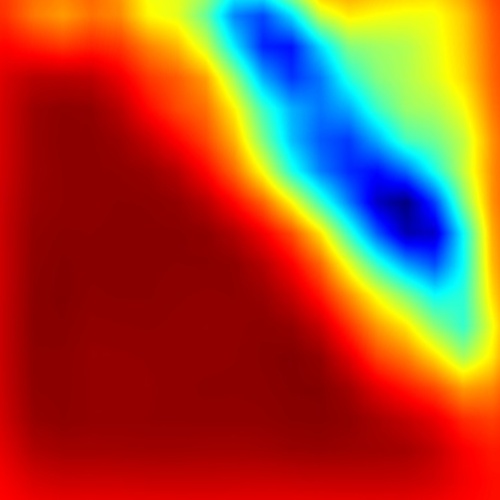}&
        \includegraphics[width=0.15\textwidth,height=2.4cm]
        {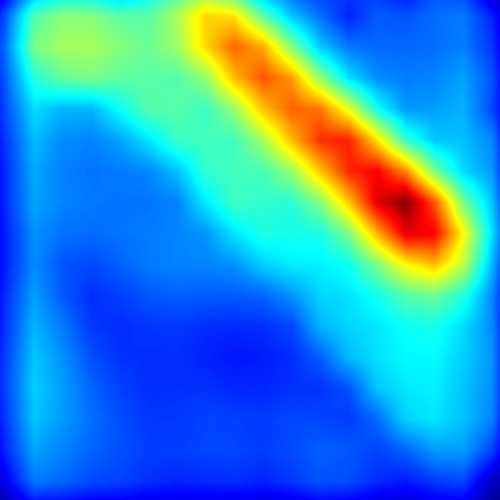}&
		\includegraphics[width=0.15\textwidth,height=2.4cm]
        {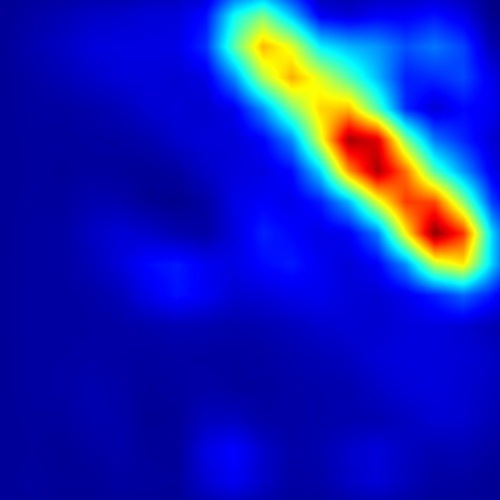}\\
        & \textit{Contextual Attention}& \\
        \includegraphics[width=0.15\textwidth,height=2.4cm]
        {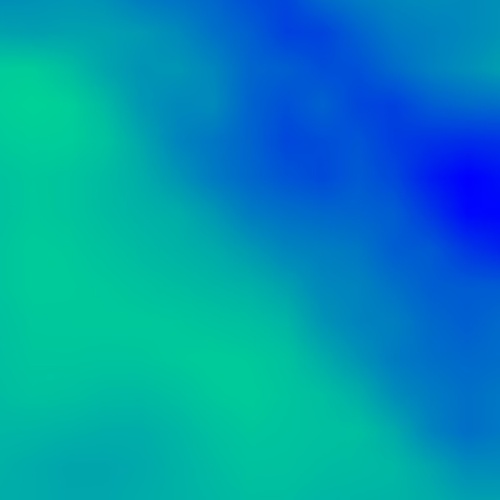}&
        \includegraphics[width=0.15\textwidth,height=2.4cm]
        {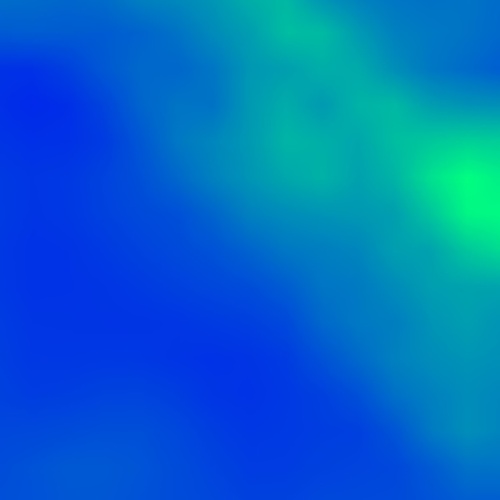}&
		\includegraphics[width=0.15\textwidth,height=2.4cm]
        {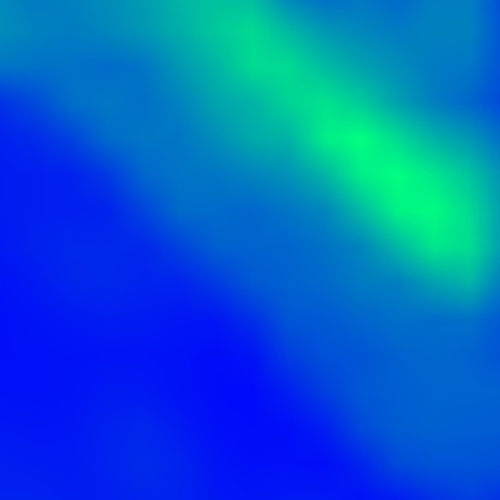}\\
        & \textit{Weight Maps}& \\
        Local Context & Medium Context & Large Context\\
	\end{tabular}
	\caption{Illustration of an example with and without using contexts, as well as their contextual attention maps that imply the correlation between contexts and local patch. Besides, we also visualize the corresponding estimated weight maps for fusing contexts.}
	\label{fig:vis3}
\end{figure}

% \begin{figure}[t]
%     \centering
%     \footnotesize
% 	\begin{tabular}{c@{}c@{}c}
        
%         \includegraphics[width=0.15\textwidth,height=2.4cm]
%         {imgs/vis/old_vis/343215.jpg} &
% 		\includegraphics[width=0.15\textwidth,height=2.4cm]
%         {imgs/vis/new_vis/343215_mask.png}&
% 		\includegraphics[width=0.15\textwidth,height=2.4cm]
%         {imgs/vis/old_vis/wo343215.png}\\
%         Local Patch & w/ Contexts & w/o Contexts \\
%         \includegraphics[width=0.15\textwidth,height=2.4cm]
%         {imgs/vis/new_vis/crop-343215local.jpg}&
%         \includegraphics[width=0.15\textwidth,height=2.4cm]
%         {imgs/vis/new_vis/crop-343215medium.jpg}&
% 		\includegraphics[width=0.15\textwidth,height=2.4cm]
%         {imgs/vis/new_vis/crop-343215large.jpg}\\
%         & \textit{Contextual Attention}& \\
%         \includegraphics[width=0.15\textwidth,height=2.4cm]
%         {imgs/vis/new_vis/343215local_g.jpg}&
%         \includegraphics[width=0.15\textwidth,height=2.4cm]
%         {imgs/vis/new_vis/343215medium_g.jpg}&
% 		\includegraphics[width=0.15\textwidth,height=2.4cm]
%         {imgs/vis/new_vis/343215large_g.jpg}\\
%         & \textit{Weight Maps}& \\
%         Local Context & Medium Context & Large Context\\
% 	\end{tabular}
% 	\caption{Illustration of an example with and without using contexts, as well as their contextual attention maps that imply the correlation between contexts and local patch and the corresponding estimated weight maps.}
% 	\label{fig:vis4}
% \end{figure}

\begin{figure}[t]
    \centering
    \footnotesize
	\begin{tabular}{c@{}c@{}c}
    
        \includegraphics[width=0.15\textwidth,height=2.4cm]
        {imgs/vis/old_vis/7892.jpg} &
		\includegraphics[width=0.15\textwidth,height=2.4cm]
        {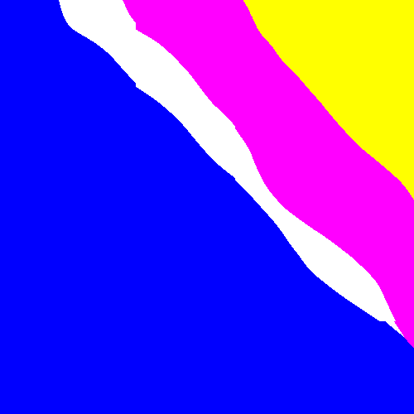}&
		\includegraphics[width=0.15\textwidth,height=2.4cm]
        {imgs/vis/old_vis/wo7892.png}\\
        Local Patch & w/ Contexts & w/o Contexts \\
        \includegraphics[width=0.15\textwidth,height=2.4cm]
        {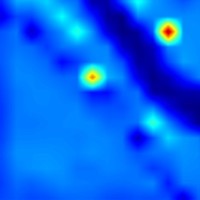}&
        \includegraphics[width=0.15\textwidth,height=2.4cm]
        {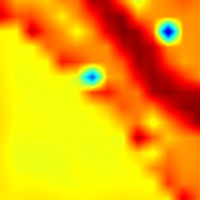}&
		\includegraphics[width=0.15\textwidth,height=2.4cm]
        {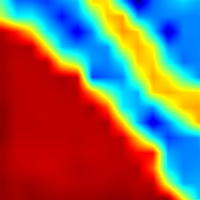}\\
        & \textit{Contextual Attention}& \\
        \includegraphics[width=0.15\textwidth,height=2.4cm]
        {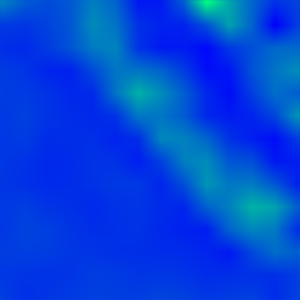}&
        \includegraphics[width=0.15\textwidth,height=2.4cm]
        {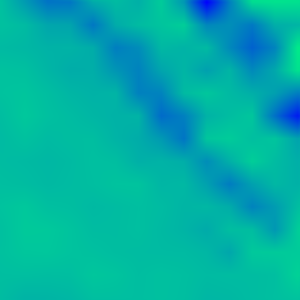}&
		\includegraphics[width=0.15\textwidth,height=2.4cm]
        {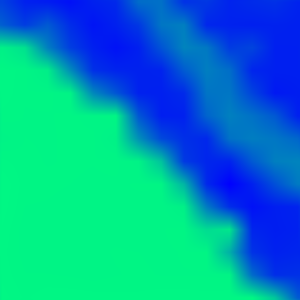}\\
        & \textit{Weight Maps}& \\
        Local Context & Medium Context & Large Context\\
	\end{tabular}
	\caption{Illustration of an example with and without using contexts, as well as their contextual attention maps and the corresponding estimated weight maps for fusion. In contrast to Fig.~\ref{fig:vis3}, the attention and weight maps are generated by the model \textbf{w/o gated positional embeddings}. }
	\label{fig:vis1}
\end{figure}

% \begin{figure}[t]
%     \centering
%     \footnotesize
% 	\begin{tabular}{c@{}c@{}c}
        
%         \includegraphics[width=0.15\textwidth,height=2.4cm]
%         {imgs/vis/old_vis/343215.jpg} &
% 		\includegraphics[width=0.15\textwidth,height=2.4cm]
%         {imgs/vis/old_vis/w343215.png}&
% 		\includegraphics[width=0.15\textwidth,height=2.4cm]
%         {imgs/vis/old_vis/wo343215.png}\\
%         Local Patch & w/ Contexts & w/o Contexts \\
%         \includegraphics[width=0.15\textwidth,height=2.4cm]
%         {imgs/vis/old_vis/343215local.jpg}&
%         \includegraphics[width=0.15\textwidth,height=2.4cm]
%         {imgs/vis/old_vis/343215medium.jpg}&
% 		\includegraphics[width=0.15\textwidth,height=2.4cm]
%         {imgs/vis/old_vis/343215large.jpg}\\
%         & \textit{Contextual Attention}& \\
%         \includegraphics[width=0.15\textwidth,height=2.4cm]
%         {imgs/vis/old_vis/343215glocal.png}&
%         \includegraphics[width=0.15\textwidth,height=2.4cm]
%         {imgs/vis/old_vis/343215gmedium.png}&
% 		\includegraphics[width=0.15\textwidth,height=2.4cm]
%         {imgs/vis/old_vis/343215glarge.png}\\
%         & \textit{Weight Maps}& \\
%         Local Context & Medium Context & Large Context\\
% 	\end{tabular}
% 	\caption{Illustration of an example with and without using contexts, as well as their contextual attention maps \wxb{without positional embeddings} that imply the correlation between contexts and local patch and the corresponding estimated weight maps.}
% 	\label{fig:vis2}
% \end{figure}

\subsection{Ablation Study}
\label{sec:abla}

In this section, we delve into the modules and settings of our proposed model and demonstrate their effectiveness. 

% \noindent \textbf{Efficacy of contexts.} 

\subsubsection{Efficacy of Contexts}

\wx{First of all, we would like to emphasize the importance of context for producing high-quality local segmentation results. To do so, we compare the variants of our model with different combination of contexts as input, depicted in Table \ref{tab:abla1}. 
Note that, in this ablation experiments, ALE module is temporarily excluded from the framework. 
In particular, we employ the baseline model based on FCN-8s \cite{long2015fully} without any contexts as extra inputs. Thus, the check-marks on the table represent whether we introduce the correlation between the corresponding context and the local patch in the model.
% For evaluation, we show the results using the local context as well as the medium and large context. 
As shown in Table \ref{tab:abla1}, in general, integrating contexts obviously improves the segmentation performance, which boosts the mIOU metrics from $71.84\%$ to $73.47\%$ in DeepGlobe and from $69.08\%$ to $73.50\%$ in Inria Aerial. 
Amongst these variants, the local context (i.e., the local patch itself) provides the non-local self-correlation cues to slightly improve the baseline. Yet, the self-correlation features of local patch do not provide sufficient contextual information to further infer the semantics of the patch. On the other hand, the medium and large contexts bring in rich contextual information of different scales and thus facilitate the segmentation. But, relying on medium context or large context alone may not always give rise to better results. For instance, for Inria Aerial, the results from the medium or large context are even slightly worse than the one from local context. Hence, exploiting the complementary information from the contexts of different scales is valuable for obtaining better local segmentation. 
As shown in Fig.~\ref{fig:abla_context}, we show several examples on the efficacy of contexts. 
In Fig.~\ref{fig:vis3}, we also demonstrate the results generated with or without contexts. Besides, we show the attention maps (i.e. correlation) of the local, medium, and large contexts with regards to the local patch. As shown, the contexts of various scales guide the model to focus on different parts of the local patch. Accordingly, our model can estimate the corresponding weight maps for feature fusion, which appears to be consistent with the attention maps. 
} 

% \subsubsection{Locality-aware contextual correlation} 

\subsubsection{Gated Positional Embeddings} 
\wx{In our preliminary version \cite{li2021contexts}, our LCF module is completely position-agnostic and it may impair the model due to permutation equivalence. 
% Contextual features and local features are correlated spatially due to the specific position relationship between them. Therefore, contextual features may be blind in position when it provides relevant information to local features without positional information. 
Therefore, based on on our preliminary version, we add the gated positional embeddings into locality-aware contextual correlation module to further strengthen contexts. As observed in Table \ref{tab:abla1}, the gated positional embeddings (GPE) generally plays a positive role comparing to the counterparts without GPE, especially on DeepGlobe. Furthermore, in Fig.~\ref{fig:abla_context}, the gated positional embeddings tend to refine the results via yielding the sensitivity towards position.}
\wx{Additionally, we also show the attention maps and the corresponding weight maps without gated positional
embeddings in Fig.~\ref{fig:vis1}. In contrast to Fig.~\ref{fig:vis3}, without GPE, the locality-aware features may bring noise into attention maps. With GPE, the attention maps appear to be more smooth and consistent.
}

\begin{table}
    \centering
    \setlength{\tabcolsep}{9pt}
    \footnotesize
    \caption{Impacts of different scales of contexts for local segmentation.}
    \begin{tabular}{c|ccc|c}
        \toprule
         \multirow{2}{*}{\textbf{Dataset}}& \multicolumn{3}{c|}{\textbf{Context Size} (pixels)}  & \multirow{2}{*}{\textbf{mIOU}} \\
        \cmidrule{2-4}
        & Local & Medium & Large &  \\
        \midrule
        \parbox[t]{2mm}{\multirow{5}{*}{\rotatebox[origin=c]{90}{DeepGlobe}}} & 508 & 1016 & 2448 & 73.23 \\
        &508 & 1524 & 2032 & 72.76  \\
        &\textbf{508} & \textbf{1524} & \textbf{2448} & \textbf{73.47} \\
        &508 & 2032 & 2448 & 73.05  \\
        &1016 & 1524 & 2448 & 73.34  \\
        \midrule
        \parbox[t]{2mm}{\multirow{6}{*}{\rotatebox[origin=c]{90}{Inria Aerial}}} &508 & 762 & 1524 & 73.05 \\
        &508 & 1016 & 1270 & 73.13 \\
        &\textbf{508} & \textbf{1016} & \textbf{1524} & \textbf{73.50} \\
        &508 & 1270 & 1524 & 73.31 \\
        &508 & 1016 & 2032 & 73.24 \\
        &762 & 1016 & 1524 & 73.20 \\
        \bottomrule
    \end{tabular}
    \label{tab:abla2}
\end{table}

\begin{table}
    \centering
    \setlength{\tabcolsep}{10pt}
    \footnotesize
    \caption{Analysis of the multi-context fusion scheme in LCF.}
    \begin{tabular}{c|cc}
        \toprule
        \textbf{Fusion Scheme}  & \textit{DeepGlobe} & \textit{Inria Aerial}\\
        % \midrule
        % Fusion & mIOU (\%)  \\
        \midrule
        Averaging Fusion & 72.89 & 73.20   \\
        Weighted Fusion & 73.16 & 73.37\\
        Adaptive Fusion & \textbf{73.47} & \textbf{73.50} \\
        \bottomrule
    \end{tabular}
    \label{tab:abla3}
\end{table}

\begin{figure}[t]
    \centering
    \footnotesize
	\begin{tabular}{c@{}c@{}c@{}c}
	
		\includegraphics[width=0.11\textwidth,height=1.8cm]
        {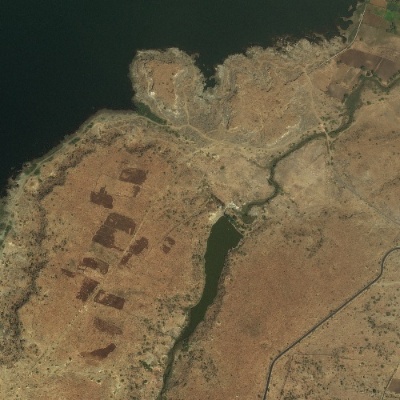} &
		\includegraphics[width=0.11\textwidth,height=1.8cm]
        {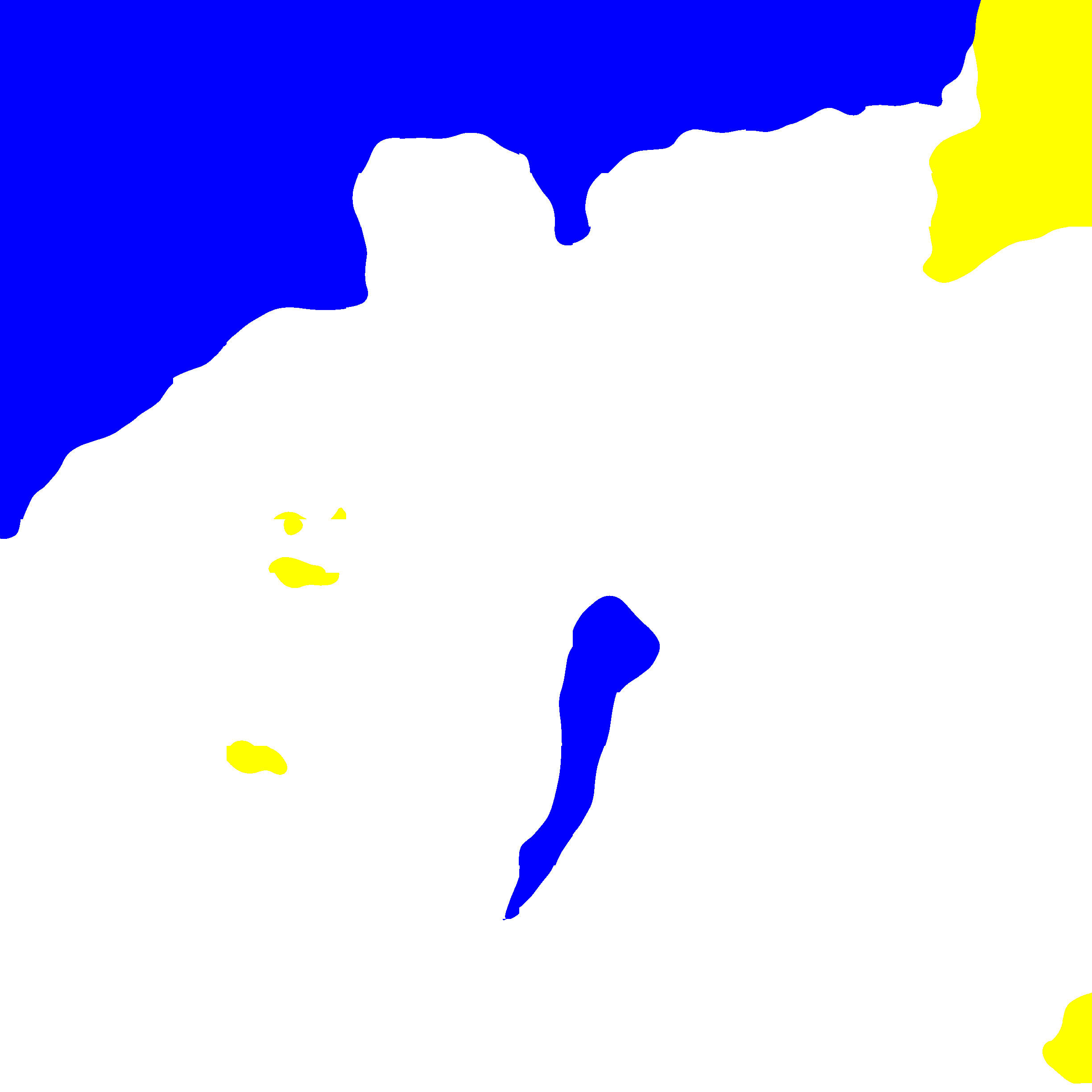}&
        \includegraphics[width=0.11\textwidth,height=1.8cm]
        {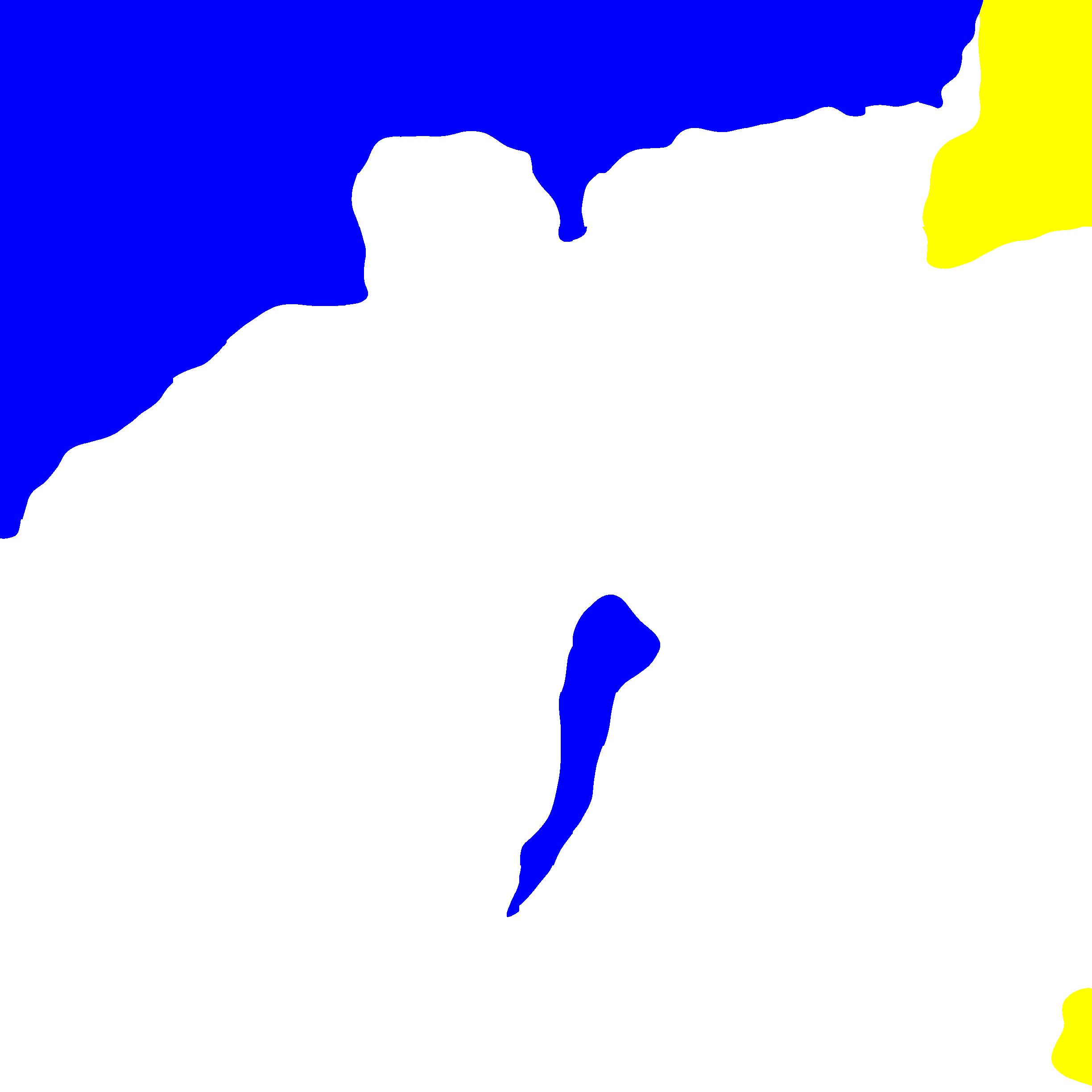}&
        \includegraphics[width=0.11\textwidth,height=1.8cm]
        {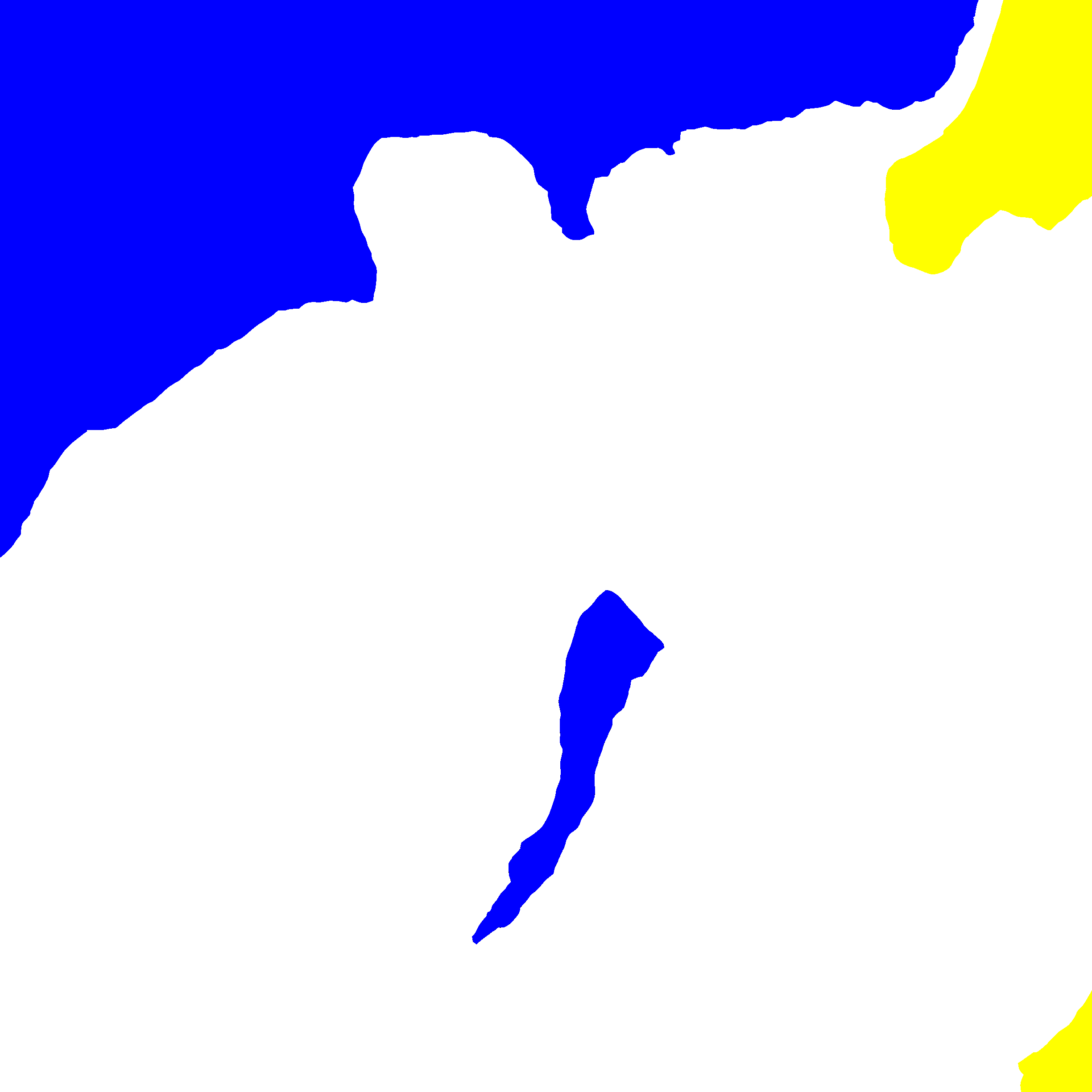}\\
        
		\includegraphics[width=0.11\textwidth,height=1.8cm]
        {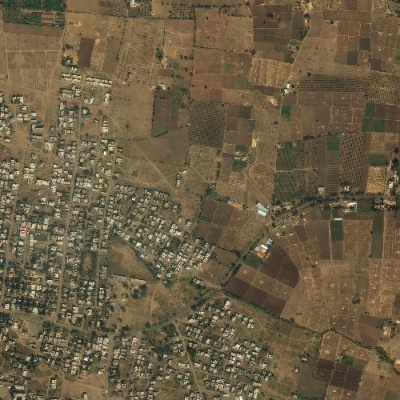} &
		\includegraphics[width=0.11\textwidth,height=1.8cm]
        {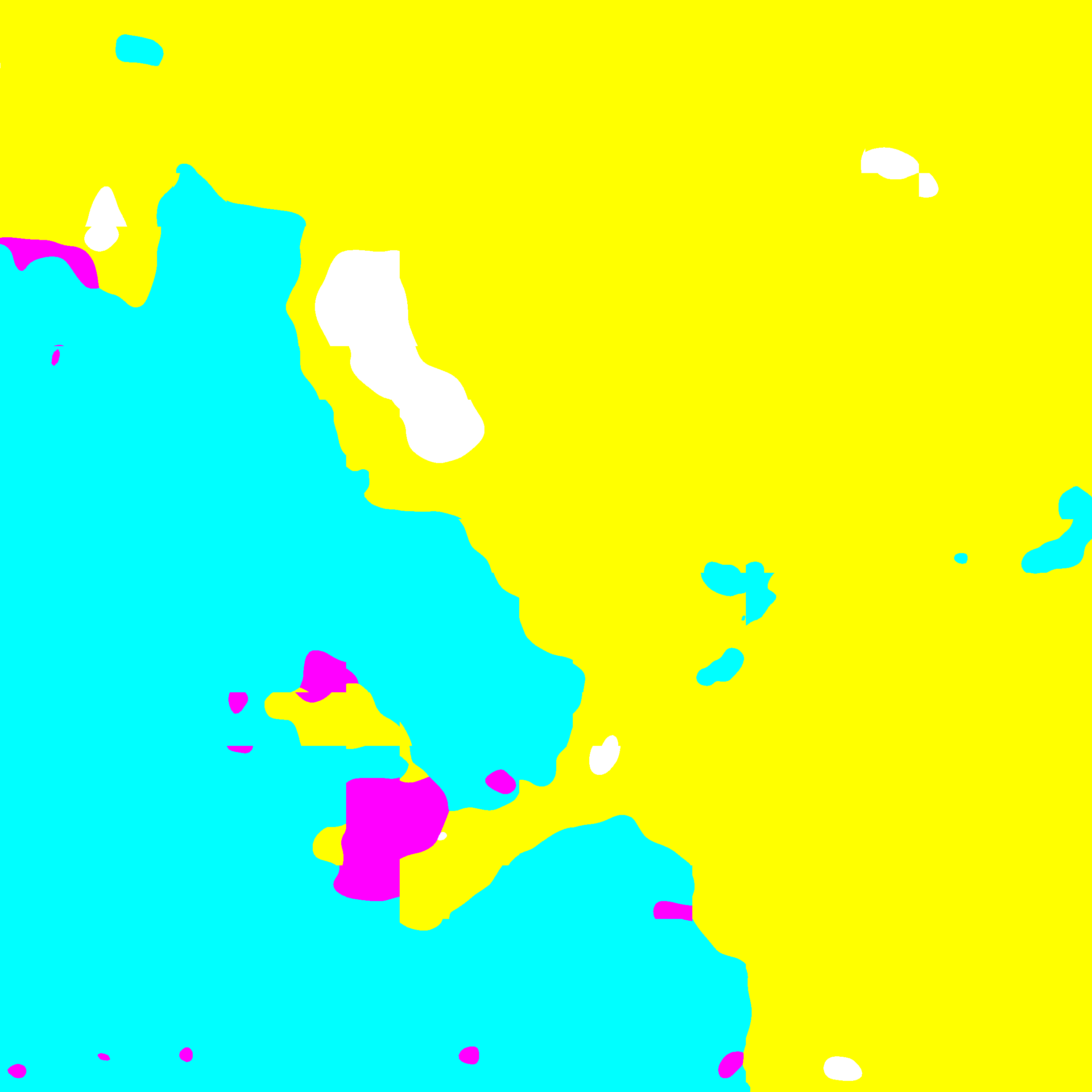}&
        \includegraphics[width=0.11\textwidth,height=1.8cm]
        {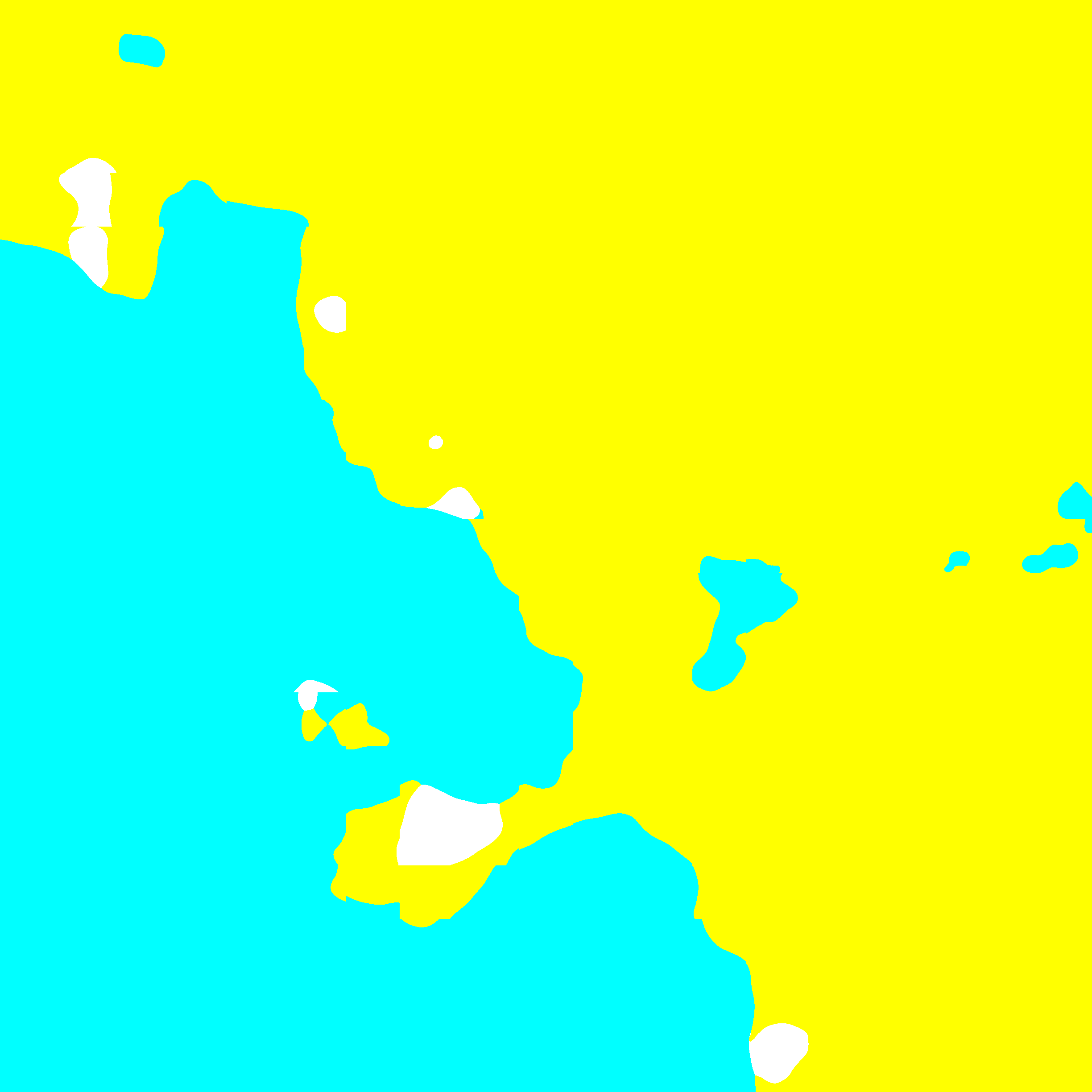}&
		\includegraphics[width=0.11\textwidth,height=1.8cm]
        {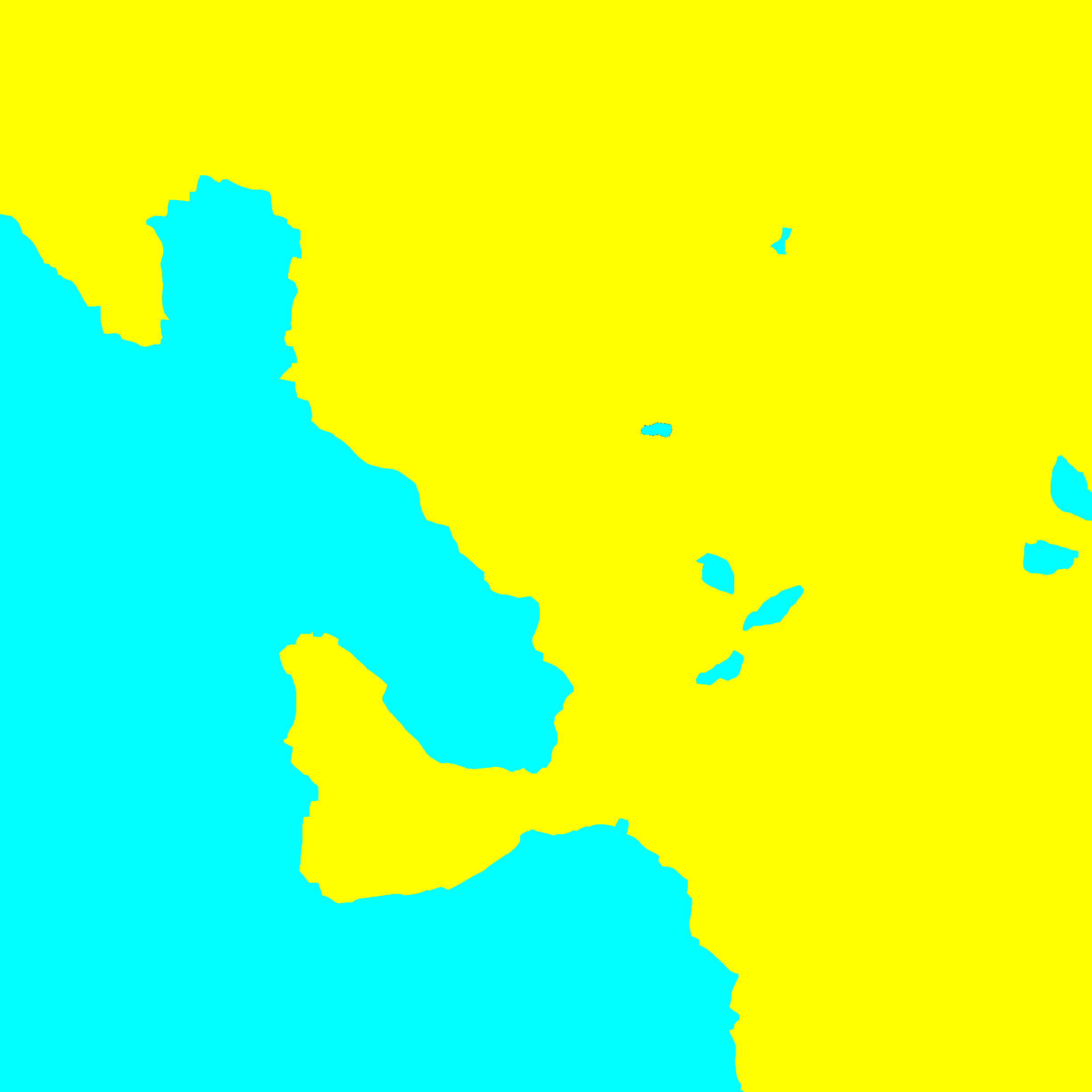}\\
        
        \includegraphics[width=0.11\textwidth,height=1.8cm]
        {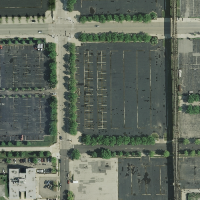} &
		\includegraphics[width=0.11\textwidth,height=1.8cm]
        {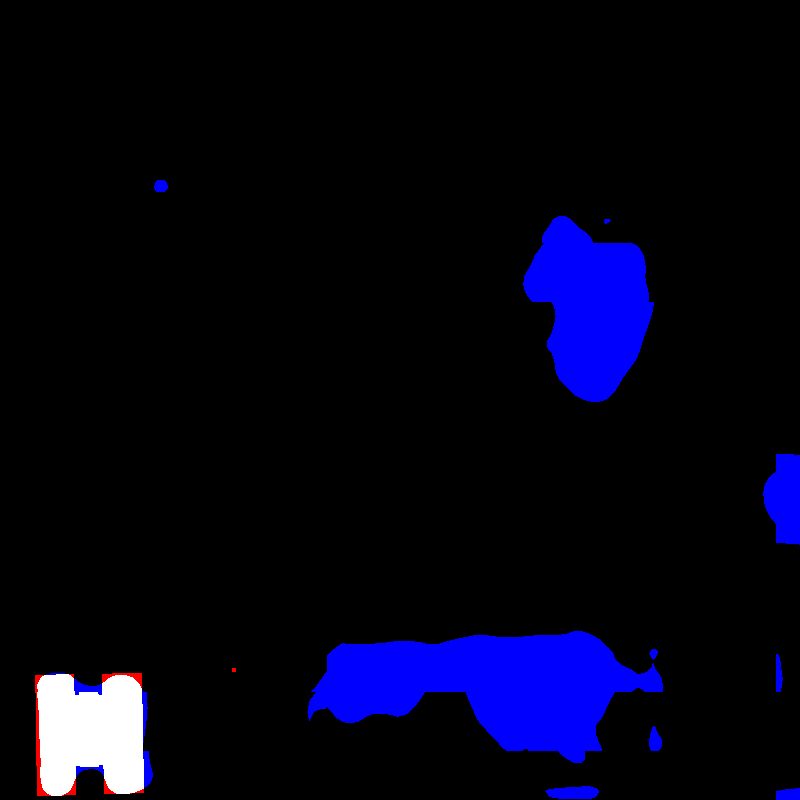}&
        \includegraphics[width=0.11\textwidth,height=1.8cm]
        {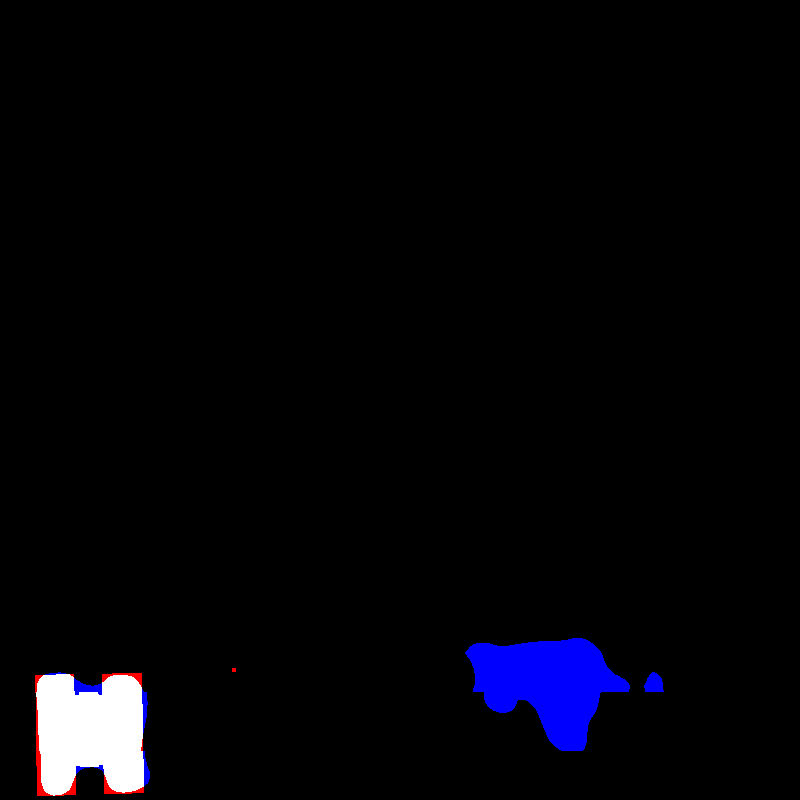}&
		\includegraphics[width=0.11\textwidth,height=1.8cm]
        {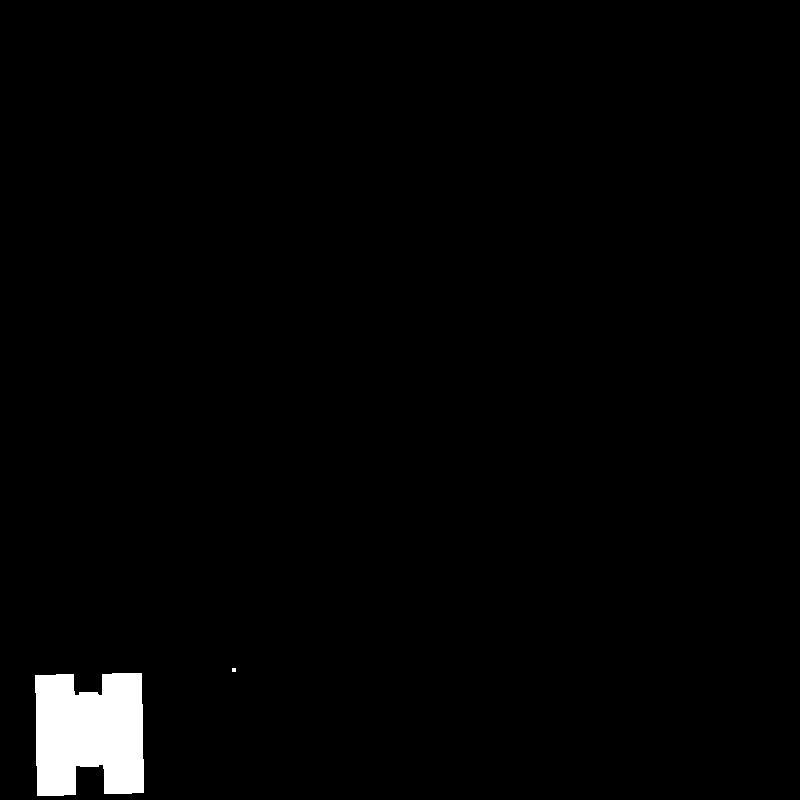}\\
        
        \includegraphics[width=0.11\textwidth,height=1.8cm]
        {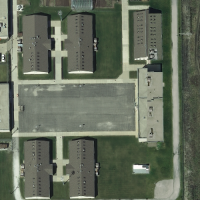} &
		\includegraphics[width=0.11\textwidth,height=1.8cm]
        {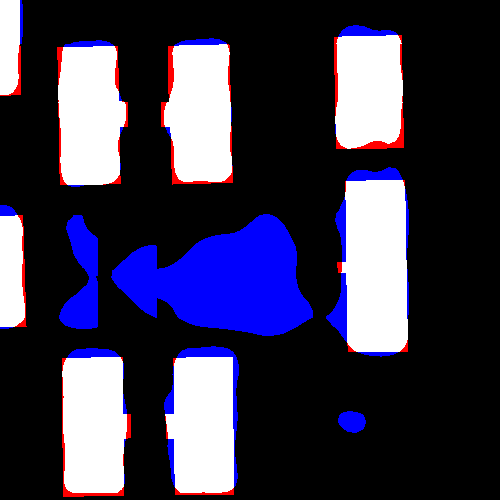}&
        \includegraphics[width=0.11\textwidth,height=1.8cm]
        {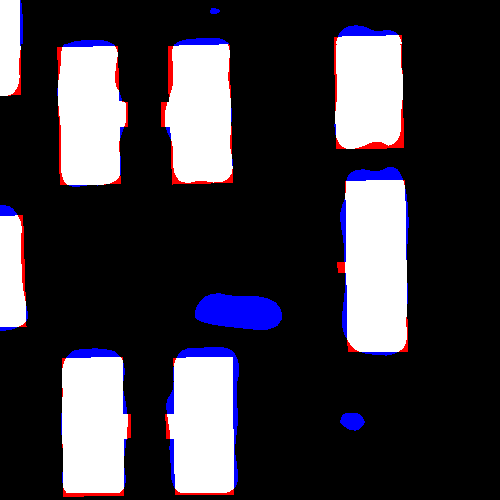}&
		\includegraphics[width=0.11\textwidth,height=1.8cm]
        {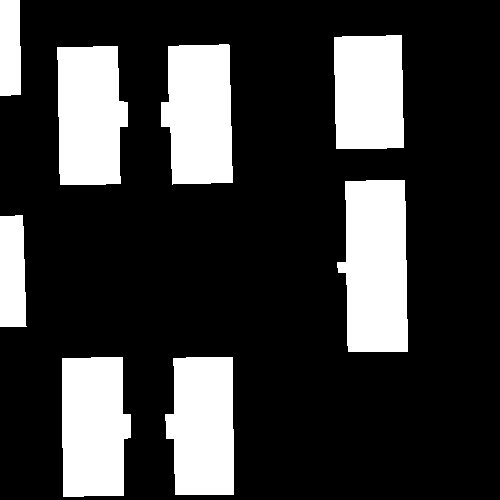}\\
        
		Input Image & w/o Fusion & w/ Fusion & Ground truth
	\end{tabular}
	\caption{Examples show the efficacy of our adaptive fusion.}
	% \vspace{-0.3cm}
	\label{fig:abla_fusion1}
\end{figure}

\subsubsection{Scales of Contexts} 
We investigate how context sizes affect the segmentation performance. We assess the models with different context sizes for both benchmarks in Table \ref{tab:abla2}. Intuitively, if the context size is close to the size of local patch, the local patch and context are highly overlapped and they share too much redundant information that may not bring much performance gain. \lqq{In addition, too large contextual patch centered on local patch may cause sampling pixels outside the bounds of the image, which leads to performance degradation to a certain extent.} Therefore, for the sizes of candidate contexts, we choose the multiples of the local patch size ($508 \times 508$) as the sizes of three contexts (i.e., small, medium, and large). 
For DeepGlobe, the optimal context sizes are $508 \times 508$, $1524 \times 1524$, and $2448 \times 2448$, in which the large context is exactly the entire image (i.e., the global context). For Inria Aerial, the optimal context sizes are $508 \times 508$, $1016 \times 1016$, and $1524 \times 1524$. The different configurations of these two datasets are due to the characteristics of their images. DeepGlobe includes the geospatial images with different terrains (e.g., water and forest), while Inria Aerial contains top-down urban views, in which a large number of buildings can be observed. Therefore, for DeepGlobe, with the entire image as our large context can help better understand the semantics. On the contrary, for Inria Aerial, too large context after being rescaled into a smaller size will lose details of cities and makes the model hard to discern the buildings in the images, which thus causes the performance degradation. 

% \begin{table}
%     \centering
%     \setlength{\tabcolsep}{3pt}
%     \footnotesize
%     \caption{Scales of contexts for local segmentation.}
%     \begin{tabular}{c|ccc|c|c|ccc|c}
%         \toprule
%          & \multicolumn{3}{c|}{Context Size (pix.)}  & \multirow{2}{*}{mIOU} & & \multicolumn{3}{c|}{Context Size (pix.)}  & \multirow{2}{*}{mIOU}\\
%         \cmidrule{2-4}\cmidrule{7-9}
%         & Small & Medium & Large & & & Small & Medium & Large &\ \\
%         \midrule
%         \parbox[t]{2mm}{\multirow{5}{*}{\rotatebox[origin=c]{90}{DeepGlobe}}} & 508 & 1016 & 2448 & 73.23 & \parbox[t]{2mm}{\multirow{6}{*}{\rotatebox[origin=c]{90}{Inria Aerial}}} &508 & 762 & 1524 & 73.05 \\
%         &508 & 1524 & 2032 & 72.76 & &508 & 1016 & 1270 & 73.13 \\
%         &\textbf{508} & \textbf{1524} & \textbf{2448} & \textbf{73.47} & &\textbf{508} & \textbf{1016} & \textbf{1524} & \textbf{73.50}\\
%         &508 & 2032 & 2448 & 73.05 & &508 & 1270 & 1524 & 73.31 \\
%         &1016 & 1524 & 2448 & 73.34 & &508 & 1016 & 2032 & 73.24 \\
%         & &  &  &  & &762 & 1016 & 1524 & 73.20 \\
%         \bottomrule
%     \end{tabular}
%     \label{tab:abla2}
% \end{table}

\begin{figure*}[t]
    \centering
    \includegraphics[width=0.9\textwidth]{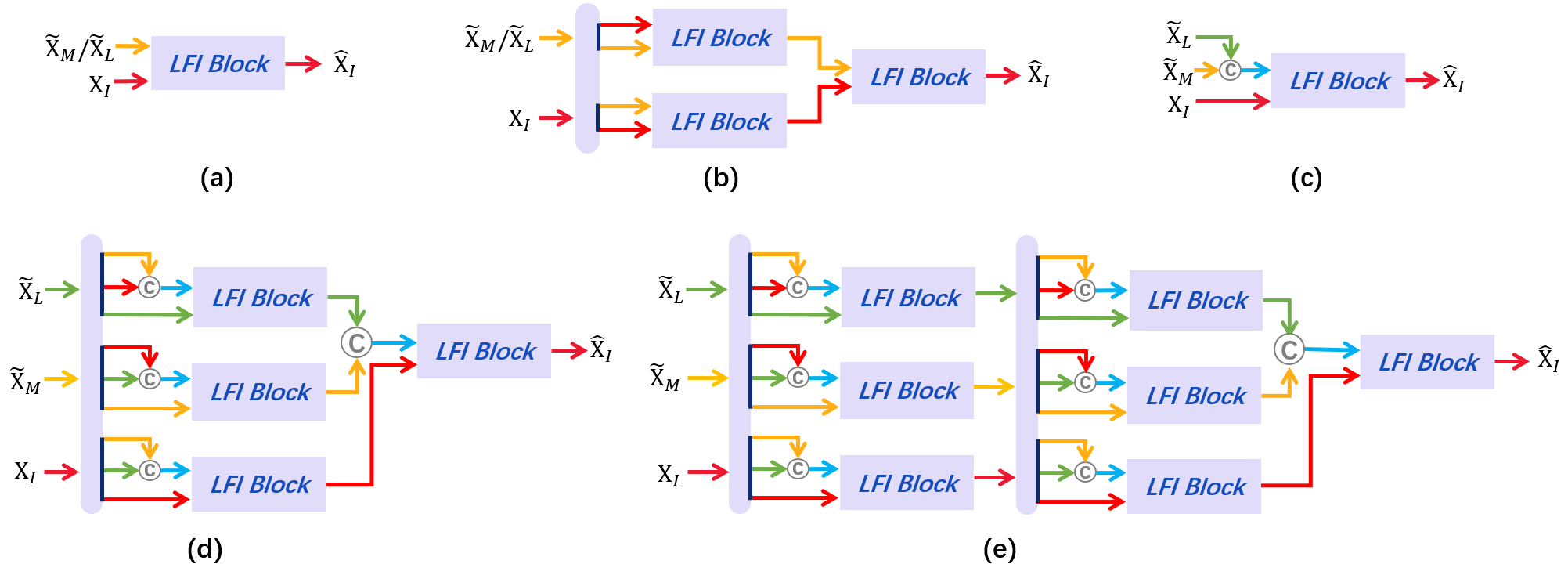}
    \caption{\wxb{We show several variants of ALE with different depths. We use different colored arrows to represent the sources of different features.}}
    \label{fig:ALEs}
\end{figure*}

\begin{table}[t]
    \centering
    \setlength{\tabcolsep}{5pt}
    \footnotesize
    \caption{Analysis of ALE structure in terms of mIOU.}
    \begin{tabular}{c|cc|cc}
        \toprule
        \multirow{2}{*}{\textbf{Structure}} & \multicolumn{2}{c|}{\textbf{Reference Patch}} & \multirow{2}{*}{\textit{DeepGlobe}} & \multirow{2}{*}{\textit{Inria Aerial}}\\
          & $\tilde{\textbf{X}}_M$ & $\tilde{\textbf{X}}_L$ & & \\
        \midrule
        w/o ALE &  &  & 73.47 & 73.50 \\
        Concat & $\checkmark$ & $\checkmark$ &  72.20 & 72.80   \\
        ALE only & $\checkmark$ & $\checkmark$ &  72.69 & 73.62   \\
        \midrule
        \multicolumn{5}{c}{\textbf{LCF + Variants of ALE}} \\\midrule
        (a) & $\checkmark$ &   & 73.76 & 74.22\\
        (a) &  & $\checkmark$ & 73.57 & 74.19\\
        (b) & $\checkmark$ &  & 73.78 & 74.39\\
        (b) &  & $\checkmark$ & 73.68 & 74.35\\
        (c) & $\checkmark$ & $\checkmark$ & 73.55 & 74.32\\
        (d) & $\checkmark$ & $\checkmark$ & \textbf{73.91} & \textbf{74.58}\\
        (e) & $\checkmark$ & $\checkmark$ & 73.77 & 74.56\\
        % +Medium EV & 73.76 & 74.22\\
        % +Medium 3EV & 73.78 & 74.39\\
        % +Large EV & 73.57 & 74.19\\
        % +Large 3EV & 73.68 & 74.35\\
        % +Medium\&Large EV & 73.55 & 74.32\\
        % +Medium\&Large 3EV & \textbf{73.91} & \textbf{74.58}\\
        % +Medium\&Large 6EV & 73.77 & 74.56\\
        \bottomrule
    \end{tabular}
    \label{tab:abla4}
\end{table}

\begin{figure}[t]
    \centering
    \tiny
	\begin{tabular}{c@{}c@{}c@{}c@{}c}
	
		\includegraphics[width=0.095\textwidth,height=1.5cm]
        {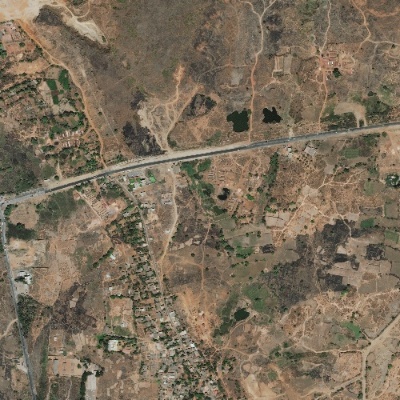} &
		\includegraphics[width=0.095\textwidth,height=1.5cm]
        {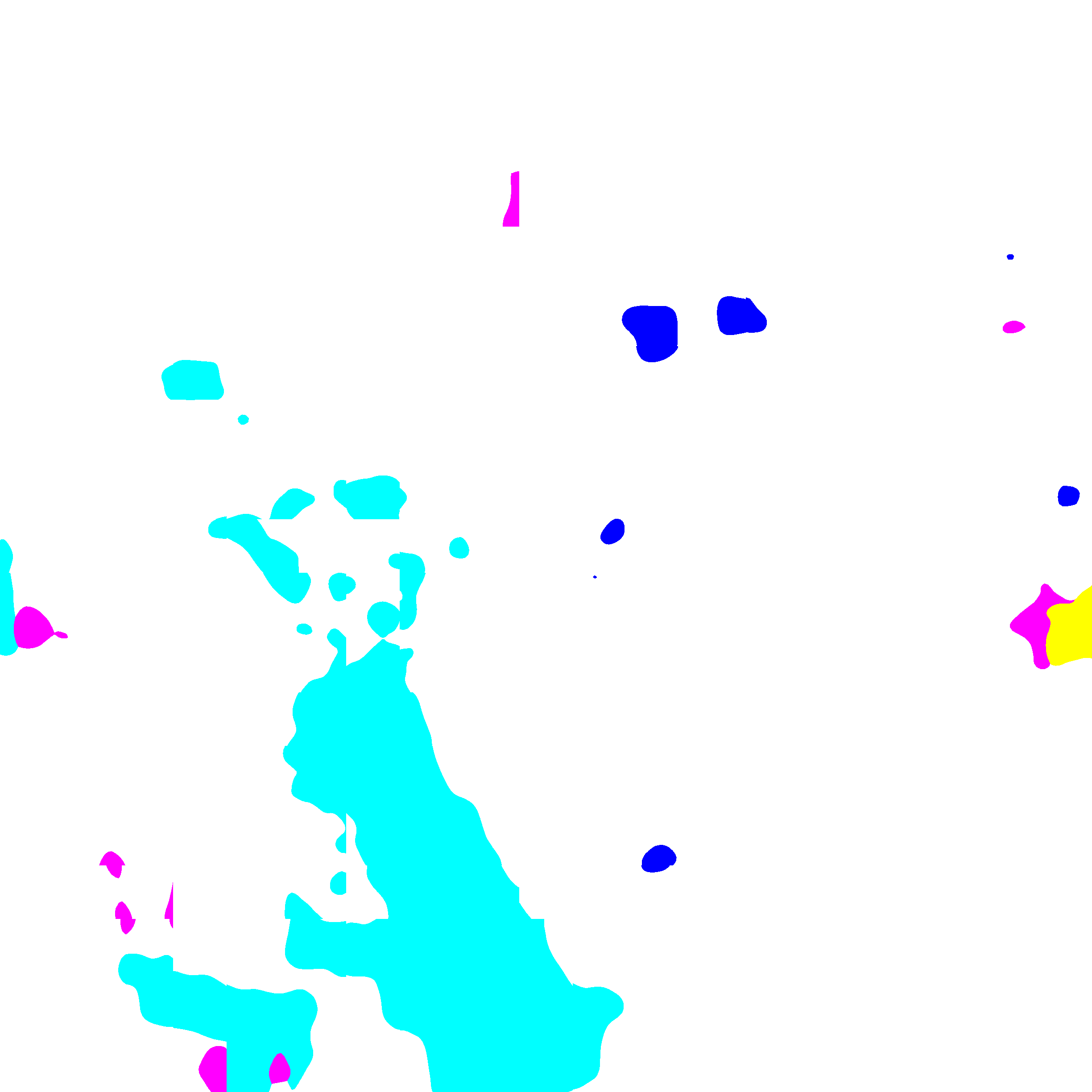}&
        \includegraphics[width=0.095\textwidth,height=1.5cm]
        {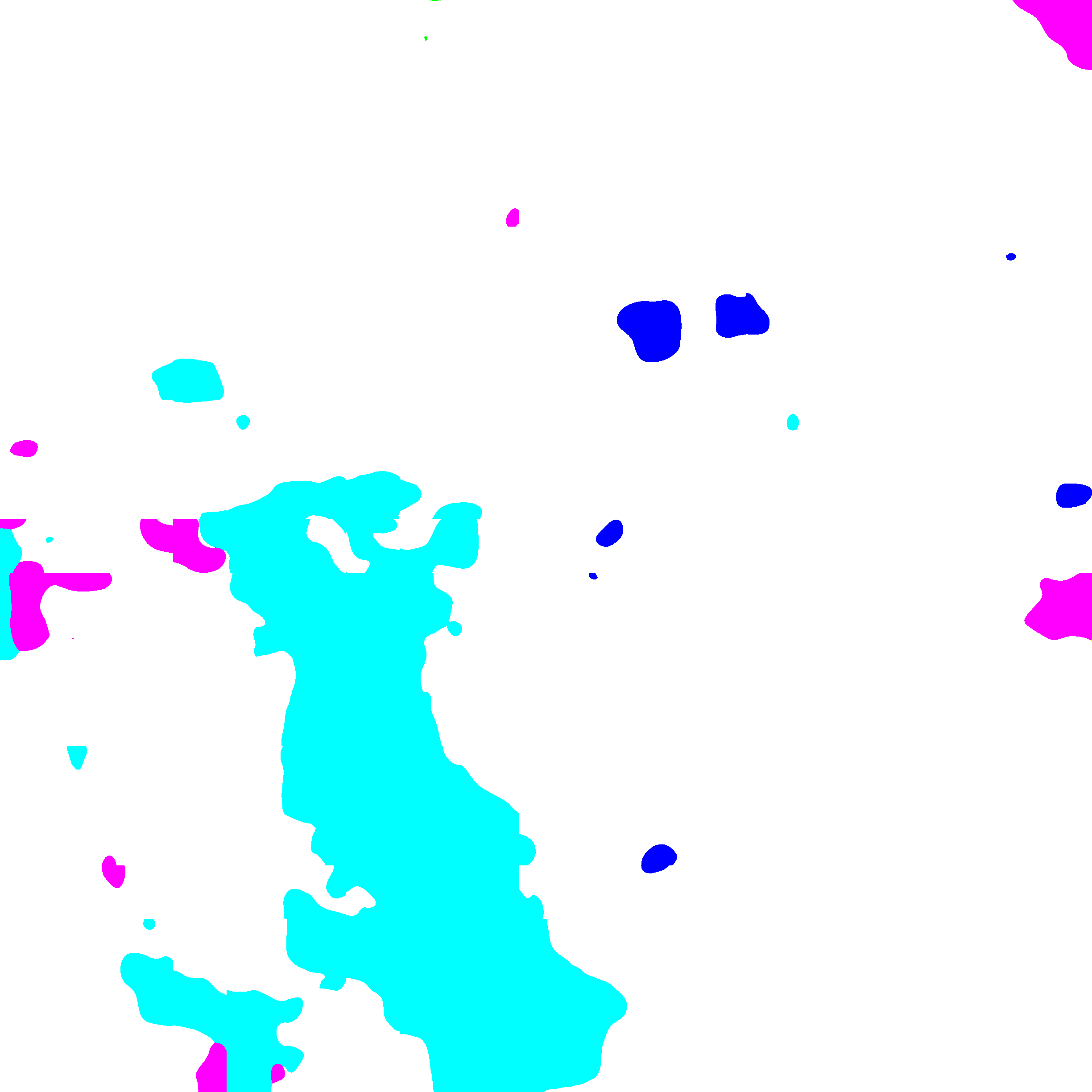}&
        \includegraphics[width=0.095\textwidth,height=1.5cm]
        {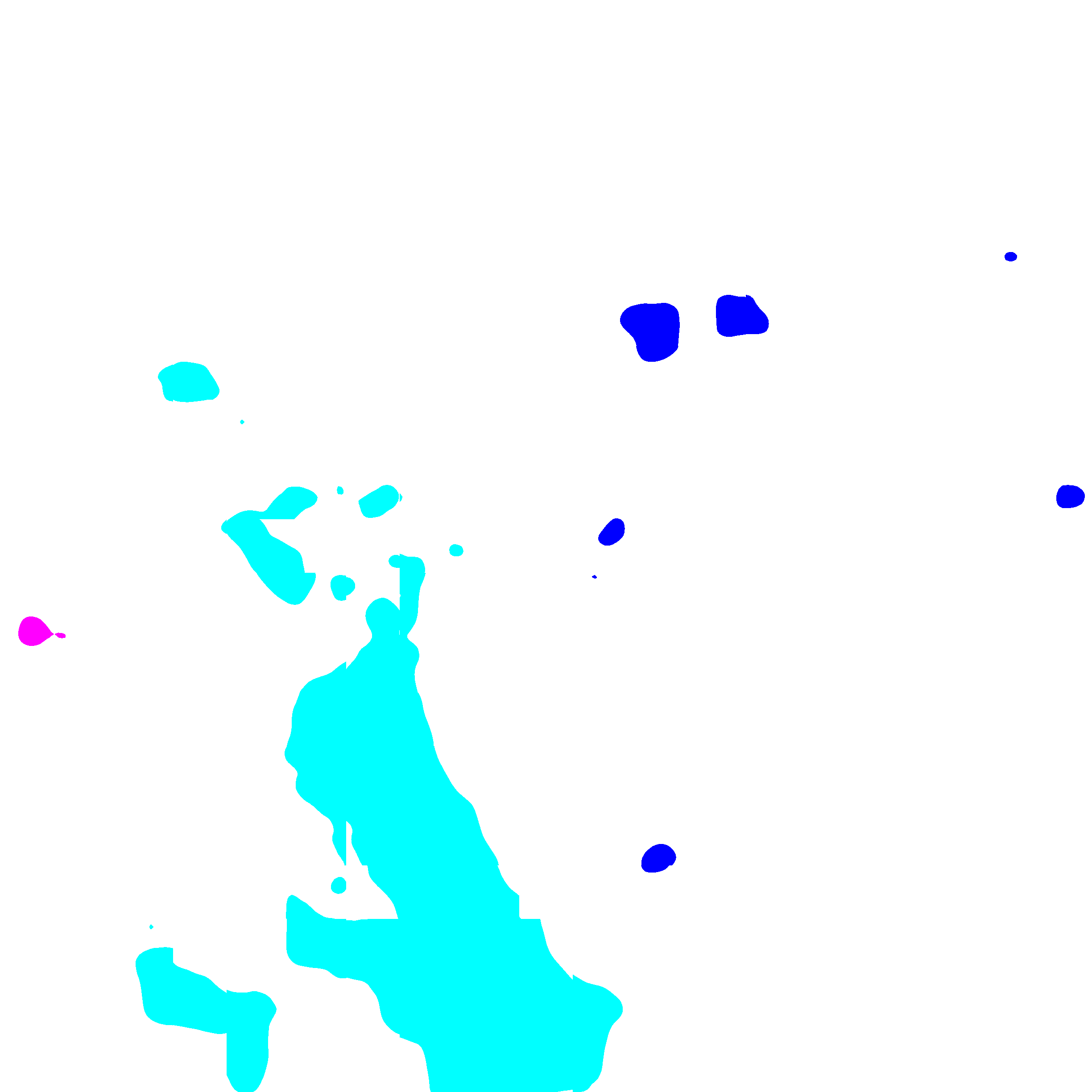}&
        \includegraphics[width=0.095\textwidth,height=1.5cm]
        {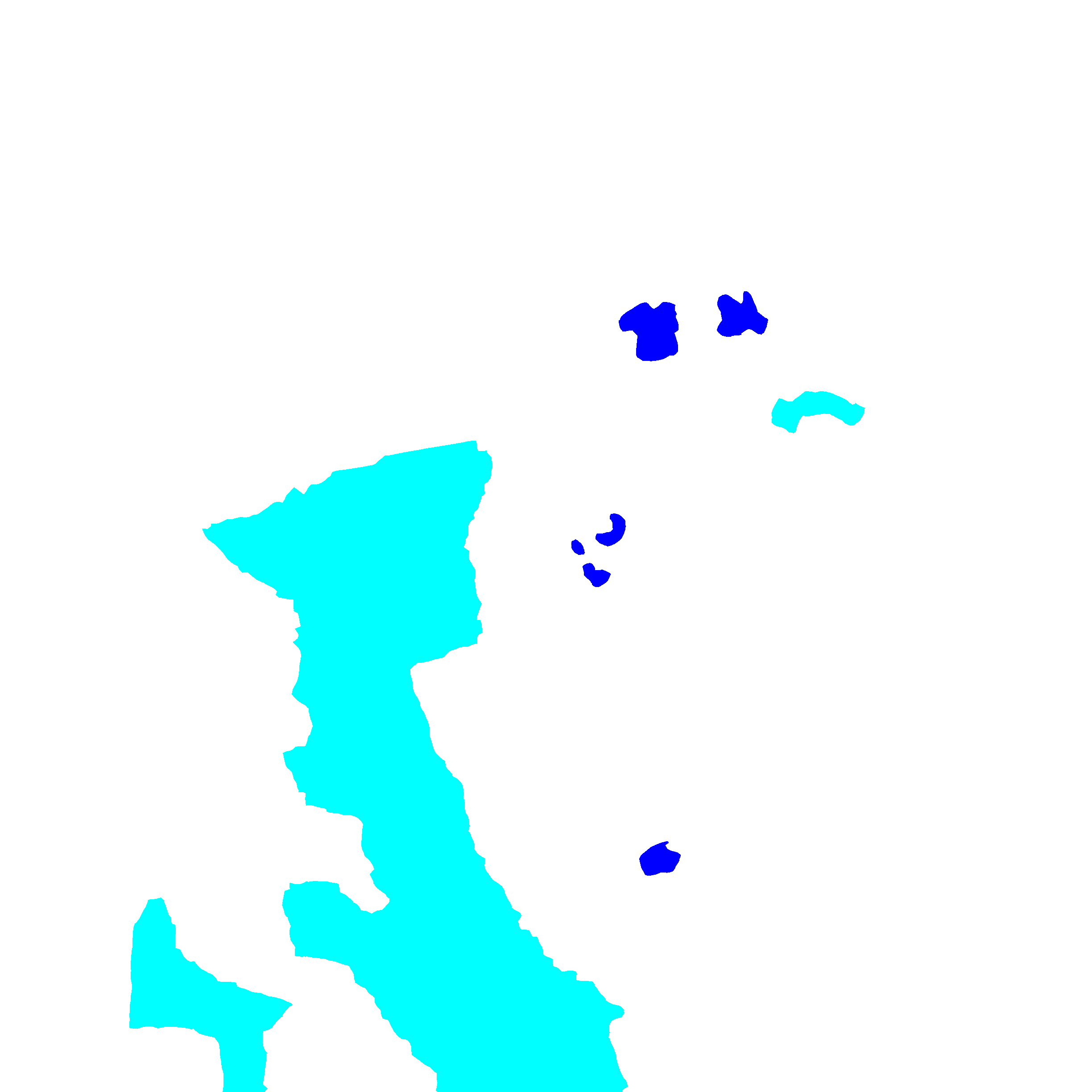}\\
        
		\includegraphics[width=0.095\textwidth,height=1.5cm]
        {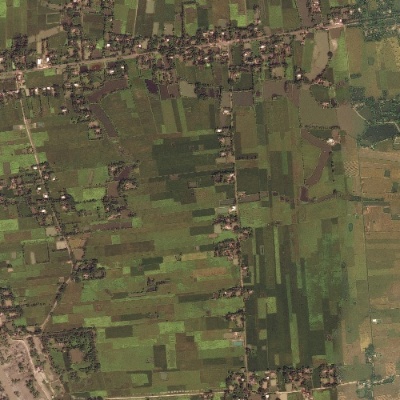} &
		\includegraphics[width=0.095\textwidth,height=1.5cm]
        {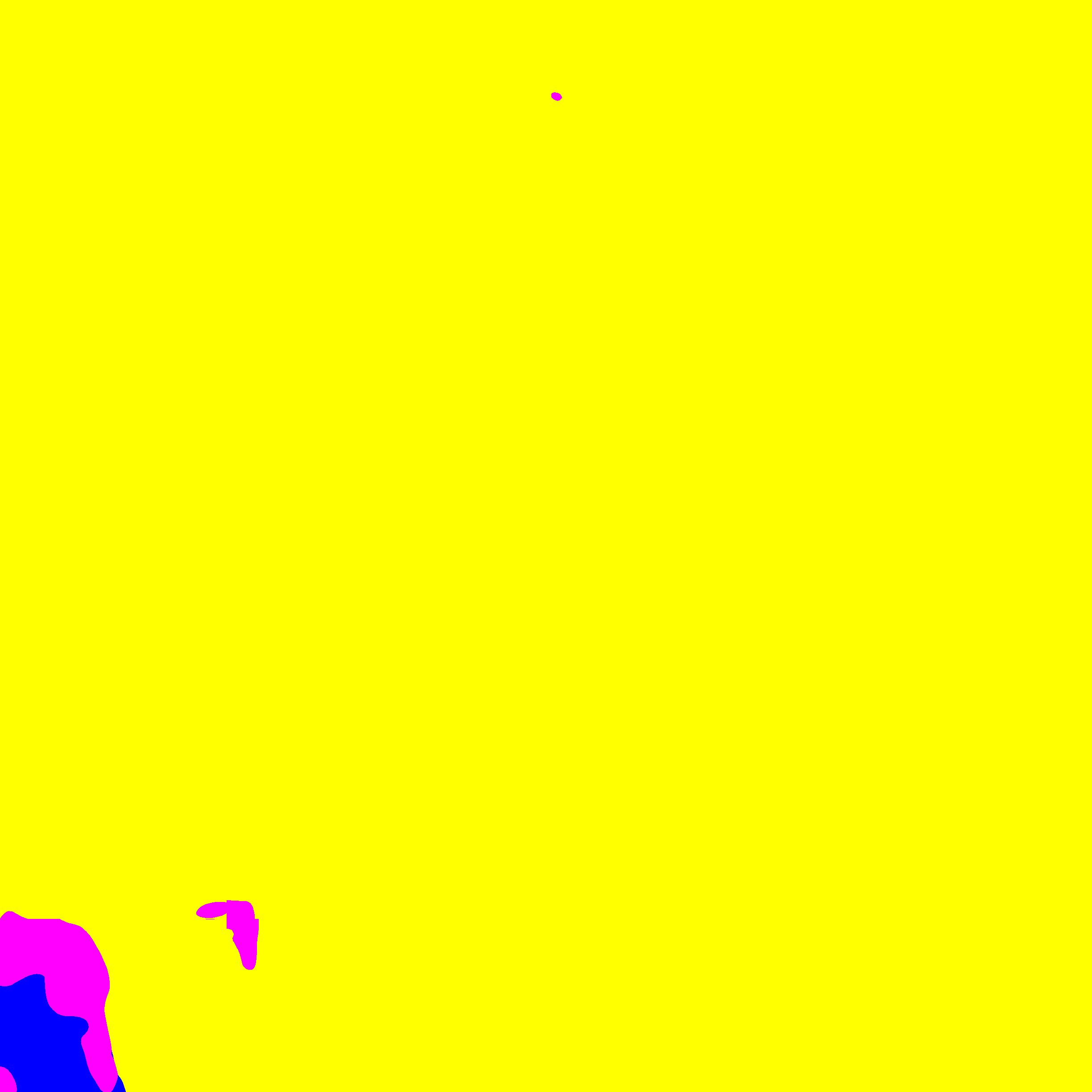}&
    	\includegraphics[width=0.095\textwidth,height=1.5cm]
        {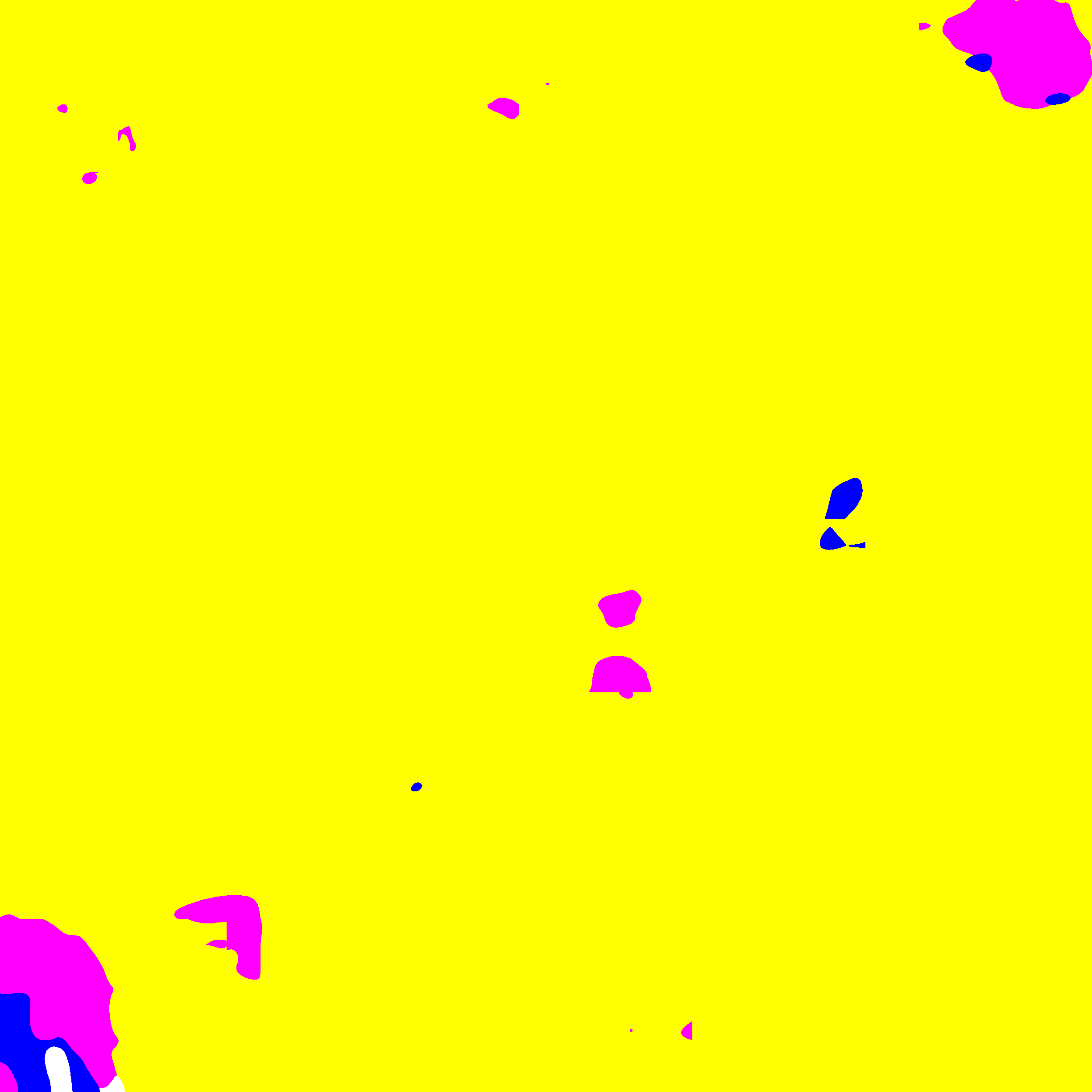}&
        \includegraphics[width=0.095\textwidth,height=1.5cm]
        {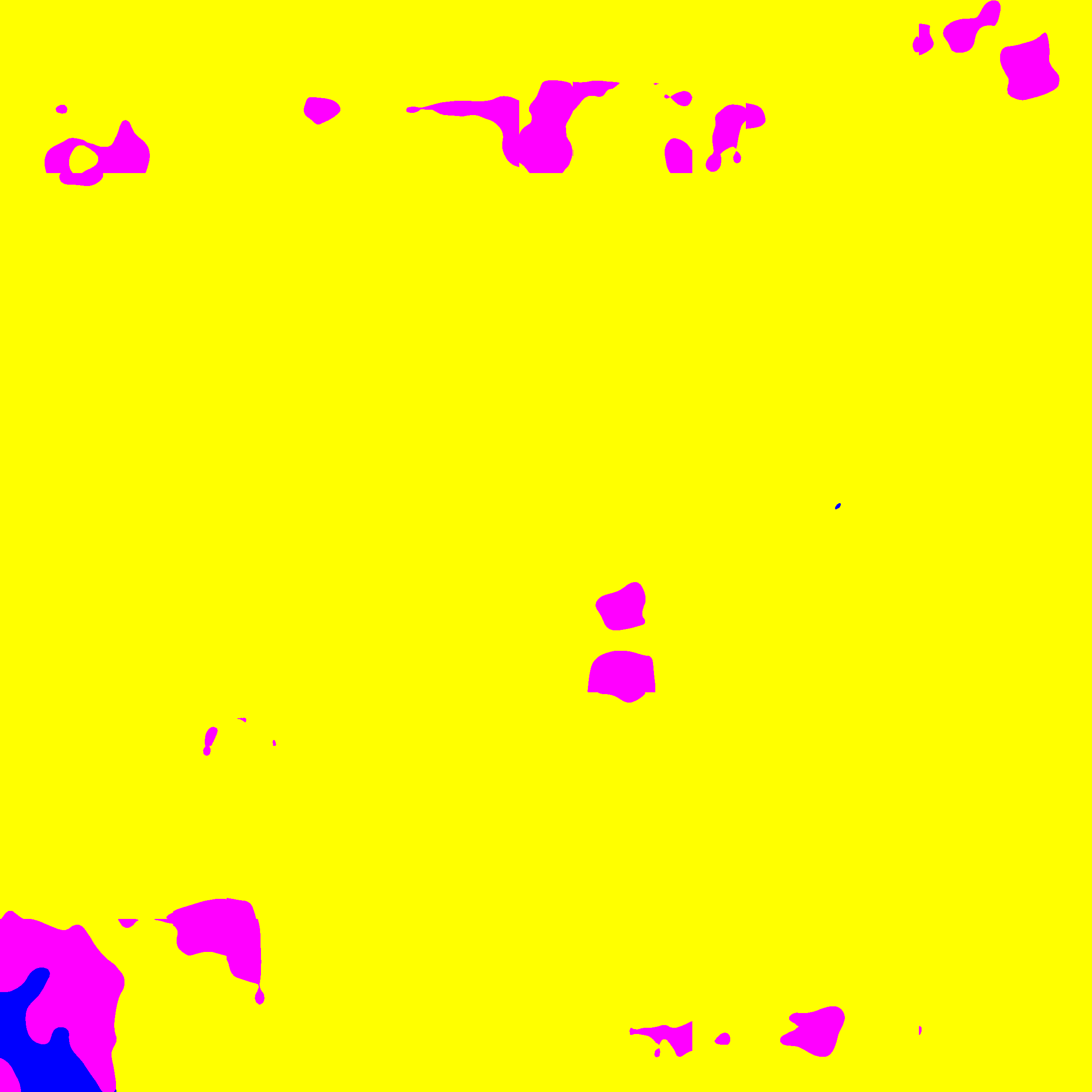}&
        \includegraphics[width=0.095\textwidth,height=1.5cm]
        {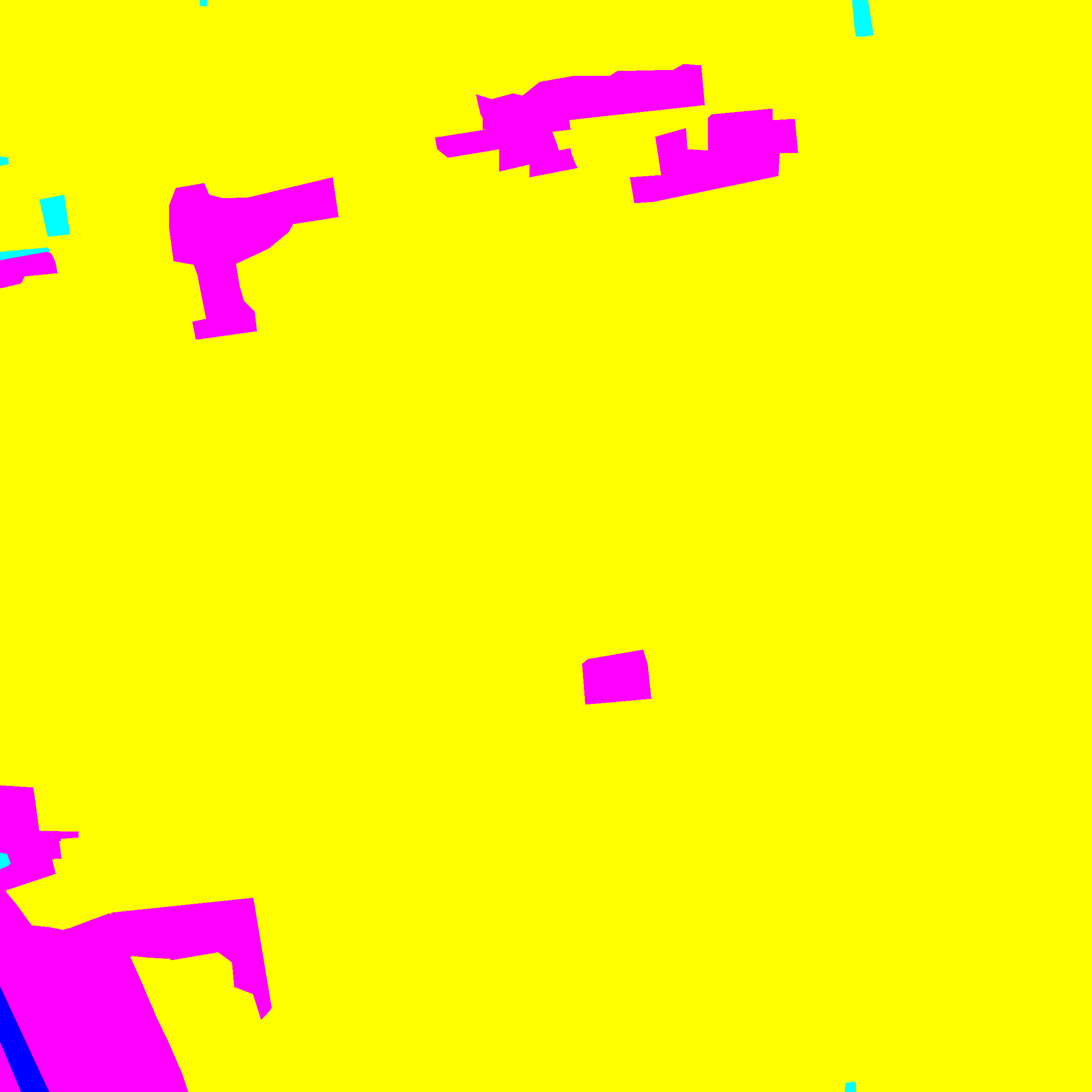}\\
        
        \includegraphics[width=0.095\textwidth,height=1.5cm]
        {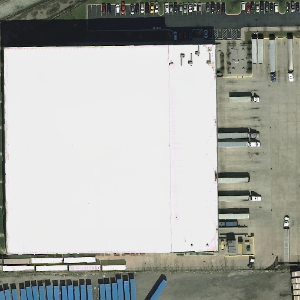} &
		\includegraphics[width=0.095\textwidth,height=1.5cm]
        {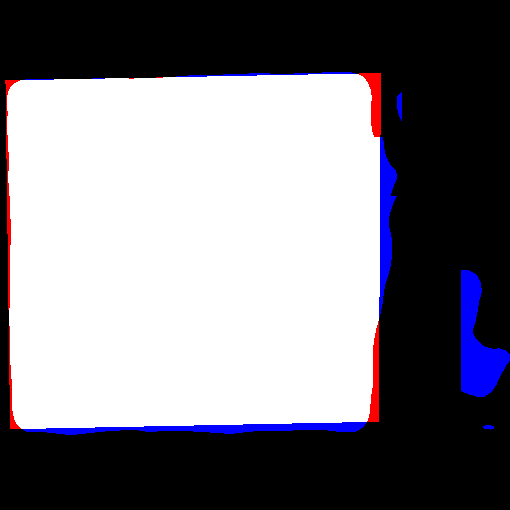}&
		\includegraphics[width=0.095\textwidth,height=1.5cm]
        {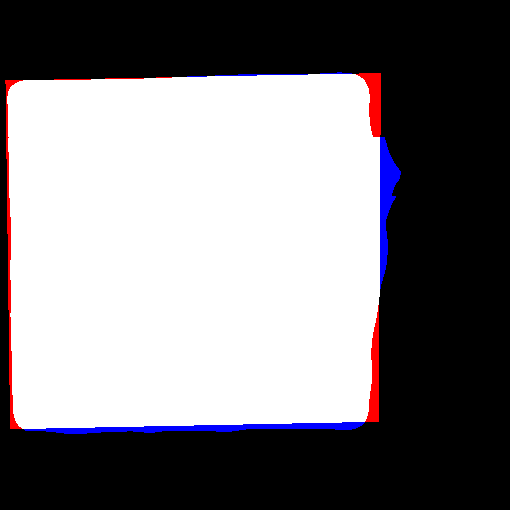}&
        \includegraphics[width=0.095\textwidth,height=1.5cm]
        {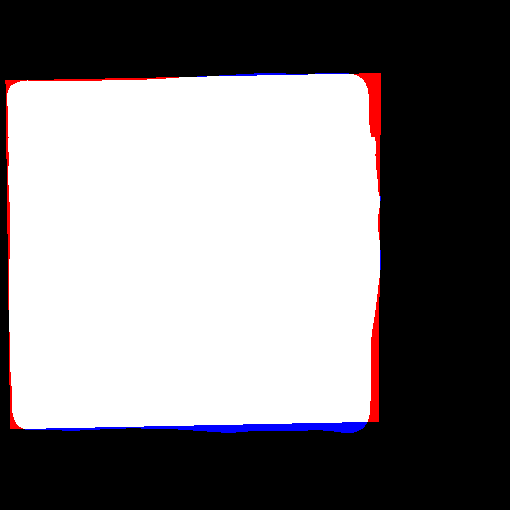}&
        \includegraphics[width=0.095\textwidth,height=1.5cm]
        {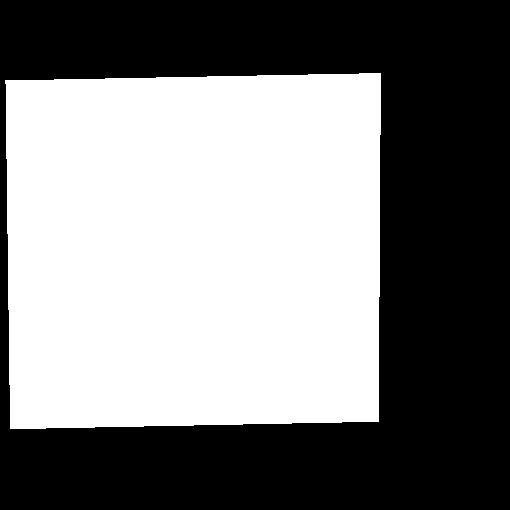}\\
        
        \includegraphics[width=0.095\textwidth,height=1.5cm]
        {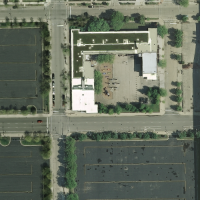} &
		\includegraphics[width=0.095\textwidth,height=1.5cm]
        {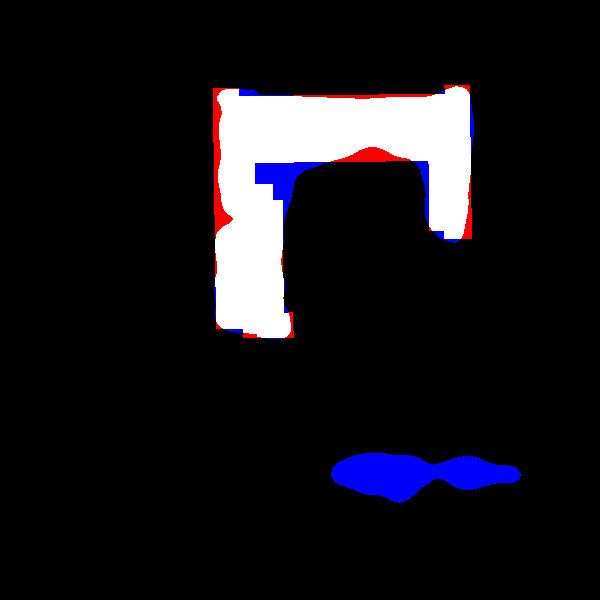}&
		\includegraphics[width=0.095\textwidth,height=1.5cm]
        {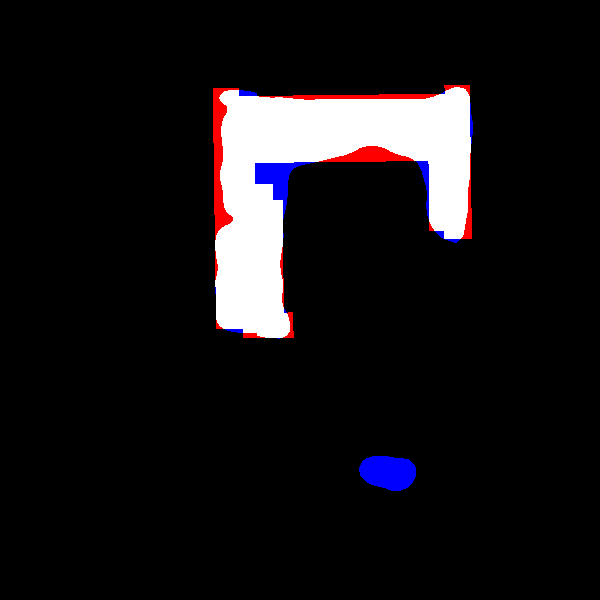}&
        \includegraphics[width=0.095\textwidth,height=1.5cm]
        {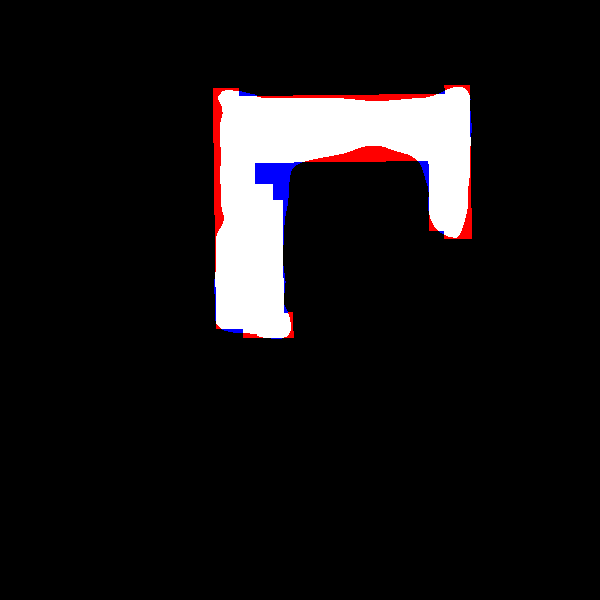}&
        \includegraphics[width=0.095\textwidth,height=1.5cm]
        {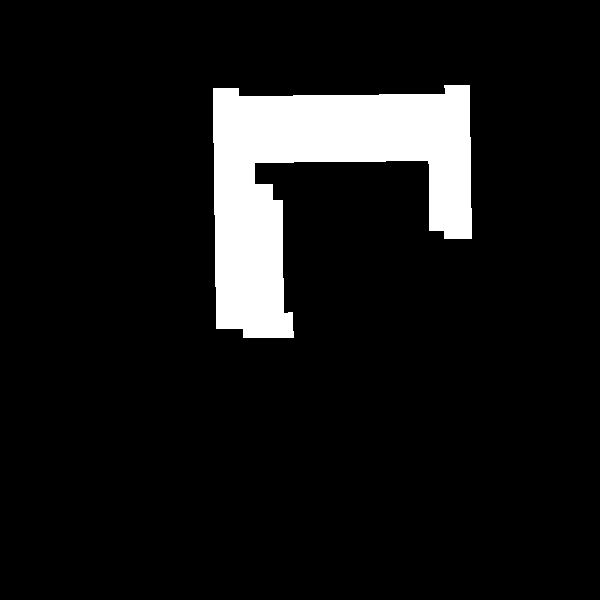}\\

		Input Image &  w/o ALE &  w/o LCF & Both & Ground truth
	\end{tabular}
	\caption{Examples show the efficacy of our proposed LCF and ALE.}
% 	\vspace{-0.3cm}
	\label{fig:abla_ev}
\end{figure}

\subsubsection{Fusion Scheme of LCF} 

To show the advantage of our fusion scheme, we compare our module against two trivial fusion methods: 1) simply averaging the locality-aware features (i.e., $\frac{1}{T}\sum_{t=1}^{T} \overline{\textbf{X}}^t$); and 2) offline estimating the optimal weights of locality-aware features (i.e., the estimated weights remain constant for each dataset). The comparison analysis results in the term of mIOU are demonstrated in Table \ref{tab:abla3}. As observed, our adaptive fusion scheme achieves the best performance over the trivial fusion methods. In Fig.~\ref{fig:abla_fusion1}, we illustrate the exemplar results of our fusion scheme comparing to the results without fusion. Note that, our fusion scheme is essentially the generic version of the average fusion and weighted fusion, so its advantage shown on Table \ref{tab:abla3} may not be significant yet indicates the effectiveness.

% \textbf{Contextual semantics refinement network.} We evaluate the effectiveness of the contextual semantics refinement network. We compare our approach to the trivial methods using non-overlapping montaging and overlapping merging (i.e. averaging). As shown in Table \ref{tab:abla4}, our semantics refinement network surpasses the trivial methods in the term of mIOU. Besides, we study how the context size of this network influences the refinement network. In specific, we evaluate the results computed by the context sizes $762 \times 762$, $1016 \times 1016$, and $1270 \times 1270$, in which $1016 \times 1016$ context mask leads to the best result. In Fig.~\ref{fig:abla_merge}, we demonstrate that our model is able to decrease the boundary artifacts and refine the contour of the semantic mask. Our network can collaborate with prior models. E.g., along with GLNet\cite{chen2019collaborative}, it can also boost its mIOU from 71.6 to 72.6 on DeepGlobe.

\subsubsection{ALE structure}

\wx{To validate the necessity of ALE, we employ the model without ALE as baseline, which is essentially the model with LCF only. 
First of all, to show the advantage of our proposed module, we replace ALE with a simple feature concatenation operation by concatenating the local features with the cropped features $\tilde{\textbf{X}}_M$ and $\tilde{\textbf{X}}_L$ (denoted as ``Concat" in the table). For reference, we also show the metrics of the model with ALE only. As observed in Table \ref{tab:abla4}, the naive feature concatenation results in performance drop on both benchmarks, even comparing to the baseline. It may be caused by the spatial misalignment of the features $\tilde{\textbf{X}}_M$ and $\tilde{\textbf{X}}_L$ with $\textbf{X}_I$, since $\tilde{\textbf{X}}_M$ and $\tilde{\textbf{X}}_L$ are cropped and upscaled from the original contextual features. Thus, the pixel-wise minor offsets between the feature maps may cause the errors in prediction. According to the results in Table \ref{tab:abla4}, ALE can well handle this issue and compensate the deficit of LCF. Combining LCF and ALE is able to yield better performance than the models with LCF or ALE only. Moreover, we show the predicted results with regard to two proposed modules in Fig.\ref{fig:abla_ev}, the results with both modules are optimal with refined results while the noise can be reduced by a large extent.
}

% \wxr{For validating the necessity of LCF and ALE, we replace them with the naive local-global feature concatenation for comparison, in which our proposed schemes are superior to feature concatenation ($73.47\%$ and $72.69\%$ vs $72.20\%$ on DeepGlobe; $73.50\%$ and $73.62\%$ vs $72.80\%$ on Inria Aerial in mIOU as Table \ref{tab:abla4}. For concatenating local-global features, we need to crop center regions that align with the local patches from the contexts and resize them to the same size as local patches, then concatenate them with local patch and reduce the dimension by $1 \times 1$ convolutional layer. Although it also benefits from contextual information from the contexts, it will be limited because it fuses local-global features pixel-to-pixel but the features are not perfectly aligned, which may leads to minor errors in position. A simple convolution operation is difficult to solve this defect, but our proposed scheme doesn't require pixel alignment. We uses all pixels to weight every pixel with positional information for LCF, and we apply a spatial attention with different window kernels to focus on local regions for ALE. In both strategies, they enlarges the receptive field and all contextual information is considered to enrich the local features with the misalignment.}

To delve into the structure design of ALE, we propose several variants of ALE, as illustrated in Fig.~\ref{fig:ALEs}. 
% For (a) and (b), we only employ one of $\tilde{\textbf{X}}_{M}$ and $\tilde{\textbf{X}}_{L}$ as the reference features of ALE, while we utilize both $\tilde{\textbf{X}}_{M}$ and $\tilde{\textbf{X}}_{L}$ for (c), (d), and (e). 
Since our ALE follows the principle of our algorithm, we can design the network structures with different topology and depths.
% , according to Eqs.~\ref{eq:optim1},~\ref{eq:optim2}, and~\ref{eq:optim3}. 
Firstly, we apply $\tilde{\textbf{X}}_{M}$ or $\tilde{\textbf{X}}_{L}$ to enhance local features for one or two recurrent steps, as show in Fig.~\ref{fig:ALEs}(a) and (b), respectively. Besides, we also apply $\tilde{\textbf{X}}_{M}$ and $\tilde{\textbf{X}}_{L}$ jointly to enhance local features for one, two, or three recurrent steps, as show in Fig.~\ref{fig:ALEs}(c), (d), and (e), respectively. Here, for the sake of simplicity, we denote the structures of (a) and (c) as the first-order local enhancement structure. (b) and (d) are denoted as the second-order local enhancement structure, while (e) is the third-order local enhancement structure.
% As the simplest design, we can 
% rewrite \wxr{Eq.~\ref{eq:optim1}} as $\hat{\textbf{X}}_I = \mathop{\arg\min}_{\textbf{X}_{I}} f(\textbf{X}_{I} ; \tilde{\textbf{X}}_{M})$, $\hat{\textbf{X}}_I = \mathop{\arg\min}_{\textbf{X}_{I}} f(\textbf{X}_{I} ; \tilde{\textbf{X}}_{L})$, or $\hat{\textbf{X}}_I = \mathop{\arg\min}_{\textbf{X}_{I}} f(\textbf{X}_{I} ; \{\tilde{\textbf{X}}_{M}, \tilde{\textbf{X}}_{L}\})$, corresponding to Fig.~\ref{fig:ALEs}(a) and Fig.~\ref{fig:ALEs}(c), respectively. We can denote them as the variants of ALE with the depth as $1$. For the variants with the depth as $2$, we can imitate the entire process of Eqs.~\ref{eq:optim1},~\ref{eq:optim2}, and~\ref{eq:optim3}. According to the different inputs, we have the network design as Fig.~\ref{fig:ALEs}(b) and Fig.~\ref{fig:ALEs}(d). Last, we duplicate the optimize process for two times and design the structure with the depth as $3$ in Fig.~\ref{fig:ALEs}(e). 
% In terms of iterations, there is one iteration in (a) and (c), (b) and (d) show two iterations structure and (e) is considered as three iterations with multiple LFI blocks.

As observed in Table \ref{tab:abla4}, the second-order structures (i.e., (b) and (d)) are better than those with the first-order ones (i.e., (a) and (c)). However, due to the introduction of too many LFI blocks in (e), there will be information over-exchange to degrade the performance. So, in the end, we select the structure (d) that achieves the best performance (73.91\% mIOU on DeepGlobe and 74.58\% mIOU on Inria Aerial).

\subsubsection{Network Efficiency}
\lqq{During the inference stage, we measure the memory costs and inference time using a single Titan Xp GPU with CUDA 11.0, CuDNN 8, and PyTorch 1.7.1. 
% Specifically, we use command ``nvidia-smi" to observe memory costs and use commands "torch.cuda.synchronize()" and "time.time()" to calculate inference time.
Our local segmentation model costs around 3691MB memory for each patch of the images in DeepGlobe and Inria Aerial, which do not increase much computation overhead over FCN-8s (2477MB). The memory usage of our model is comparable to existing semantic segmentation models. For the timing performance, our segmentation model processes each $508 \times 508$ patch in 0.16s per patch during inference. Overall, for each instance from DeepGlobe and Inria Aerial without TTA, our model costs 6 and 20 seconds, respectively, compared with FCN-8s (3 and 9 seconds), GLNet (6 and 19 seconds), and CascadePSP (9 and 37 seconds).}
% As an independent model from the segmentation model, our refinement network costs 1165MB memory. Thus, the memory usage of our model is comparable to existing semantic segmentation models.For the timing performance, our segmentation model processes each patch in 0.15s and our refinement model requires 0.06s per patch during inference. Overall,for each instance from DeepGlobe and Inria Aerial, our model needs to cost 8s and 26s, compared with FCN-8s (3s and 9s), GLNet (6s and 19s), and CascadePSP (9s and 37s).

% \begin{table}
%     \centering
%     \setlength{\tabcolsep}{4pt}
%     \footnotesize
%     \caption{Context sizes for mask refinement on public datasets.}
%     \begin{tabular}{c|c|cc}
%         \toprule
%         Refinement & Context Size  & DeepGlobe & Inria Aerial\\
%         % \midrule
%         % Fusion & mIOU (\%)  \\
%         \midrule
%         Montaging & - & 73.22 & 73.45\\
%         Averaging & - & 73.22 & 73.53   \\
%         \midrule
%          &  762 & 73.24 & 73.63\\
%         Ours & \textbf{1016} & \textbf{73.45} & \textbf{73.66}\\
%          & 1270 & 73.36 & 73.63\\
%         \bottomrule
%     \end{tabular}
%     \label{tab:abla4}
% \end{table}

\section{Conclusion}
\label{sec:conclusion}
In this paper, we innovate the widely used high-resolution image segmentation pipeline. In particular, we introduce a novel locality-aware context fusion based segmentation model to process local patches, where the relevance between local patch and its various contexts are jointly and complementarily utilized to handle the semantic regions with large variations. Additionally, we present the alternating local enhancement module that restricts the negative impact of redundant information introduced from the contexts, and thus is endowed with the ability of fixing the locality-aware features to produce refined results.
% we present a contextual semantics refinement network that is enabling to reduce the boundary artifacts and refine mask contours during the process of creating the final high-resolution mask. 
% Furthermore, in comprehensive experiments, we demonstrate that our model outperforms other state-of-the-arts in public benchmarks. 

However, it remains challenging for our model to identify the complex landscapes in the geospatial images. For instance, ``agriculture", ``rangeland", and ``forest" are easily confused in some situations, where even humans also have to take time to discern their semantics carefully. Besides, our model suffers from class imbalance problem to some extent, which leads to sub-optimal performance in certain subject categories with less training samples.

For ultra-high resolution image segmentation, the major issues rest in the data scarcity and the difficulty to manually annotate the semantics for those images. As the future work, we may consider to fully utilize the existing segmentation datasets and pretrained models to strengthen the task of ultra-high resolution image segmentation via overcoming the domain gap. On the other hand, existing models tend to sacrifice the segmentation performance for model efficiency. Moreover, deploying the ultra-high resolution image segmentation model on hardware devices with limited computation resources is a more challenging task along this direction.
\lqq{Deep network compression technique can be applied to achieve a memory-efficient and hardware-deployable model. 
Also, we can explore a network structure design strategy that allows the model to automatically select the number of contextual levels instead of a fixed number of contexts, which may adaptively reduce the computation resource.}

\noindent\textbf{Data Availability Statement.} The employed datasets DeepGlobe\footnote{https://competitions.codalab.org/competitions/18468} and Inria Aerial\footnote{https://project.inria.fr/aerialimagelabeling} in our work are publicly available for research.

\bibliography{egbib}% common bib file


\end{document}